\documentclass[review]{elsarticle}
\usepackage[10pt]{extsizes}
\usepackage[a4paper, total={6in, 9in}]{geometry}

\usepackage{hyperref}
\usepackage{amsmath}
\usepackage{amssymb}
\usepackage{bbold}
\usepackage{mathtools}
\usepackage{stmaryrd}
\usepackage{tabulary}
\usepackage{placeins}
\usepackage[usenames,dvipsnames]{xcolor}
\usepackage[export]{adjustbox}
\usepackage{subfig}
\usepackage{float}
\usepackage{cleveref}
\usepackage{xcolor}
\usepackage[T1]{fontenc}
\newcommand{\david}[1]{\textcolor{black}{#1}} 
\newcommand{\chiara}[1]{\textcolor{black}{#1}} 

\newcommand{\eugenia}[1]{\textcolor{black}{#1}}
\newcommand{\final}[1]{\textcolor{black}{#1}}

\usepackage{comment}

\usepackage{array}

\newcolumntype{F}[1]{%
    >{\raggedright\arraybackslash\hspace{0pt}}p{#1}}%
\newcolumntype{T}[1]{%
    >{\centering\arraybackslash\hspace{0pt}}p{#1}}%
\newcolumntype{P}[1]{>{\centering\arraybackslash}p{#1}}
\newcolumntype{M}[1]{>{\centering\arraybackslash}m{#1}}

\usepackage{graphicx} 
\usepackage{float}	
\usepackage[colorinlistoftodos]{todonotes}\reversemarginpar
\graphicspath{{img/}} 
\usepackage{xspace}
\newcommand{\ourmethod}{AcME\xspace}
\newcommand{\scoresname}{standardized effects\xspace}
\newcommand{\scorename}{standardized effect\xspace}

\begin{document}

\begin{frontmatter}

\title{
\final{AcME - Accelerated Model-agnostic Explanations:\\Fast Whitening of the Machine-Learning Black Box}}

\author[1]{Dandolo David\corref{cor2}}
\author[1]{Masiero Chiara}
\author[2]{Mattia Carletti}
\author[2]{Davide Dalle Pezze}
\author[2]{Gian Antonio Susto\corref{cor1}}

\cortext[cor2]{A Python implementation of the approach proposed in this work is available at \url{https://github.com/dandolodavid/ACME}.}
\cortext[cor1]{Corresponding author. \\
This work has been partially supported by the Regione Veneto project ExplAIn 4.0 and by MIUR (Italian Minister for Education) under the initiative "Departments of Excellence" (Law 232/2016). Statwolf Data Science is also acknowledged for its financial support.}

\address[1]{Statwolf Data Science, Padova, Italy. E-mail: david.dandolo@statwolf.com, davide.dallepezze@statwolf.com, chiara.masiero@statwolf.com }
\address[2]{Universit\'a degli Studi di Padova, Padova, Italy. E-mail: mattia.carletti@unipd.it, gianantonio.susto@unipd.it}


\begin{abstract}


\final{In the context of human-in-the-loop Machine Learning applications, like Decision Support Systems, interpretability approaches should provide actionable insights without making the users wait. In this paper, we propose Accelerated Model-agnostic Explanations (AcME), an interpretability approach that quickly provides feature importance scores both at the global and the local level. AcME can be applied \textit{a posteriori} to each regression or classification model. Not only AcME computes feature ranking, but it also provides a \textit{what-if} analysis tool to assess how changes in features values would affect model predictions. We evaluated the proposed approach on synthetic and real-world datasets, also in comparison with SHapley Additive exPlanations (SHAP), the approach we drew inspiration from, which is currently one of the state-of-the-art model-agnostic interpretability approaches. We achieved comparable results in terms of quality of produced explanations while reducing dramatically the computational time and providing consistent visualization for global and local interpretations. To foster research in this field, and for the sake of reproducibility, we also provide a repository with the code used for the experiments. }

\end{abstract}


\end{frontmatter}


\section{Introduction}
\label{sec:intro}
With the increase of computational power and the growing availability of data, Machine Learning (ML) approaches have been successfully applied to many scientific, technological, and business areas \cite{jordan2015machine, KANG2020106773, SMITI2020100280, VANKLOMPENBURG2020105709, li58social, harb2020framework}. In particular, ML has paved the way for more effective Decision Support Systems (DSS). Indeed, ML can play a key role in processing the ever-growing amount of data available to organizations into actionable insights provided through DSS 
\cite{bohanec2021heartman, 8987964, diagnosis_DSS, guerrero2022decision,  LEE2020157, MA2020481, Pirani2017ApplicationOB, yu2021medical}.
To make the most from the advanced data analytics capabilities enabled by ML, analysts should be able to \chiara{get explanations} about model predictions so they can make \chiara{faster}, better, \david{and more reliable} decisions based on the information processed by ML algorithms \cite{AYOUB2021102569, Bauer2021ExplAInIT,   fiok2021analysis, PRZYBYLA2021102653, rico2021machine, SHIN2021102551}. 

On the one hand, some ML methods (e.g. linear regression, simple decision trees) are intrinsically interpretable and provide a feature importance score for each of the input variables.  On the other, the most advanced ML models that allow for better performance (e.g. neural networks) work just like "black boxes": they provide predictions to the users, but no hint about how they compute them. \chiara{Not only may this lack of interpretability conceal fairness issues \cite{MISZTALRADECKA2021102519,PANIGUTTI2021102657}, but it also often translates into a lack of trust} that hinders the adoption of ML. This is particularly detrimental in the context of DSS \cite{Ebadi2019HowCA,doshivelez2017rigorous,MILLER20191} \final{and other socio-technical systems \cite{andras2018trusting}}.
Interpretable Machine Learning is a research field that aims at shedding light on how black-box ML model predictions depend on input features \cite{Murdoch2019DefinitionsMA}.
ML interpretability methods can be classified into \emph{model-specific} and \emph{model-agnostic} ones. The former, also known as \emph{ad-hoc} methods, are designed for a specific class of models \cite{molnar, diffi}. The latter, also known as \emph{post-hoc} methods, can be applied on top of every ML model \cite{expl}. 

In this work, with DSS in mind, we focus on model-agnostic approaches. Indeed, in many practical scenarios, different ML models can be used to address many data processing tasks needed to distil the information to be exposed through the DSS. However, users may lack a background in ML, so it can be difficult for them to deal with different feature importance indicators produced by different methods. For them, having to deal with a unique set of feature importance indicators, such as those computed by a model-agnostic explainability approach, is undoubtedly a great advantage. Indeed, less training is needed and, given a task (e.g. regression or classification),  users are always confronted with the same model evaluation interface, independently from the actual ML model implemented to solve it.

From the perspective of the level of granularity, interpretability algorithms can be further divided into two classes: \emph{global} interpretability methods, aimed at providing explanations of the model as a whole, and \emph{local} interpretability methods, aimed at providing explanations associated with individual predictions. Of course, both global and local interpretability are useful in DSS. On the one hand, global interpretation can be used to build trust in the predictive model; on the other, local interpretation provides users with contextual information that can be relevant in the decision-making process (e.g. in root cause analysis).

While it is apparent that in human-in-the-loop applications interpretability insights must be provided to the user in a reasonable amount of time, most state-of-the-art interpretability approaches require time-consuming procedures for computation that do not allow for 'on-the-fly' operation.

In this work, after addressing the advantages and disadvantages of the most popular ML interpretability solutions in a DSS-oriented perspective, we propose a novel methodology, called Accelerated Model-agnostic Explanations (AcME), that works with each regression or classification model; not only AcME provides both global and local interpretability, but it is also computationally efficient when compared to current state-of-the-art approaches.

\section{Related Work}
\label{sec:related_work}
Among model-agnostic approaches, Partial Dependence Plots (PDPs)  \cite{pdp} are one of the most well-known: PDPs simply make each variable vary in a range and show the annexed predictions in a plot. While PDPs provide a rich description of the effect each feature has on the prediction, such approach requires a non-negligible effort by the user to analyse and compare a potentially large number of plots (one for each feature). Moreover, PDPs focus on global interpretability only. 
An alternative approach is given by Permutation Importance, introduced for Random Forests in \cite{breiman2001random}, and its variants, such as the  model-agnostic variant proposed in \cite{permutation} or the Conditional Permutation Importance described in \cite{strobl2008conditional}. These approaches are based on random permutations of values for each feature and produce a ranking of features according to the resulting importance score. Results are very easy to understand, but they provide only global level interpretability.
 
As for local interpretability, a key contribution was given in \cite{LIME}, where authors propose Local Interpretable Model-agnostic Explanations (LIME) approach. LIME is one of the interpretability techniques that take advantage of data disturbances, such as EXPLAIN and IME \cite{vstrumbelj2009explaining}. 
LIME is based on the training of local surrogate model to approximate the predictions of the underlying black-box model, to explain individual predictions \cite{perturbation}.

Among explainability techniques that can provide both global and local interpretation, one that has received a lot of interest recently is SHapley Additive exPlanation (SHAP) \cite{shap_article}. This technique has encountered great success and it has also been extended in different ways \cite{antwarg2021explaining, giudici2021shapley,interpretLSTM, xgboost-shap}; however, SHAP has some drawbacks. First, SHAP is computationally burdensome, as it will be discussed in the next section. Second, it is a complex approach, not only  for its theoretical basis, but also in terms of the resulting plots, that can be difficult to read, especially for users with no ML background. 
About the methodological foundations, they seem to be not fully understood by many users, potentially leading to misuses of the tool \cite{shap-problem}. While the last issue could be eased by a better training on the theoretical aspects (even though many end-users may lack the technical background to grasp them or the time to invest in such training), the computational complexity could be a bottleneck in situations where multiple models have to be compared, or limited computational resources are available, or fast reaction based on model prediction is required.
As for the computational speed, it is nowadays a challenge, particularly in the Web-of-Things context, where the amount of data is considerable, often in the form of live streams with extremely fast update. 
Besides, a fast update of data leads to a more frequent retrain of the ML models, and when a model changes, it is necessary to rerun the entire ML interpretability procedure, which worsens the computational burden associated to SHAP.
In this paper, we draw inspiration from the versatility of SHAP and the simplicity of the computations behind PDPs, and, as stated above, we propose \ourmethod, a new model-agnostic approach for both global and local interpretability. Similarly to SHAP, \ourmethod produces effective data visualization for global interpretability. Also, it brings the same effectiveness to visualization for local interpretability. Moreover, it remarkably improves computational efficiency. Despite being much faster, the proposed approach proves to be comparable to SHAP in terms of feature impact evaluation.
The rest of the paper is organized as follows: in \Cref{sec:shap} we will explain the SHAP procedure, with particular focus on KernelSHAP, while in \Cref{sec:acme} we present \ourmethod. Finally, in \Cref{sec:exp_results} we describe experiments on the two methods, comparing and analyzing the results on both synthetic and real datasets.

\section{SHAP}\label{sec:shap}
As mentioned in \Cref{sec:related_work}, SHAP \cite{shap_article} is a framework for interpreting predictions produced by a ML model based on the concept of Shapley value from cooperative game theory. Given a trained model $f$ and a generic data point $\mathbf{x}$, represented by a $p$-dimensional feature vector (i.e. $\mathbf{x} \in \mathbb{R}^p$), SHAP computes, for each feature, a real-valued quantity that represents the contribution of the feature to the prediction $f(\mathbf{x})$. 
The main idea is that the prediction can be explained by treating it as a "payout" that needs to be distributed across features, which act as “players” in a coalition. As many other interpretability methods, SHAP relies on the definition of an \emph{explanation model} $g$ which is simpler than the predictive model $f$ to be explained, while being a good approximation thereof at least locally. Specifically, the explanation model $g$ is a linear function of binary variables $z_j$, indicating the presence or absence of the corresponding feature in the sampled coalition, and takes the form

\begin{equation} \label{eq:expl_model}
    g(\mathbf{z}) = \phi_0 + \sum_{j=1}^{p} \phi_j z_j.
\end{equation}

The $p$-dimensional vector of ones and zeros $\mathbf{z} = [z_1, \dots, z_p]$ represents the sampled coalition: $z_j=1$ denotes presence of the $j$-th feature in the coalition, while $z_j=0$ indicates that the $j$-th feature is not in the coalition. At the core of the SHAP method is the estimation of the linear model $g$ given $K$ sampled coalitions $\mathbf{z}^{(1)}, \dots, \mathbf{z}^{(K)}$, which allows for a straightforward computation of the feature attribution coefficients $\phi$'s, i.e. the Shapley values.

Notice that, in light of the nature of the underlying explanation model \eqref{eq:expl_model}, SHAP can be considered as an additive feature attribution method. The most popular model-agnostic version of SHAP is \emph{KernelSHAP}, characterized by high portability as it can be used on any ML model. Model-specific versions have been proposed for tree-based models (\emph{TreeSHAP} \cite{tree_shap}) and for Deep Learning models (\emph{DeepSHAP} \cite{shap_article}), both with the aim of leveraging the internal structure of the model at hand to improve the computational performance of the algorithm. {Since our main goal is the design of a completely model-agnostic interpretability method, in the remainder of this work we mainly consider KernelSHAP as benchmark for comparisons}. 

\subsection{KernelSHAP}\label{kernel_shap}
At the core of the \textit{KernelSHAP} method is the estimation of the linear explanation model \eqref{eq:expl_model} by means of the optimization of a squared loss function in which the contribution of each sampled coalition $\textbf{z}$ is weighted according to a specific weighting kernel (whose rationale and properties are discussed below). 
The algorithm can be summarized in the following steps:
\begin{itemize}
    \item For every data point in the dataset $i=1,\dots,N$: 
    \begin{enumerate}
    \item Sample $K$ coalitions $\textbf{z}^{(k)} \in\{0,1\}^p, k\in\{1,\dots,K\} $, and obtain $Z = \{\textbf{z}^{(1)},\dots, \textbf{z}^{(K)}\}$;
    \item Map each coalition vector $\textbf{z}$ to the original feature space and obtain $\textbf{x}' = h_{\mathbf{x}}(\textbf{z})$, where $h_{\mathbf{x}}$ is a function from $\{0,1\}^p$ to $\mathbb{R}^p$ mapping $1$s into the original feature values of data point $\mathbf{x}$ and $0$s into feature values randomly sampled from the dataset (see the example in \Cref{tab:coalition_to_feature});
    \item Get the prediction produced by the trained model $f$:  $\widehat{y}'=f(\mathbf{x}')$;
    \item Compute the weight for each coalition $\mathbf{z}$ according to the SHAP kernel:
        $$\pi_{x}(\mathbf{z})=\frac{(p-1)}{\binom{p}{|\mathbf{z}|}|\mathbf{z}|(p-|\mathbf{z}|)},$$
    where $|\mathbf{z}|$ denotes the number of present features in coalition $\mathbf{z}$. Notice that the coalitions that are given the highest weights, and therefore considered as most informative, are those consisting of few features (where $|\mathbf{z}|$ is small) and those consisting of many features (where $|\mathbf{z}|$ is large);
    \item Fit the linear explanation model $g$ by optimizing the loss function:
    $$ L(f,g,\pi_{x})=\sum_{\mathbf{z} \in Z} [f(h_{\mathbf{x}}(\mathbf{z}))-g(\mathbf{z})]^2 \pi_{x}(\mathbf{z}) ;$$
    \item Return Shapley values, i.e. the estimated coefficients from the linear explanation model $\phi_j$ for $j=1, \dots, p$.
    \end{enumerate}
    \item To get the global feature importance scores, the absolute Shapley values can be aggregated across the data points in the dataset. The global feature importance score for the generic feature $j$ can be computed as follows:
    $$ I_j=\sum_{i=1}^{N}|\phi_j^{(i)}|.$$
\end{itemize}

\begin{table}[]
\scriptsize
\centering
\begin{tabular}{llllllllll}
\hline
Age & Weight & Height & Sex & &  Age & Weight [kg] & Height [cm] & Sex\\
\hline
1 & 1 & 1 & 1 & $\longrightarrow$ & 24 & 88 & 185 & M \\ 
1 & 0 & 0 & 0 & $\longrightarrow$ & 24 & \textbf{60}  & \textbf{193} & \textbf{F} \\ 
1 & 1 & 0 & 0 & $\longrightarrow$ & 24 & 88 & \textbf{178}& \textbf{F} \\ 
1 & 1 & 1 & 0 & $\longrightarrow$ & 24 & 88 & 185 & \textbf{F} \\ 
\hline
\end{tabular}
\caption{An example of mapping from the space of coalition vectors to the original feature space, with absent features replaced with feature values (in \textbf{bold}) randomly sampled from the dataset.}
\label{tab:coalition_to_feature}
\end{table}

The authors of the method suggest to use $K = 2048 + 2 \cdot p$ coalitions \cite{shap_user_guide} to obtain the best results in the majority of situations. The higher the number of evaluated coalitions, the higher the computational time, due to the larger dimension of the matrix to be inverted in the estimation of the linear explanation model. On the other hand, a larger number of coalitions could ensure the estimation of a more accurate explanation model, assumed the sampled coalitions are sufficiently informative. For this purpose, the coalitions are not chosen at random: the weight $\pi_{x}(\mathbf{z})$ is exploited to select top $K$ most informative ones, i.e. those whose associated weights are the highest.
Notice that the number of sampled coalitions (according to the authors' suggestion) increases with the number of features. In addition, the number of linear models that have to be estimated increases with the number of data points in the dataset. As a consequence, the computational cost does not scale well with the dataset dimensions and this is the reason why \textit{KernelSHAP} is computationally burdensome, especially in big data regimes.
In \Cref{sec:real_data} we will analyze different solutions the authors provide to lower the computational time, highlighting the impact such trade-offs have on the quality of the produced explanations.

\subsection{SHAP results visualization}\label{shap_viz}
\begin{figure}[t]
    \centering
    \subfloat[SHAP waterfall plot. \label{fig:shap_waterfall}]{\includegraphics[width=0.55\textwidth]{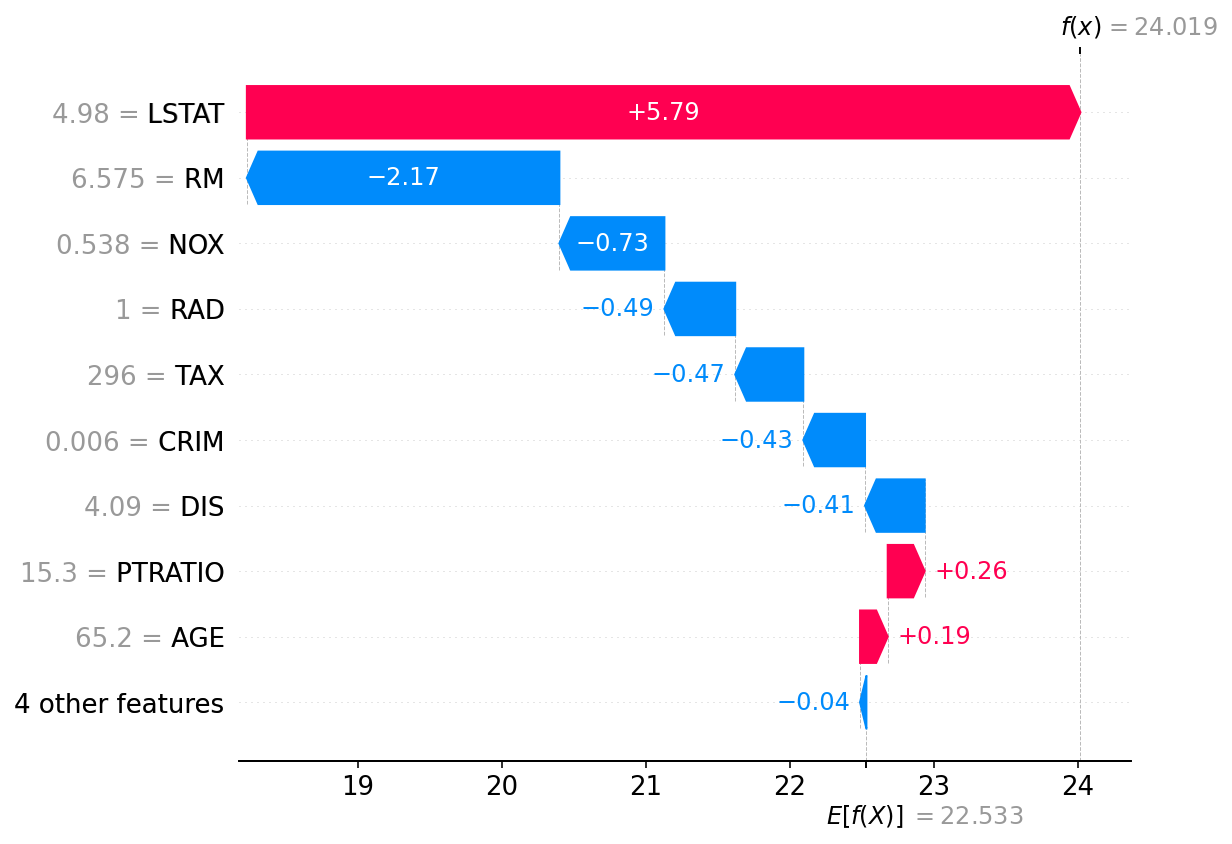}}
    \vspace{0.5cm}
    \subfloat[ SHAP for local importance. \label{fig:SHAP_local}]{\includegraphics[width=0.8\textwidth]{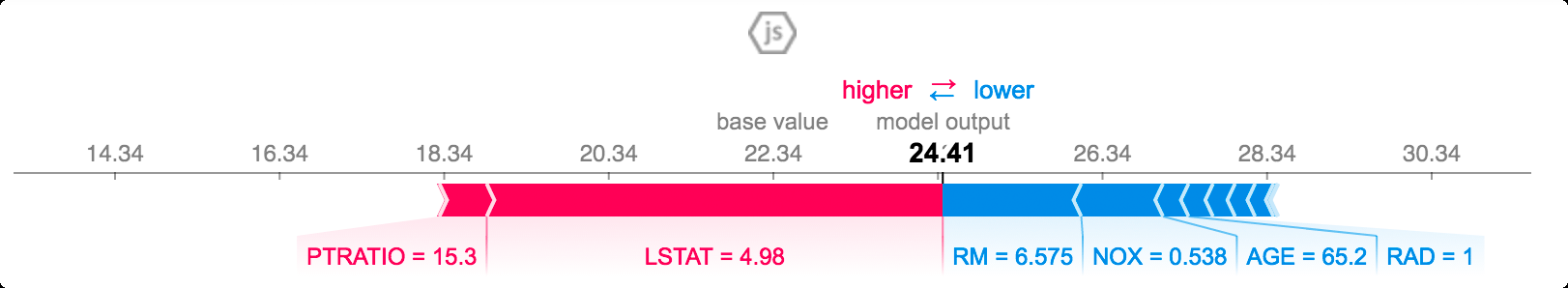}} 
    \caption{SHAP local importance plot : Waterfall and Force plot} 
\end{figure}

\chiara{Before introducing \ourmethod, we briefly introduce the most frequently encountered visualizations of SHAP results, that will be useful to better assess the visual efficacy of the novel approach detailed in the following sections.}
\david{ SHAP global importance \chiara{is a jitter plot that} shows both feature effects and feature importance. The y-axis has the feature names sorted by decreasing importance score, while on the x-axis there are the Shapley values. The color map, \eugenia{from} blue to red, represents the value of the feature from low to high. Overlapping points are jittered in y-axis direction to get a sense of the distribution (Figure \ref{fig:model1}b). 
As for local interpretation, SHAP provides two different plots: the waterfall plot (Figure \ref{fig:shap_waterfall}) and the force plot (Figure \ref{fig:SHAP_local}).
They are both based on the concept that each \chiara{SHAP} value is a force that either increases or decreases the estimation. 
The prediction starts from a baseline, that is the average of all predictions. In \chiara{local importance plots}, each positive Shapley value is an arrow that increases the prediction, while negative values decrease it. Balancing each other, the arrows point to the actual prediction for the selected observation. The difference is that, while the force plot has all the arrows on the same row divided in positive values (on the left) and negative values (on the right), the waterfall plot has one row for each arrow, ordered by impact. \chiara{Both visualizations are very different from the one used for global interpretation. One shortcoming is that they do not provide any hint about features distributions to contextualize the values of current observation.}}

\section{Proposed approach: \ourmethod}\label{sec:acme}
In this work, we propose a new approach to model explainability, \href{https://github.com/dandolodavid/ACME}{\ourmethod}\footnote{A Python implementation of the proposed approach is made available at the following link https://github.com/dandolodavid/ACME}, aimed at analyzing the role played by each feature at both global and local scale. The importance scores provided by \ourmethod rely on perturbations of the data based on quantiles of the empirical distribution of each feature. These perturbations are performed with respect to a reference point in the input space, which we call \emph{baseline vector} (denoted $\mathbf{x^b}$). \chiara{First we deal with regression tasks}. The procedures exploited to get global and local importance scores, described in \Cref{sec:acme_global} and \Cref{sec:acme_local}, respectively, mainly differ in the choice of the baseline vector and the produced visualizations. 
In \Cref{sec:classification}, we extend the proposed approach to classification tasks.
As supported by the experimental results in \Cref{sec:exp_results}, the rationale behind \ourmethod allows for a substantial reduction in computational time, while retaining a quality of explanations comparable to state-of-the-art interpretability method SHAP. 
\chiara{Besides, \ourmethod provides very similar visualizations both for global and local interpretability. They are reminiscent of SHAP global importance visualization, but simpler, provided the number of considered quantiles in score evaluation is low. Moreover, local visualization can be used as \textit{what-if} analysis tools to assess how changes in each feature values would impact model prediction for a specific observation.} 

\subsection{Global interpretability for regression tasks} \label{sec:acme_global}
When dealing with global interpretability, we consider the mean vector $\mathbf{\bar{x}}$, i.e. the $p$-dimensional vector whose components are the mean values of the features (computed over the whole dataset), as the baseline vector $\mathbf{x^b}$. As previously mentioned, the baseline vector represents the point with respect to which perturbations (and corresponding predictions) are computed. \david{\chiara{To evaluate the importance of each feature $j \in {1,\dots,p}$}, we create a variable-quantile matrix $\mathbf{Z}_j \in \mathbb{R}^{Q \times p}$ whose rows are identical to the baseline vector except for the $j$-th component, which is substituted by the $Q$ quantiles of the empirical distribution of the processed feature $j$}. 
Notice that the number of rows in $\mathbf{Z}_j$ can be easily tuned by changing the number of selected quantiles $Q$. By comparing the predictions associated with the rows of the variable quantile matrix $\mathbf{Z}_j$ with the prediction associated with the baseline vector, an importance score representing the relevance of \chiara{the $j$-th feature} can be computed. Specifically, the whole procedure for the computation of the global importance score \chiara{for feature $j$} can be summarized as follows:
\begin{enumerate}
    \item Compute the baseline vector: \quad $\mathbf{x^b} =  \mathbf{\bar{x}} = [ \bar{x}_1, \dots , \bar{x}_{j-1}, \bar{x}_j, \bar{x}_{j+1}, \dots, \bar{x}_p ]^T$.
    \item For each 
    $q\in\lbrace 0,1/(Q-1),2/(Q-1), \dots, 1 \rbrace$, create the new vector $\mathbf{z}_{j,q} \in \mathbb{R}^p$. This is obtained from $\mathbf{\bar{x}}$ by substituting $\bar{x}_j$ with $x_{j,q}$ i.e., the value of quantile $q$ for the $j$-th variable:
    \begin{equation*}
         \mathbf{z}_{j,q} = [ \bar{x}_1, \dots , \bar{x}_{j-1}, x_{j,q}, \bar{x}_{j+1}, \dots, \bar{x}_p ].
    \end{equation*}
   To be more robust, we can avoid the use of quantile 0 and 1, limiting the range of $q$ to an interval that excludes the possible outliers; in this case, an appropriate interval could be for example from $q=0.1$ and $q=0.9$. For simplicity, in the rest of the paper we will use the complete range of quantiles, however the procedure remains unchanged.
    \item \final{Create the variable-quantile matrix for feature $j \in \{1,\dots,p\}$: 
    $$
    \mathbf{Z_j} = \begin{bmatrix} 
        \mathbf{z}_{j,0} \\
        \mathbf{z}_{j,1/(Q-1)} \\
        \vdots \\
        \mathbf{z}_{j, \; 1}\\
    \end{bmatrix} = \begin{bmatrix} 
        \bar{x}_1 & \bar{x}_{2} & \dots & x_{j,\;0} & \dots & \bar{x}_p  \\
        \bar{x}_1 & \bar{x}_{2} & \dots & x_{j,\;1/(Q-1)} & \dots & \bar{x}_p  \\
        \vdots & \vdots & \ddots & \vdots & \ddots & \vdots\\
        \bar{x}_1& \bar{x}_{2}& \dots& x_{j, \;1}& \dots& \bar{x}_p
    \end{bmatrix}
    $$}
    \item \final{Compute predictions associated with the variable-quantile matrix rows: 
    $$
    \hat{\mathbf{y}}_j = \begin{bmatrix} 
        \hat{{y}}_{j, 0} \\
        \hat{{y}}_{j,1/(Q-1)} \\
        \vdots \\
        \hat{{y}}_{j, 1} \\
    \end{bmatrix} = \begin{bmatrix} 
        {f}(\mathbf{z}_{j,0}) \\
        {f}(\mathbf{z}_{j,1/(Q-1)}) \\
        \vdots \\
        {f}(\mathbf{z}_{j,1})
    \end{bmatrix}
    $$}
    
    \item Calculate the \scorename: 
    \begin{equation}
    \Delta_{j,q} = \frac{ \hat{y}_{j,q}-{f}(\mathbf{x^b}) }{\sqrt{{var}(\hat{\mathbf{y}}_j)}}(max(\hat{\mathbf{y}}_j) - min(\hat{\mathbf{y}}_j)).\label{eq:std_eff}
    \end{equation}
    \item The global feature importance score for the generic feature $j$ can be computed by averaging the magnitude of standardized effects over the quantiles:
    \begin{equation}
        I_j = \frac{1}{Q} \sum_{q=1}^{Q}|\Delta_{j,q}|.
        \label{eq:global_importance}
    \end{equation}
\end{enumerate}

The first multiplicative factor in \Cref{eq:std_eff} is similar to the well-known standard score: the objective of this standardization is to achieve scale invariance in the evaluation of the change in prediction caused by using the quantile $q$ (in place of the baseline value). The second multiplicative factor, instead, accounts for the overall impact of the variable in terms of how the applied perturbations spread out predictions. Indeed, it is reasonable that the wider the range of changes in the prediction, the more relevant the variable is for the model.

Notice that for categorical variables, the mode is used in place of the mean value as baseline. Then, the $M$ distinct values the feature could assume the role of quantiles.

\ourmethod has in general lower computational complexity than KernelSHAP. 
\chiara{Indeed, \ourmethod only needs to apply the model on $Q\times p$ observations, corresponding to the vectors $\mathbf{z}_{j,q}$. The number of observation $N$ in the dataset that is used to get explanations only affects the computational burden required to calculate the quantiles (for efficient approaches to compute approximate quantiles in large scale data, we refer the reader to \cite{Chen2020_quantile_survey}). On the other hand, as detailed in \Cref{sec:shap}, to provide global interpretability, KernelSHAP requires to train a linear model $N$ times, once for each point in the evaluation dataset. First, for each data point $x_i$, $i \in \{1, \dots, N\}$, KernelSHAP samples $K$ coalitions. Then, it applies the model that has to be interpreted to each coalition. Finally, KernelSHAP uses the $K$ predictions as a training dataset to fit a local linear model. Notice that, according to the documentation, the choice of $K$ should also depend on $p$. \\
The effectiveness 
of \ourmethod is amplified when explanations of high-dimensional datasets are required in real-time, a scenario in which computationally intensive algorithms are not viable.}

In addition, the use of the quantiles enhances robustness against the presence of outliers (not rare in large datasets), and could be a solution to the lack of information in small datasets. In fact, with few observations the estimated density could allow to generate unobserved values, instead of using only observed values permutation.

\paragraph{Results visualization}

\begin{figure}
\centering
		\subfloat[\ourmethod global importance scores \label{rf_acme_bar}]{\includegraphics[width=0.47\textwidth]{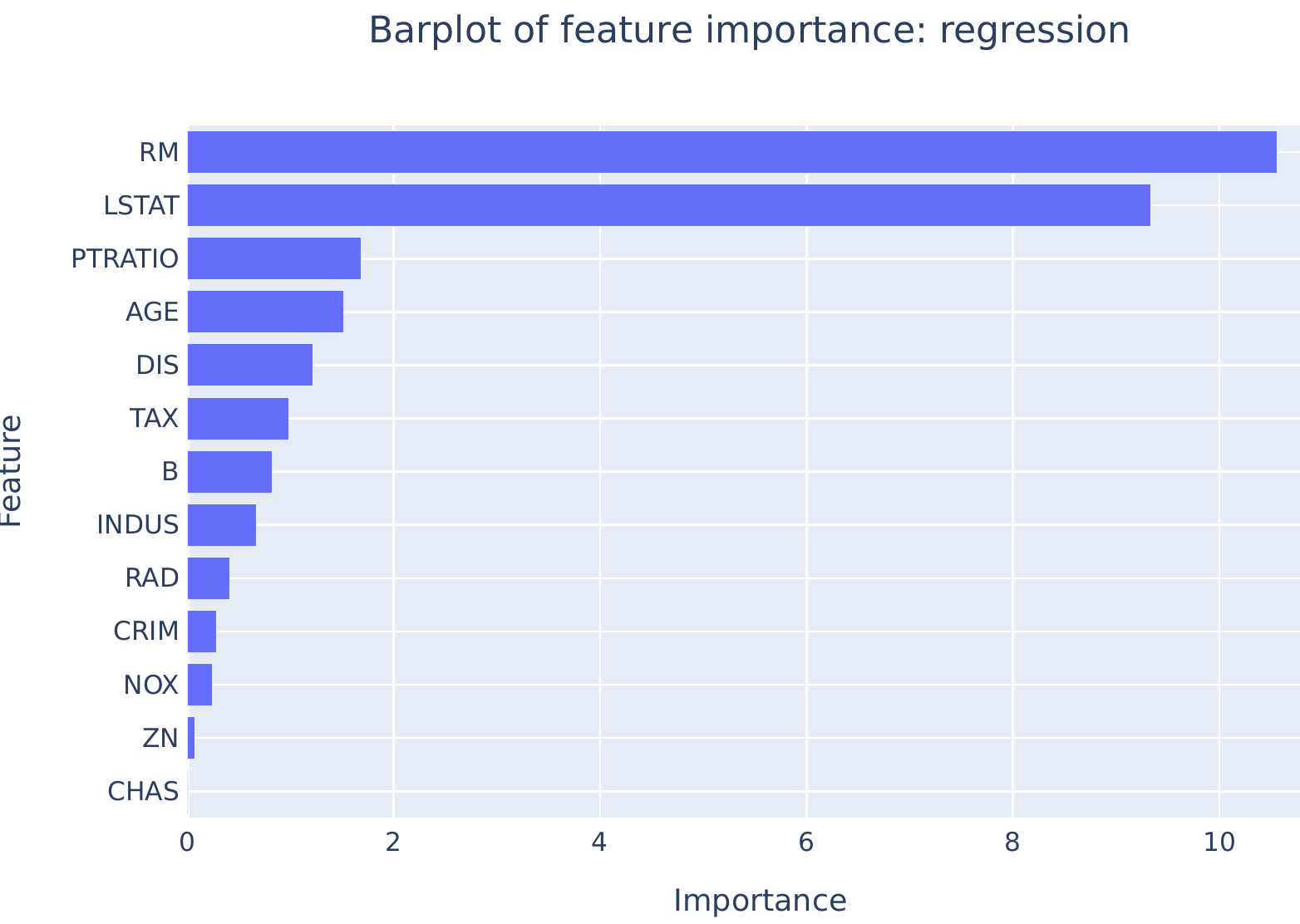}}
		\hspace{0.2cm}
		\subfloat[\ourmethod global importance detailed visualization]{\includegraphics[width=0.5\textwidth]{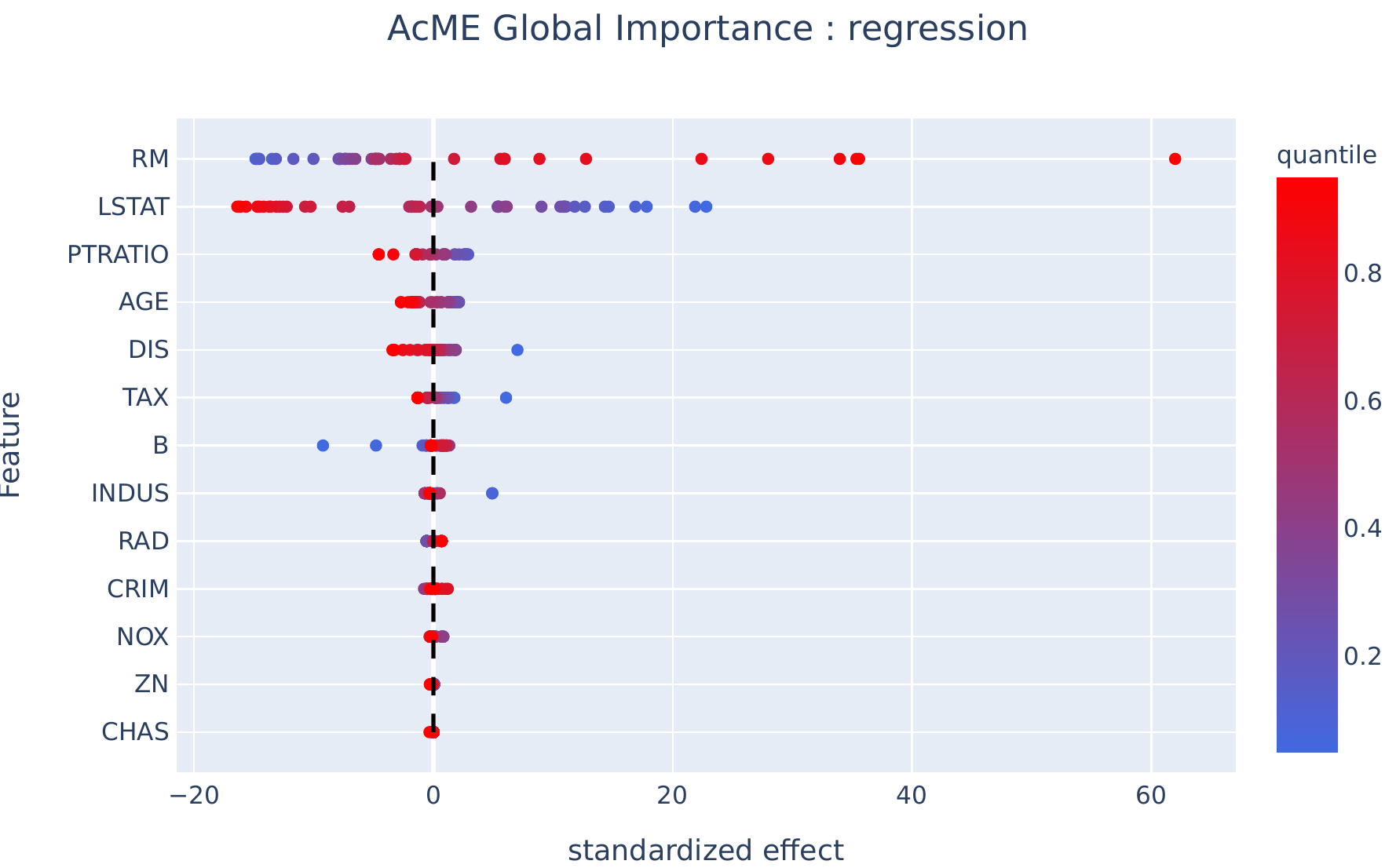}}
    \caption{\ourmethod on a regression task (Boston Housing dataset, RF model, see \Cref{sec:real_data} for details)}
    \label{fig:regression_acme}
\end{figure}

\chiara{For the visualization of global feature importance scores calculated by \ourmethod, we propose two different kinds of plots. The first one is simpler, while the second one is more informative. \\
The former is just a bar plot that shows the feature scores computed according to Equation \eqref{eq:global_importance}, in decreasing order. An example is depicted in \Cref{fig:regression_acme}(a). This high-level visualization is designed to figure out quickly which features are the most relevant for the model. Thus, we can use this visualization as a diagnostic tool, for example, when comparing different models.\\
As for the second visualization, we draw inspiration from SHAP global interpretability plots, since they are concise and effective.} An example of AcME global importance detailed score visualization is given in in \Cref{fig:regression_acme}(b): on the $y$ axis the features are sorted in decreasing order of importance according to \Cref{eq:global_importance}, while the \scoresname for each element of the variable-quantile matrix are plotted along the $x$ axis. In other words, each horizontal line (associated with a specific feature) represents the \scoresname for the $Q$ perturbations based on quantiles.
\eugenia{The color represents the quantile level of the feature from low (marked in blue) to high (marked in red). Moreover the ACME visualization provides a black dashed line, \chiara{corresponding to the prediction for the base point}, to separate positive effects, i.e. those pushing the prediction to higher values, from negative effects, i.e. those pushing the prediction to lower values.}

\begin{figure}
    \centering
    \includegraphics[scale=0.32]{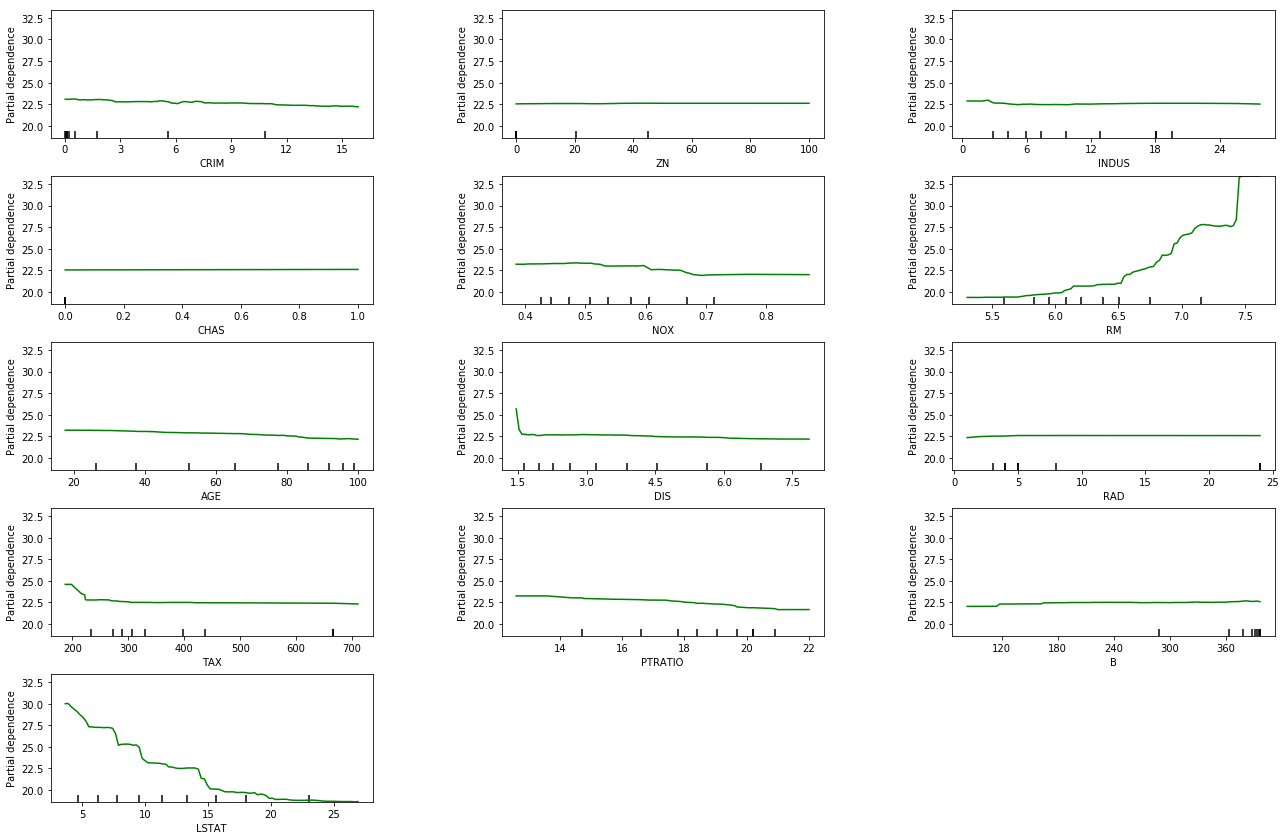}
    \caption{PDP calculated for a Random Forest with 100 trees on the Boston Housing dataset (see \Cref{sec:real_data} for details).}
    \label{fig:pdp}
\end{figure}

Compared to PDPs visualization (see the example in \Cref{fig:pdp}), \ourmethod offers a remarkable improvement as it provides a condensed visualization that makes the analysis of results immediate and effective. We recall that the PDP method produces $p$ different plots (one for each feature) and the time and human effort required for the analysis could easily become unsustainable as the dimensionality of data grows. Indeed, the complexity of the visualization (in terms of the number of displayed plots) may collide with the limited human ability to elaborate simultaneous information, that has been proven to be restricted to a set of seven univariate simultaneous stimuli \cite{memory}.

\subsection{Local interpretability for regression tasks} \label{sec:acme_local}
\chiara{When the scope of the analysis is the interpretation of individual predictions, we set the baseline vector $\mathbf{x^b}$ equal to the specific data point to be explained $\mathbf{x}^*$, instead of setting $\mathbf{x^b}=\mathbf{\bar{x}}$ as in the scenario of global interpretability.}

The procedure to get importance scores is similar to that outlined in \Cref{sec:acme_global}, but in the local case it only serves to order features in the displayed plot, which is meant to convey a different kind of information compared to the global case. The visualization strategy adopted for local interpretability is described in the next paragraph.

\paragraph{Results visualization}
In \ourmethod, we produce a local interpretability visualization that is reminiscent of its counterpart used for global interpretability, but at the same time we introduce fundamentals modifications aimed at making its interpretation more intuitive and its usage more actionable. Specifically, in the local case we do not display \scoresname but the actual predictions associated with the perturbed data points (based on the selected $Q$ quantiles).

Such a design choice allows for immediate understanding since the underlying \emph{what-if} approach is well-aligned with human tendency to reason in counterfactual terms. A dashed line is placed in correspondence of the prediction associated with the original observation \chiara{$\mathbf{x}^*$}, so that it is clear which variables are pushing to increase (or decrease) the prediction. The advantage of \ourmethod visualization for local interpretability, if compared to the SHAP force plot, is that also information about feature distributions is implicitly provided by means of quantile levels. In particular, the knowledge of the quantile corresponding to each feature value in the original observation 
makes it possible to understand how much the value of a specific feature can be reduced or increased, and how the corresponding prediction will be affected.
\chiara{Figure \ref{fig:acme_local} shows an example of local interpretability provided by \ourmethod, which is further detailed in \Cref{sec:real_data}}.

\subsection{\ourmethod for classification}\label{sec:classification}
In \Cref{sec:acme_global}  we described \ourmethod for regression tasks. As for classification, by considering the problem of estimating the probability assigned to each class instead of the problem of assigning a label, we can easily resort to a regression task. Thus, the procedure detailed above is still valid. 
In particular, for global feature importance, we consider, for each class $l$ and each feature $j$, the \scoresname computed by evaluating the changes in the predicted probability of class $l$ under perturbations of feature $j$ (as usual, perturbations are obtained by replacing the original feature value with the selected $Q$ quantile). The only drawback of this approach is that we should produce as many plots as the number of distinct classes in the dataset, as shown in \Cref{fig:GlassSingleClasses}. 
\david{The barplot simplified visualization can circumvent the problem, evolving to a stacked barplot}: each feature is assigned a bar partitioned in as many blocks as the number of distinct classes and the length of each block is proportional to the \scorename associated with the specific class. \Cref{fig:Glass} depicts a comparison between this plot and the on provided by SHAP.
Similar considerations hold for the local interpretability. For each class, \ourmethod displays the predicted probabilities under perturbations with respect to the original observation feature values (see, for example, \Cref{fig:acme_local_class}).
An application of \ourmethod to a classification task is described in Section \ref{sec:real_data_classification}.


\section{Experimental results}\label{sec:exp_results}
In this section, we describe the experiments carried out to assess the effectiveness of \ourmethod. \chiara{It is worthwhile to notice that interpretability problems are non-supervised tasks, meaning that, in general, there is no ground-truth available to assess feature ranking and importance scores. To overcome this issue, in \Cref{sec:syn_data}, we resort to synthetic dataset generation that gives us the chance to have a ground truth, so it enables a more objective evaluation. 
Then, to assess the efficiency of \ourmethod, we compare it with KernelSHAP, a model-agnostic variant of SHAP, by considering both computational time and explanation quality. For the sake of reproducibility, experiments were conducted on well-known, publicly available ML datasets that are paradigmatic for regression (\Cref{sec:real_data}) and classification tasks (\Cref{sec:real_data_classification}).}

\subsection{Synthetic datasets} \label{sec:syn_data}

We created two synthetic datasets with controlled characteristics in order to assess the robustness of \ourmethod w.r.t. such conditions. Specifically, we study a regression task and consider the data points generated by the following linear model
\begin{equation} \label{eq:true_model}
    \mathbf{y} = \mathbf{X} \boldsymbol{\beta} + \boldsymbol{\epsilon}
\end{equation}
where $\mathbf{y} = [y_1, \dots, y_N]^T \in \mathbb{R}^N$ is the vector of responses, $\mathbf{X} \in \mathbb{R}^{N \times p}$ is the design matrix, $\boldsymbol{\beta} \in \mathbb{R}^p$ the vector of model's coefficients and $\boldsymbol{\epsilon} \sim \mathcal{N}_p(\mathbf{0_p}, \mathbf{\Sigma})$. 
The vector of model coefficients $\boldsymbol{\beta} = [\beta_1, \dots, \beta_p]^T$ is designed so that some variables have a real effect ($\beta_j \neq 0$), while others are not important ($\beta_j = 0$). 
In both the experiments detailed below, we generate $N=200$ observations according to \Cref{eq:true_model}, then we fit a linear model to the obtained data. \david{What we expect is that both KernelSHAP (with default parameters) and \ourmethod (with $Q=50$) succeed to identify the \chiara{correct feature importance ranking. Notice that this is known in advance, since we can compute it based on vector $\boldsymbol{\beta}$}. To certificate the quality of the obtained results, we use Normalized Discounted Cumulative Gain \cite{ndcg_wang}. NDCG sums the true scores ($\boldsymbol{\beta}$) ranked in the order induced by the predicted scores (the \ourmethod features importance) after applying a logarithmic discount. Then it divides by the best possible score (ideal DCG, obtained for a perfect ranking) to obtain a normalized score between 0 and 1. }

\subsubsection{Experiment \#1: variables with the same scale}
In the first experiment, we analyze the simplest scenario, \chiara{ where variables all share the same scale}. In particular, the dataset is obtained by setting the following parameters for the linear model in \Cref{eq:true_model} and \chiara{generating $N=200$ observations}:
\begin{align*}
    \boldsymbol{\mu} &= [10,10,10,10,10,10,10,10]^T \\ 
    \boldsymbol{\Sigma} &= 10 * \text{I}_8 \\
    \mathbf{X}_i &\sim \mathcal{N}_8( \boldsymbol{\mu}, \boldsymbol{\Sigma})\\
    \boldsymbol{\beta} & = [10,20,-10,0.3,1,0,0,-0.5]^T \\
    \boldsymbol{\epsilon}_i & \sim \mathcal{N}_8(\boldsymbol{0}_8, \boldsymbol{\Sigma}).
\end{align*}

\begin{figure}[t]
\centering
    \subfloat[AcME]{\includegraphics[width=0.48\textwidth]{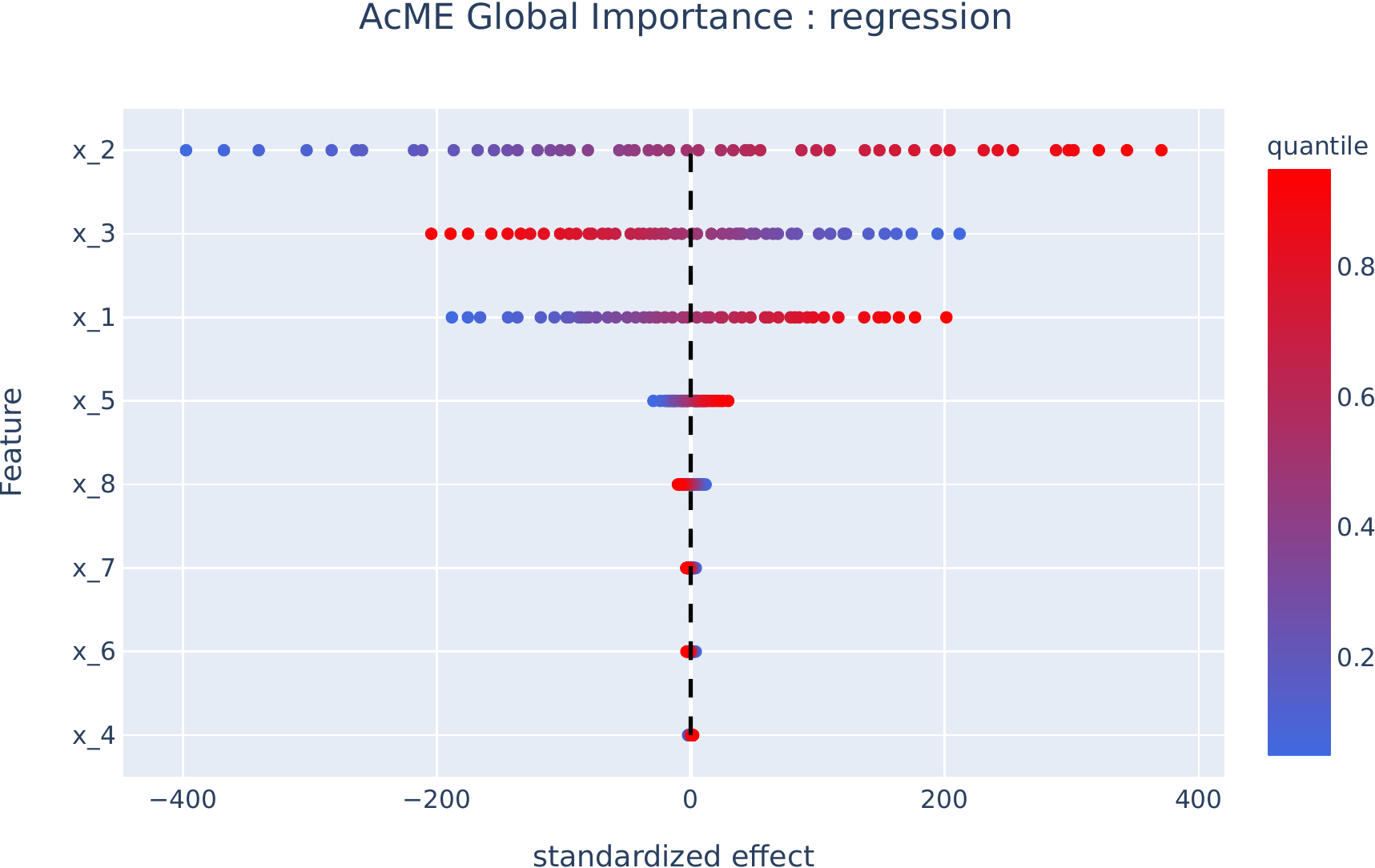} }
    \hspace{0.2cm}
    \subfloat[KernelSHAP]{\includegraphics[width=0.48\textwidth]{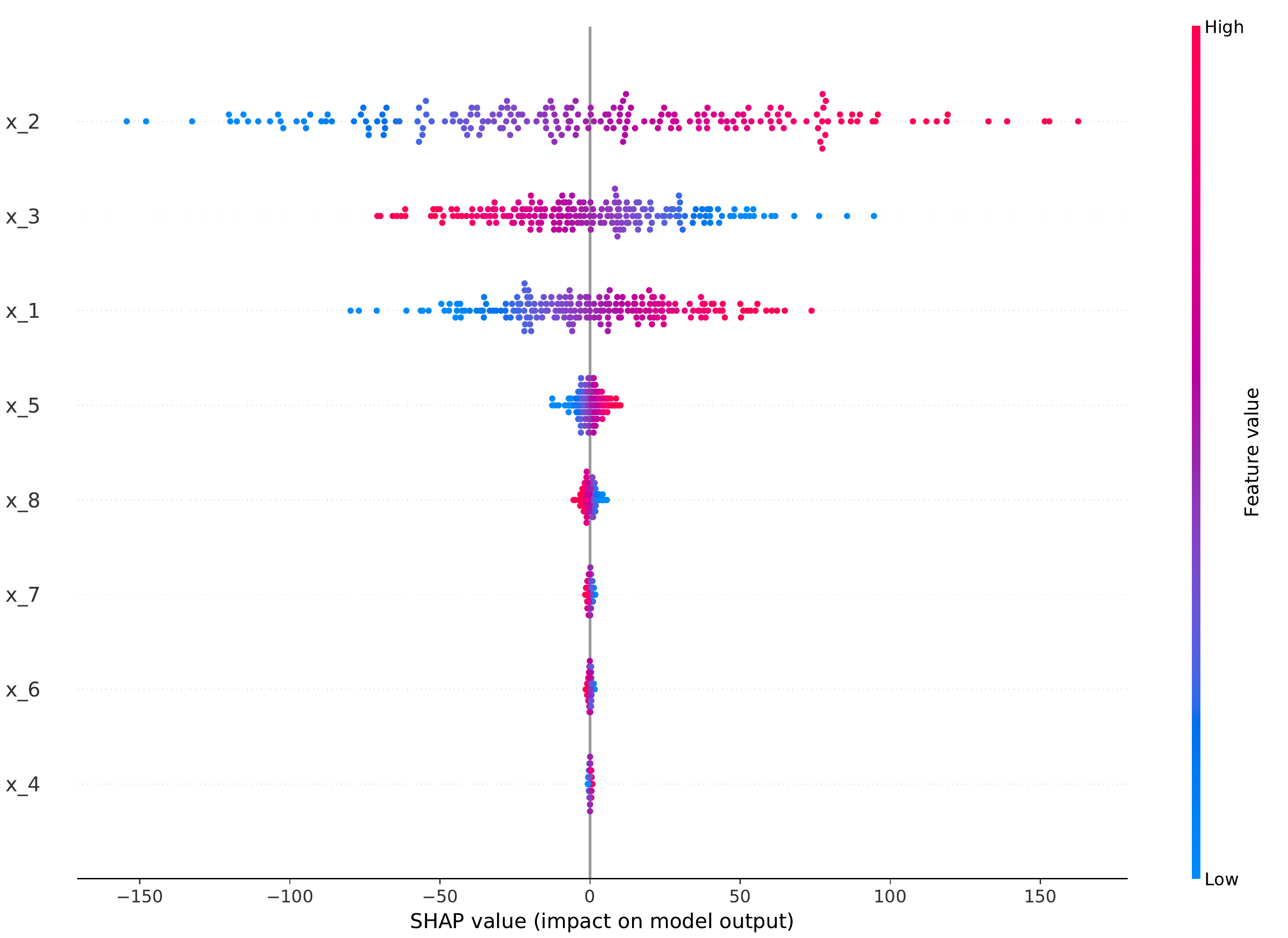}}
    \caption{[Experiment \#1] Result comparison of \ourmethod and KernelSHAP. \ourmethod elapsed time: 0.9864 s; KernelSHAP elapsed time: 242.0958 s.}
    \label{fig:model1}
\end{figure}

Notice that such a design choice for the vector of model coefficients $\boldsymbol{\beta}$ leads to specific considerations with regard to features ranking: $x_1$, $x_2$ and $x_3$ are the variables that have the largest impact on the response $y$ ($x_2$ being the most relevant), while $x_4$, $x_6$, $x_7$ and $x_8$ have negligible or no effect. Moreover, $x_3$ and $x_8$ affect the model response in the opposite direction w.r.t. the other variables, resulting in a negative input-output correlation.
As can be seen in \Cref{fig:model1}, both \ourmethod and KernelSHAP can correctly identify which features are actually important for the fitted model. Also, both methods attribute a negative effect (on the predicted output) to the highest quantiles for features $x_3$ and $x_8$, reflecting the fact that $\beta_3<0$ and $\beta_8<0$. Features $x_4$, $x_6$ and $x_7$ exhibit the lowest feature importance scores, in accordance with the true model specifications. The main difference is in the computational time: \ourmethod requires less than a second, while KernelSHAP requires about 4 minutes in the tested hardware\footnote{As a reference, the experiment reported in this work were achieved on a Macbook with CPU i5 2,9 GHz, SSD 256 GB and 8GB RAM.}. \david{ The NDCG calculated on the feature ranking produced by \ourmethod is reported in Table \ref{tab:ndcg_metric}. \chiara{It is very close to 1, showing that the feature ranking returned by \ourmethod is always similar to the expected one, as the model was able to infer the actual underlying data generation process, and AcME could explain how the model maps input to outputs. }}

\begin{table}[]
\centering
\footnotesize
\begin{tabular}{l|l|l|l|l}
\hline
Experiment & True ranking & AcME ranking & NDCG  \\
\hline
1 & $[x_2,x_3,x_1,x_5,x_8,x_4,x_6,x_7]$ & $[x_2,x_3,x_1,x_5,x_8,x_7,x_6,x_4]$ & 0.9998 \\
2 & $[x_1,x_2,x_3,x_5,x_8,x_4,x_6,x_7]$ & $[x_1,x_2,x_3,x_5,x_8,x_7,x_4,x_6]$ & 0.9998 \\
\hline
\end{tabular}
\caption{NDGC score for Experiment 1 and 2. It is strictly near to 1 in both of them, showing how the ranking produced by AcME is consistent with the real one. }
\label{tab:ndcg_metric}
\end{table}

\subsubsection{Experiment \#2: variables with different scale}

In this experiment, we aim at studying the effect of exploiting variables with different scales. 
The data-generating model is instantiated as previously, with the exception of $\boldsymbol{\mu}$ and $\boldsymbol{\Sigma}$, now set as follows:
\begin{align*}
    \boldsymbol{\mu} & = [100, 10, 10, 10, 100, 10, 10, 100]^T \\
    \boldsymbol{\Sigma} & = diag(100,10,10,10,100,10,10,100).  
\end{align*}

\begin{figure}[t]
\centering
    \subfloat[AcME]{ \includegraphics[width=0.48\textwidth]{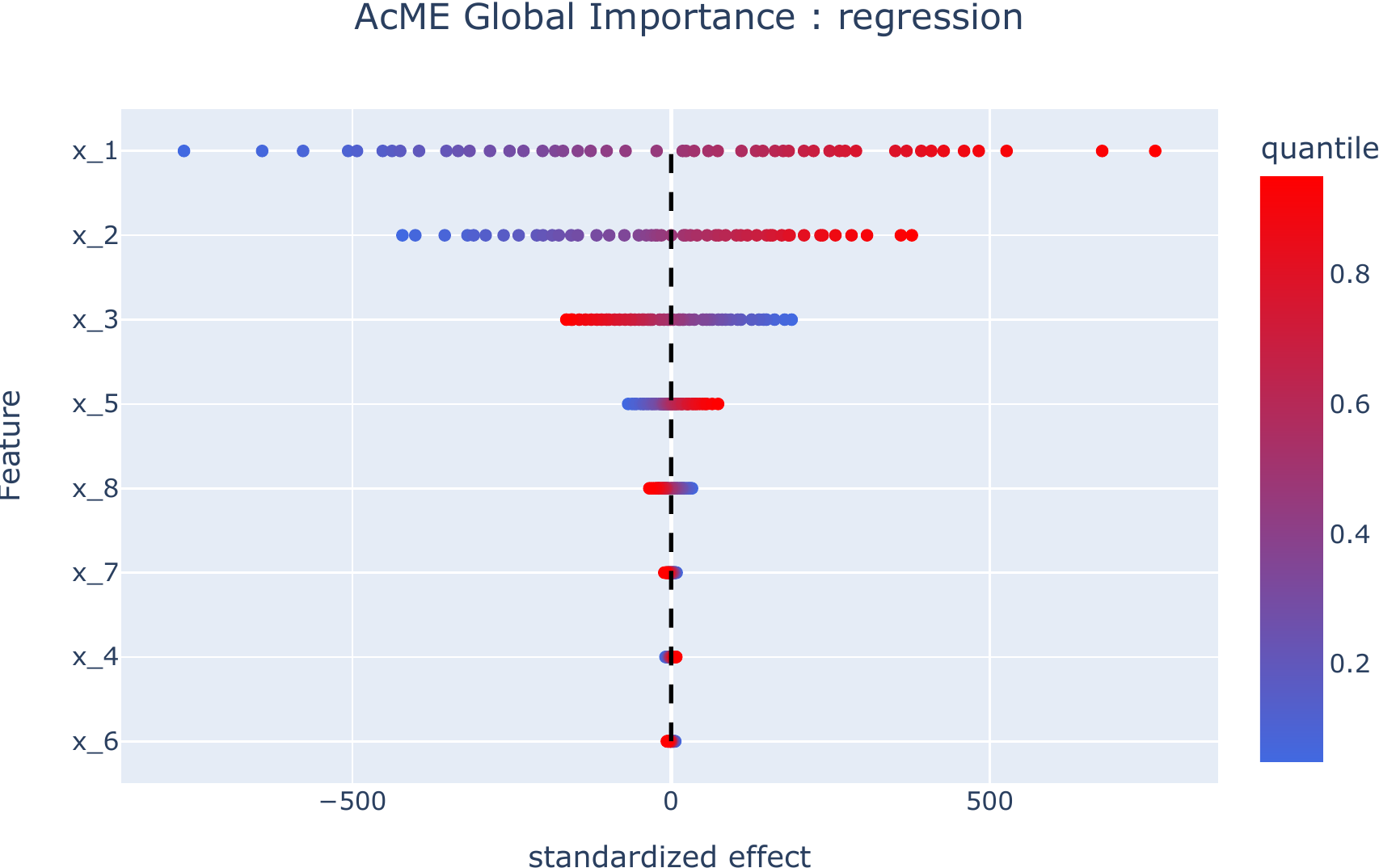} }
    \hspace{0.2cm}
    \subfloat[KernelSHAP]{\includegraphics[width=0.48\textwidth]{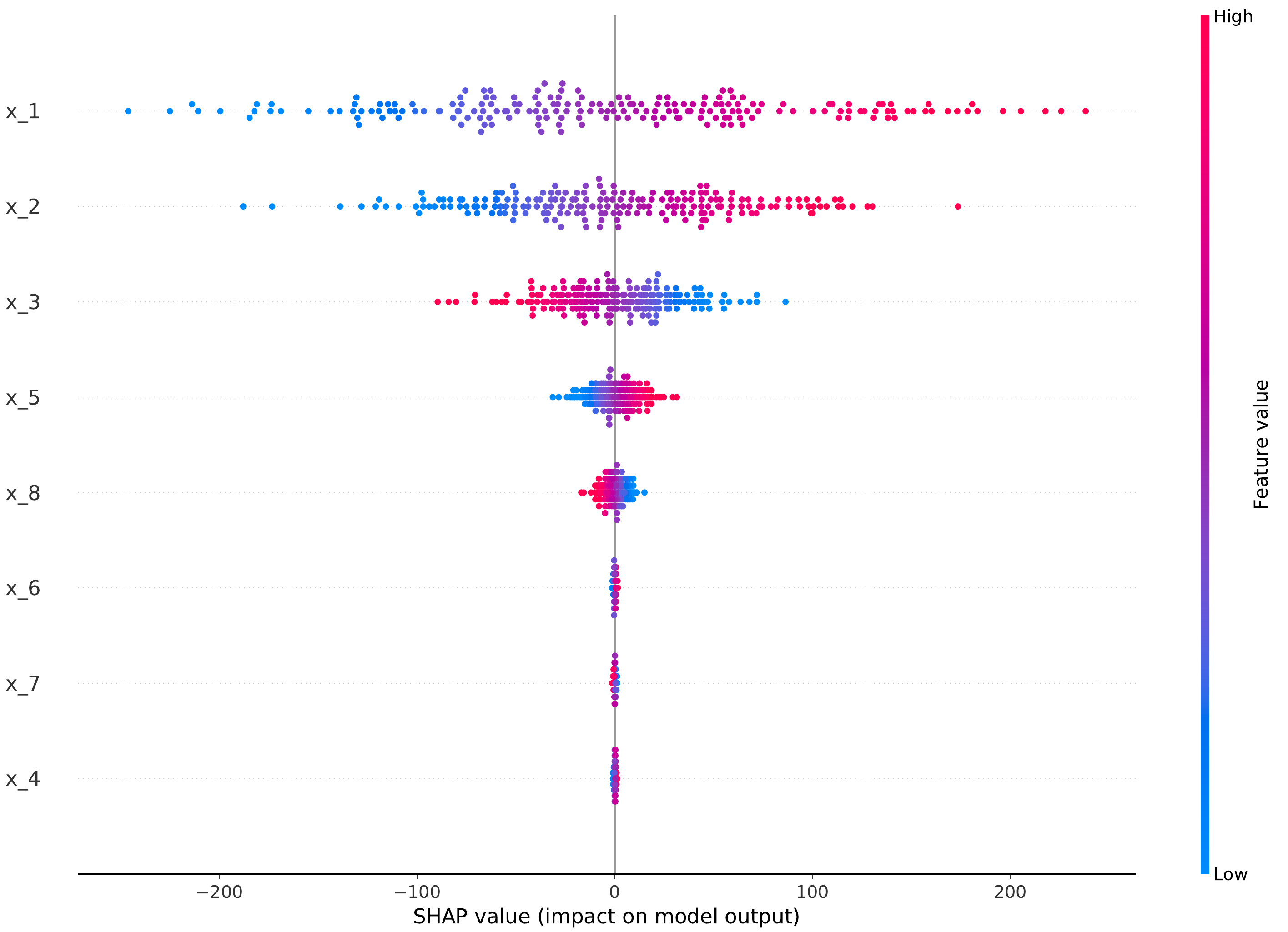}}
    \caption{ [Experiment \#2] Result comparison of \ourmethod and KernelSHAP. \ourmethod elapsed time:  0.9450; KernelSHAP elapsed time: 242.1215. }
    \label{fig:model2}
\end{figure}

What we expect is that the impact of $x_1$ will be very higher than before, surpassing $x_2$ and $x_3$. Moreover, we expect that the importance of $x_8$ with $\beta_8 = -0.5$ will be comparable more with the importance of $x_4,x_6,x_7$, even if the variable receives a great increase both in terms of position and variance. This expected behaviour is reasonable even if the user performs data normalization in the preprocessing step, in fact larger scale variables effects would be captured by an associated larger $\beta$, and as demonstrated in the previous experiment, \ourmethod correctly describes that situation.  
In \Cref{fig:model2}, we could see that the proposed approach behaves as expected, supporting the quality of the results. As for experiment \#1, the most obvious difference between the two methods is the elapsed time: approximately 1 second for \ourmethod and 4 minutes for KernelSHAP. \david{As happened in the previous experiment, the ranked list metric is near to the upper limit (Table \ref{tab:ndcg_metric}).}

\subsection{Experiments on a regression task: Boston Housing dataset} \label{sec:real_data}
\chiara{The goal of this experiment is to show the results obtained by \ourmethod on a regression task, and to compare them with those achieved by KernelSHAP. To this aim, we consider a well-known publicly available dataset,  \href{https://archive.ics.uci.edu/ml/machine-learning-databases/housing/}{UCI Boston Housing dataset}\footnote{https://archive.ics.uci.edu/ml/machine-learning-databases/housing}\label{boston}. The Boston Housing dataset is composed of 506 observations with 14 variables describing houses in the area of Boston Mass. The task is predicting the median value of owner-occupied homes. 
\\
KernelSHAP and \ourmethod are compared in terms of both evaluated feature importance and computational time.}
\subsubsection{Efficiency comparison between KernelSHAP and \ourmethod}
{As seen in \Cref{sec:shap}, what makes \textit{KernelSHAP} so computationally burdensome is:
\begin{enumerate}
    \item the high number of linear models that are estimated;
    \item the computational complexity of the matrix inversion, that increases with the number of coalitions.
\end{enumerate}
To overcome the first problematic, the authors of \cite{shap_user_guide} suggest to reduce the number of rows in the dataset, sampling $\widetilde{N}$ rows from the original ones. 
For the second issue, the only option is to reduce the number of coalitions $K$.
In this section, we will explore these solutions, evaluating the quality of results with respect to changes in parameters $\widetilde{N}$ and $K$ in terms of their similarity to results obtained with the complete original dataset and the suggested number of coalitions. \chiara{Our experiments suggest that using dataset sampling and coalitions reduction to lessen the computational time required by KernelSHAP may translate into unreliable results. Moreover, elapsed time for KernelSHAP is still much higher than that required by \ourmethod.}

\begin{table}[]
\centering
\footnotesize
\begin{tabular}{l|l|r}
\hline
& Number of samples & Elapsed Time (in seconds) \\
\hline
\ourmethod & complete & 0.36 \\
KernelSHAP & 5   &  357.23 \\
KernelSHAP & 10  &  425.61 \\
KernelSHAP & 20  &  875.85 \\
KernelSHAP & 100 &  1855.65 \\
\hline
\end{tabular}
\caption{[Boston Housing Dataset] Elapsed time for KernelSHAP with different dataset sampled. }
\label{tab:shap_sampled}
\end{table}

\begin{figure}
\centering
   \subfloat[SHAP with sampling $g = 5$ rows]{\includegraphics[width=0.5\textwidth]{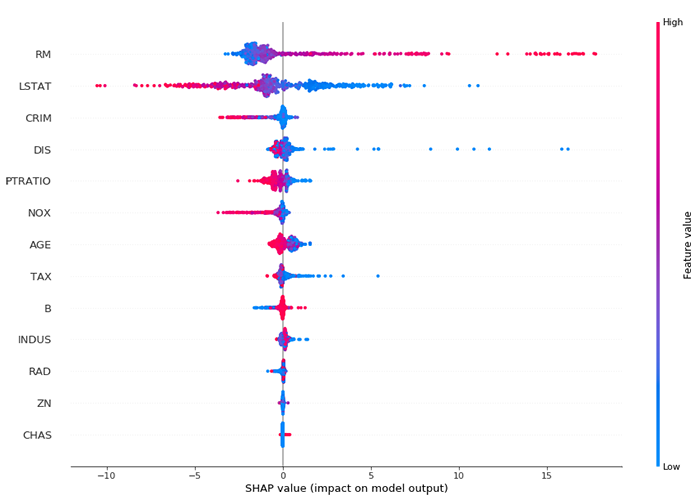}}
	\subfloat[SHAP with sampling $g = 10$ rows]{\includegraphics[width=0.5\textwidth]{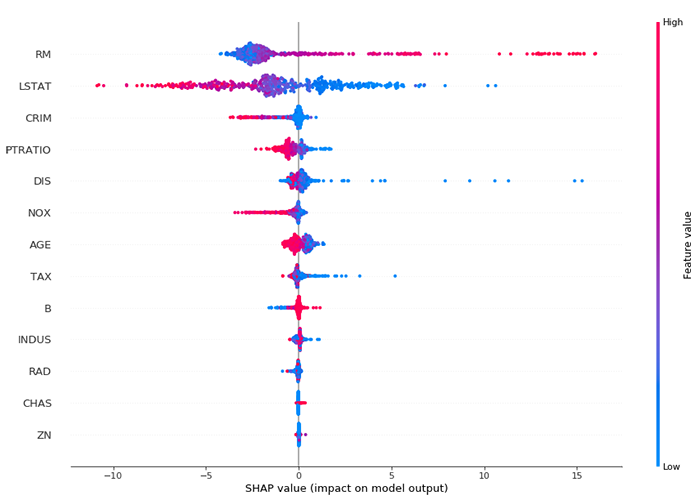}}\\
	\subfloat[SHAP with sampling $g = 20$ rows]{\includegraphics[width=0.5\textwidth]{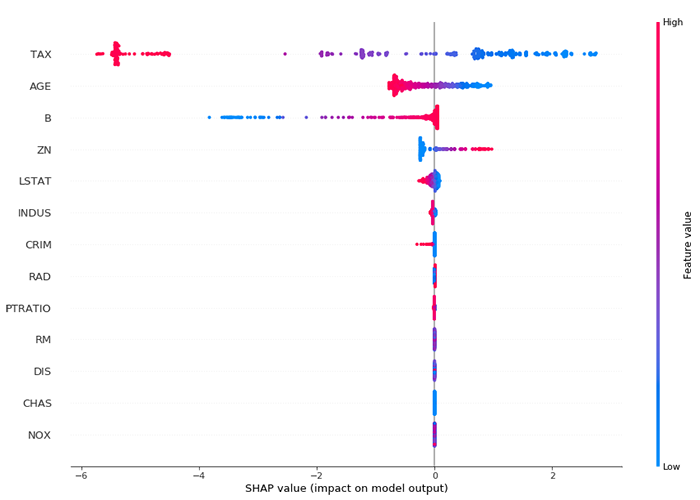}}
	\subfloat[SHAP with sampling $g = 100$ rows]{\includegraphics[width=0.5\textwidth]{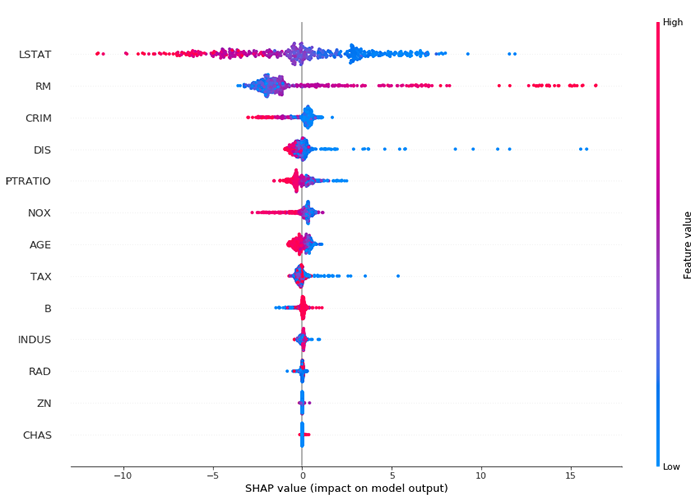}}\\
	\subfloat[SHAP calculated on the full dataset]{\includegraphics[width=0.7\textwidth]{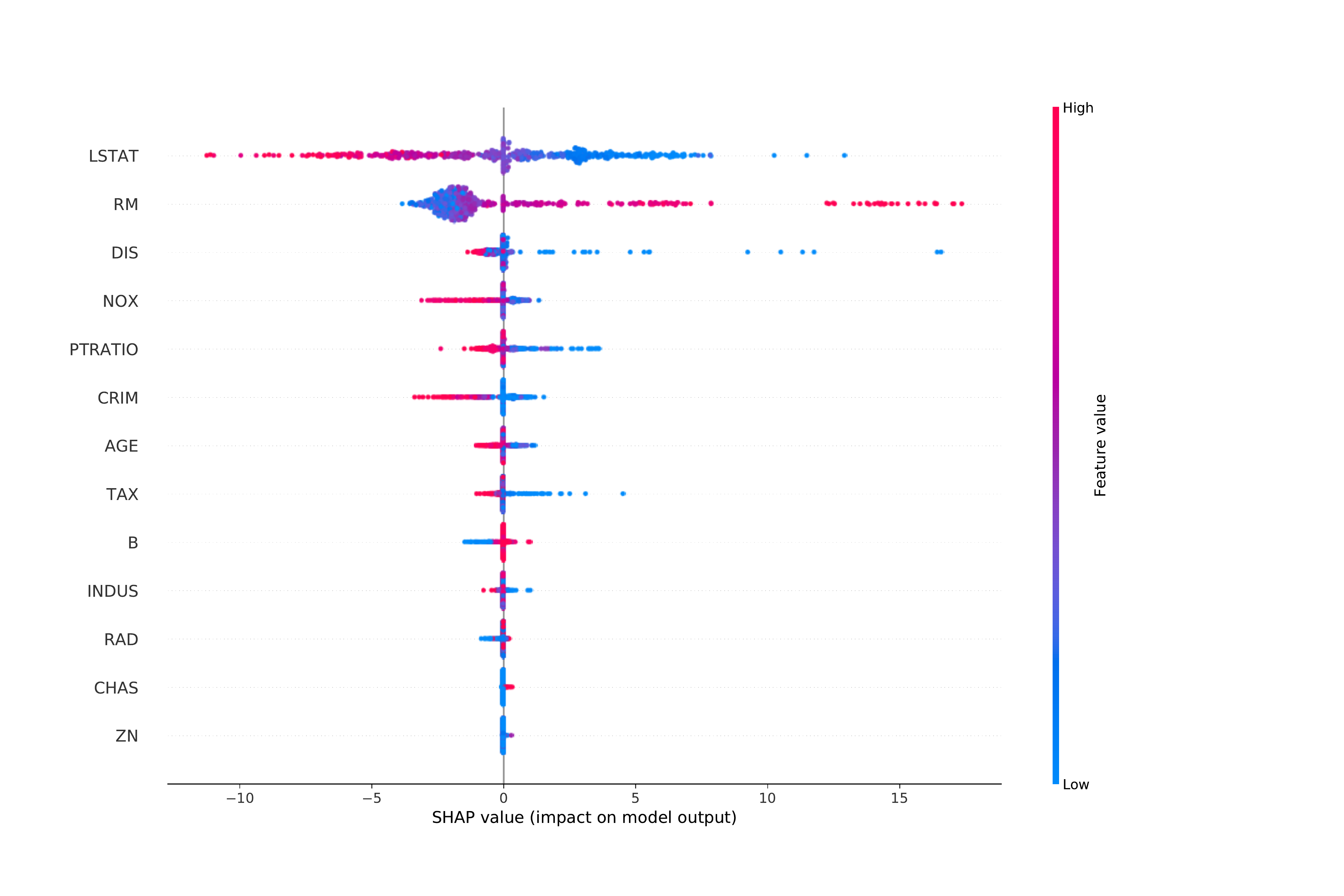}}
    \caption{[Boston Housing Dataset] KernelSHAP summary plot for the Random Forest model with 100 tree, using a sample in order to reduce the number of rows of the input dataset. As clear from the plots, the results are not stable in term of feature behaviour and importance compared to the results obtained using the full dataset. }
    \label{fig:shap_sample}
\end{figure}

\begin{table}[]
\centering
\scriptsize
\begin{tabular}{M{1.2cm}|M{9.5cm}|M{1.2cm}|M{1.5cm}}
\hline
$\tilde{N}$ ($\#$ samples) & Ranking  & Kendall Tau (full) & Kendall Tau (top 5) \\
\hline
5 & $\text{RM,LSTAT,CRIM,DIS,PTRATIO,NOX,AGE,TAX,B,INDUS,RAD,ZN,CHAS}$ & 0.231 & 0.2 \\
10 & $\text{RM,LSTAT,CRIM,PTRATIO,DIS,NOX,AGE,TAX,B,INDUS,RAD,CHAS,ZN }$ & 0.692 & 0.0 \\
20 & $\text{TAX,AGE,B,ZN,LSTAT,INDUS,CRIM,RAD,PTRATIO,RM,DIS,CHAS,NOX}$ & 0.103  & -0.2 \\
100 & $\text{LSTAT,RM,CRIM,DIS,PTRATIO,NOX,AGE,TAX,B,INDUS,RAD,ZN,CHAS }$ & 0.359 & 0.8\\
complete & $\text{LSTAT,RM,DIS,NOX,PTRATIO,CRIM,AGE,TAX,B,INDUS,RAD,CHAS,ZN}$ & 1 & 1\\
\hline
\end{tabular}
\caption{[Boston Housing Dataset] Kendall Tau score obtained with the comparison of KernelSHAP ranking list with the full dataset with KernelSHAP rankings obtained using only 5,10,20,100 samples from the dataset.}
\label{tab:SHAP_KT_samples}
\end{table}

\paragraph{Reducing KernelSHAP computational burden by sampling observations}
We compute KernelSHAP by sampling 5, 10, 20, and 100 rows at random from the original dataset, \david{keeping the default number of coalitions}. In \Cref{tab:shap_sampled}, we show the computational time of \ourmethod run on full dataset compared to that required by KernelSHAP when considering the sampled rows only. It is clear that \ourmethod is always much faster. 
In addition, even though the computational time required by the sampled KernelSHAP is much lower than the original version when using lower values of $\widetilde{N}$, the output is not reliable, as depicted in \Cref{fig:shap_sample}. For instance, when we use $\widetilde{N}=20$, not only the feature order based on computed importance is different, but also the model behaviour is not correctly detected: for example the Shapley values of \texttt{TAX, RM, AGE} are completely different from all the other cases. The instability in the results is caused by the sampling procedure on the original dataset, that does not extract \chiara{rows that are truly representative of the dataset}. 

\david{To measure the uncertainty in the ranking \chiara{induced by dataset sampling}, we use Kendall’s Tau metric, a measure of the correspondence between rankings. Being a rank correlation coefficient, it will be high for two lists with similar ranking (with the maximum in 1 when perfectly identical), whereas it will be low for two lists sorted very differently (with the minimum in -1 when all position pairs are discordant). In Table \ref{tab:SHAP_KT_samples} there are reported Kendall's Tau values for each ranking generated by KernelSHAP with different number of samples, compared with KernelSHAP ranking obtained with the full dataset, both for entire lists and for top five elements of each list. While the correlation values for the full lists are strictly positive, the same doesn't happen considering the top five lists. In addition, it should be noted that there isn't a monotonous increasing trend. In fact, worst results are obtained with $g=20$. Both KernelSHAP and \ourmethod use perturbations of the input dataset to obtain model explanation. \chiara{However, as for KernelSHAP, analysing the similarity among the rankings corresponding to different sampling strategies suggests that sampling may have a detrimental effect on the quality of explanations, especially when features distribution is not trivial}. Instead, \ourmethod relies on quantiles: the estimation of these statistics is robust to outliers, while the sampling strategy implemented by SHAP seems to be not, in general.}

With $\widetilde{N}=100$ the results appear similar to those on the entire dataset, however the elapsed time is already significantly high.

\chiara{In this experiment, we extracted rows at random. Alternatively, smarter sampling techniques could be used: for instance, a suggestion could be the usage of clustering to detect groups of observations to sample from. However, the use of clustering, and more generally of others advanced sampling techniques, requires the tuning of hyper parameters, like the number of clusters. The correct choice of the parameters is relevant because it could heavily affect the quality of the results, and it needs \textit{ad hoc} analysis. For instance, it is not possible to determine a rule-of-thumb to suggest the correct number of lines to sample, because this parameter is strongly linked to the intrinsic complexity of the \chiara{dataset}.}

\begin{table}
\centering
\footnotesize
\begin{tabular}{l|l|r}
\hline
& Number of coalitions K  & Elapsed Time (in seconds) \\
\hline
\ourmethod & complete &    0.36 \\
KernelSHAP & 10  &    92.40 \\
KernelSHAP & 25  &   221.61 \\
KernelSHAP & 50  &   327.65 \\
KernelSHAP & 100 &   506.31 \\
\hline
\end{tabular}
\caption{[Boston Housing Dataset] Elapsed time for KernelSHAP with different number of coalitions used in the estimation of the Shapley values. }
\label{tab:shap_values_sampled}
\end{table}

\begin{table}[]
\centering
\scriptsize
\begin{tabular}{M{1.3cm}|M{9.5cm}|M{1.2cm}|M{1.5cm}}
\hline
K ($\#$ coalitions) & Ranking & Kendall Tau (full) & Kendall Tau (top 5) \\
\hline
10 & $\text{LSTAT,RM,ZN,CRIM,INDUS,CHAS,NOX,DIS,AGE,PTRATIO,TAX,RAD,B}$ & -0.05 & -0.2 \\
25 & $\text{LSTAT,RM,DIS,CRIM,NOX,PTRATIO,AGE,B,TAX,INDUS,ZN,CHAS,RAD}$ & 0.231 & 0.06 \\
50 & $\text{LSTAT,RM,CRIM,DIS,PTRATIO,NOX,AGE,TAX,B,INDUS,RAD,ZN,CHAS}$ & 0.359 & 0.8 \\
100 & $\text{LSTAT,RM,DIS,CRIM,PTRATIO,NOX,AGE,TAX,B,INDUS,RAD,CHAS,ZN}$ & 0.821 & 0.6\\
default & $\text{LSTAT,RM,DIS,NOX,PTRATIO,CRIM,AGE,TAX,B,INDUS,RAD,CHAS,ZN}$ & 1 & 1\\
\hline
\end{tabular}
\caption{[Boston Housing Dataset] Kendall Tau score obtained with the comparison of KernelSHAP ranking list with the default number of coalitions and KernelSHAP rankings obtained using only 10,25,50,100 coalitions.}
\label{tab:SHAP_KT_coalitions}
\end{table}

\paragraph{Reducing KernelSHAP computational burden by reducing the number of coalitions}
In \Cref{tab:shap_values_sampled}, we report how the computational time required by the KernelSHAP procedure varies with the number of sampled coalitions, by considering 10, 20, 50 and 100 coalitions, \david{using the entire dataset}. As previously stated, reducing the number of sampled coalitions \chiara{can} really speed up the computation, but the results are once again unstable. \chiara{Besides, results exhibit higher variance when the number of considered coalitions is lower.} \david{In Table \ref{tab:SHAP_KT_coalitions} we reported the Kendall's Tau measure calculated on the ranked lists obtained reducing the number of coalitions and the ranked list obtained using the default number of coalitions}. This is strictly related to the approximation used in the computation of the Shapley values, that uses a lower number of permutations to estimate the true values. Again, feature distribution complexity is an important aspect: with simple distribution a lower number of permutations could already lead to decent results, but with more complex distributions this is not always true.

\subsubsection{Feature ranking evaluation and comparison between KernelSHAP and \ourmethod}\label{sec:ranking_evaluation_and_comparison}
\chiara{To study the importance scores assigned to each feature by \ourmethod and KernelSHAP (calculated on the full dataset with default parameters), we fit three different types of models: \textit{Linear Regression}, \textit{Random Forest} and \textit{CatBoost Regression}.  
To evaluate the generated feature ranking, we estimated other two different models for each model type. We trained the former using only the five features with the highest importance according to AcME. Instead, the latter considers all other features only. We expect that the first model will have a Mean Squared Error (MSE) closer to the value obtained using all the features, whereas the MSE for the second model will be much higher. There is no ground truth that we can use for a fair comparison. However, this procedure allows us to assess the interpretability performance quantitatively.
As detailed in Table \ref{tab:MSE}, experimental results reflect the expectations, confirming that \ourmethod correctly identified the features that are most relevant for the model.}

\begin{table}[]
\centering
\footnotesize
\begin{tabular}{l|r|r|r}
\hline
Model & MSE & Top 5 features & All features except top 5 \\
\hline
Linear Regression & 21.89 & 26.01 & 55.15 \\
Random Forest & 1.54 & 2.40 & 4.42 \\
CatBoost & 0.52 & 0.86 & 3.80 \\
\hline
\end{tabular}
\caption{[Boston Housing Dataset] Mean Square Error for the 3 model used for \ourmethod ranking evaluation. The best model here is the CatBoost, followed by the Random Forest.  }
\label{tab:MSE}
\end{table}

\begin{table}
\centering
\footnotesize
\begin{tabular}{l|l|r}
\hline
Method & Model & Elapsed Time (in seconds) \\
\hline
KernelSHAP & Linear Regression  &   3651.8556 \\
KernelSHAP & Random Forest  &   5639.9273 \\
KernelSHAP & CatBoost Regression  & 4578.0989 \\
KernelSHAP & SVR & 12094.5632 \\
\ourmethod & Linear Regression & 0.2610 \\
\ourmethod & Random Forest & 0.4107 \\
\ourmethod & CatBoost Regression & 0.9764 \\
\ourmethod & SVR &  1.5584 \\
TreeSHAP & Random Forest & 4.7886 \\
TreeSHAP & CatBoost Regression  & 3.1525\\
\hline
\end{tabular}
\caption{[Boston Housing Dataset]  Computing time for the various tested models and the different methods of model explanation on the Boston dataset.}
\label{tab:ACME_time}
\end{table}

\Cref{tab:ACME_time} reports the computing time for \ourmethod and KernelSHAP, while \Cref{fig:LR} - \Cref{fig:CT} depict the results of the two methods. In \Cref{tab:ACME_time}, we report also the elapsed time using the TreeSHAP procedure, that is much faster than the KernelSHAP on the same model, but in this case is usable only with \textit{CatBoost} and \textit{Random Forest}.
In all the four situations, the results are similar both in terms of importance scores rank and feature-to-output map behavior, but the computing time is much lower for \ourmethod, that turns out to be extremely faster than each version of SHAP.
In particular:
\begin{itemize}
\item Linear Regression: both \ourmethod and KernelSHAP recognise \texttt{LSTAT} as the most important variable, and both the methods mostly agree on how the model uses the variables. The most important difference is in importance score for RAD: while KernelSHAP ranks it at the third place, our method put RAD at the sixth place. This happens for construction of the $\Delta_{j,k}$ (\Cref{eq:std_eff}), because we give greater importance to the variables with high impact on predictions range. In fact, instead of \texttt{RAD}, \ourmethod prefers \texttt{RM} that has the bigger variation among all variables.
\item Random Forest: here the two methods agree on the first two variables as regards importance. Then, KernelSHAP gives approximately the same importance to \texttt{DIS}, \texttt{NOX}, \texttt{PTRATIO}, \texttt{CRIM}, \texttt{AGE} and \texttt{TAX}, while \ourmethod gives higher importance to \texttt{DIS}. This happens because, for very low quantiles of \texttt{DIS}, the prediction is pushed to very high values and this does not occur for the other variables, as shown in \Cref{rf_acme} and \Cref{rf_shap}. 
\item CatBoost Regression: this is the best model in terms of MSE. In this case, as in the Random Forest, the only relevant difference between \ourmethod and KernelSHAP is the impact of \texttt{DIS}, that \ourmethod considers larger than what KernelSHAP does. Besides, the other importance scores and the model's behavior explanations given by the two methods are very similar.
\end{itemize}}

\begin{figure}
    \subfloat[n coalitions = 10]{\includegraphics[width=0.5\textwidth]{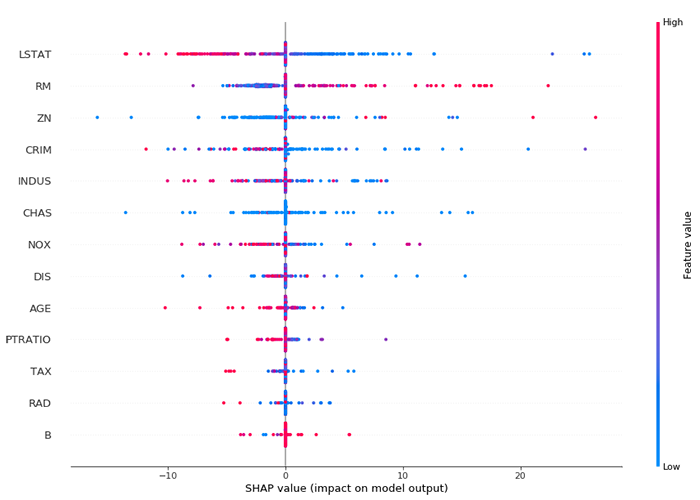}}
	\subfloat[n coalitions = 25]{\includegraphics[width=0.5\textwidth]{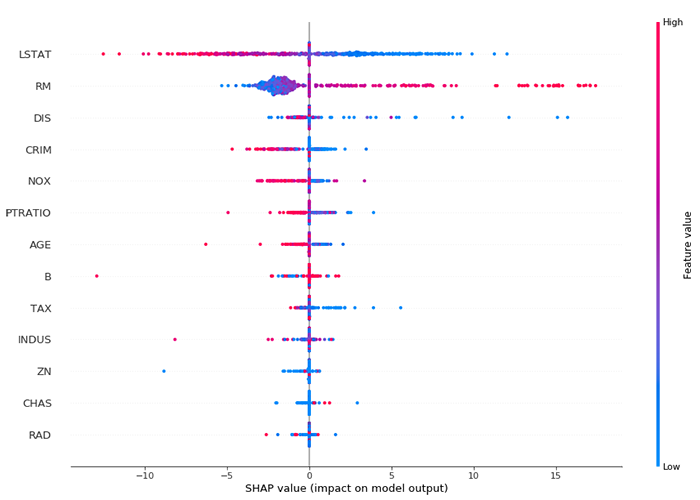}}\\
    \subfloat[n coalitions = 50]{\includegraphics[width=0.5\textwidth]{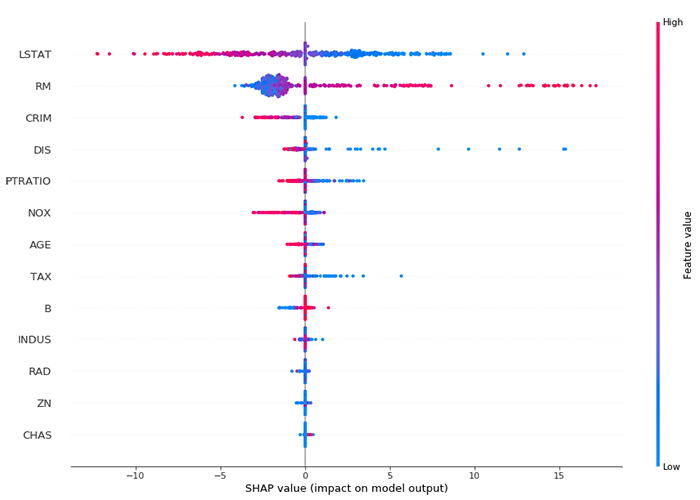}}
	\subfloat[n coalitions = 100]{\includegraphics[width=0.5\textwidth]{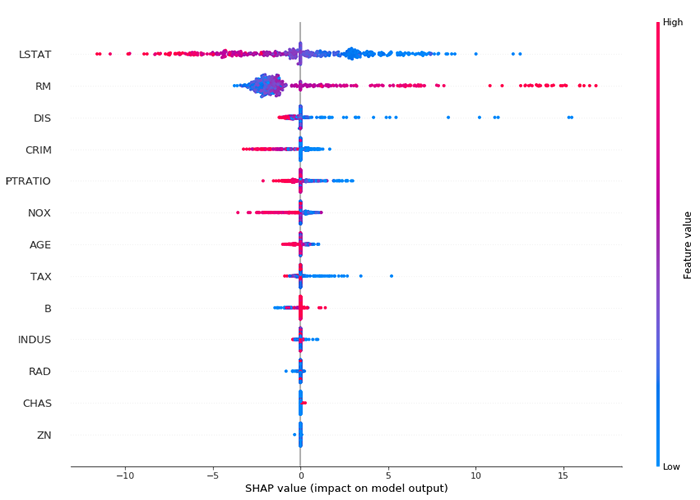}}
    \caption{[Boston Housing Dataset] KernelSHAP summary plot for the Random Forest model using a lower number of coalitions of the model to accelerate the procedure. }
        \label{fig:RF_coalitions}
\end{figure}

\begin{figure}
\centering
		\subfloat[\ourmethod on Linear Model]{\includegraphics[width=0.5\textwidth]{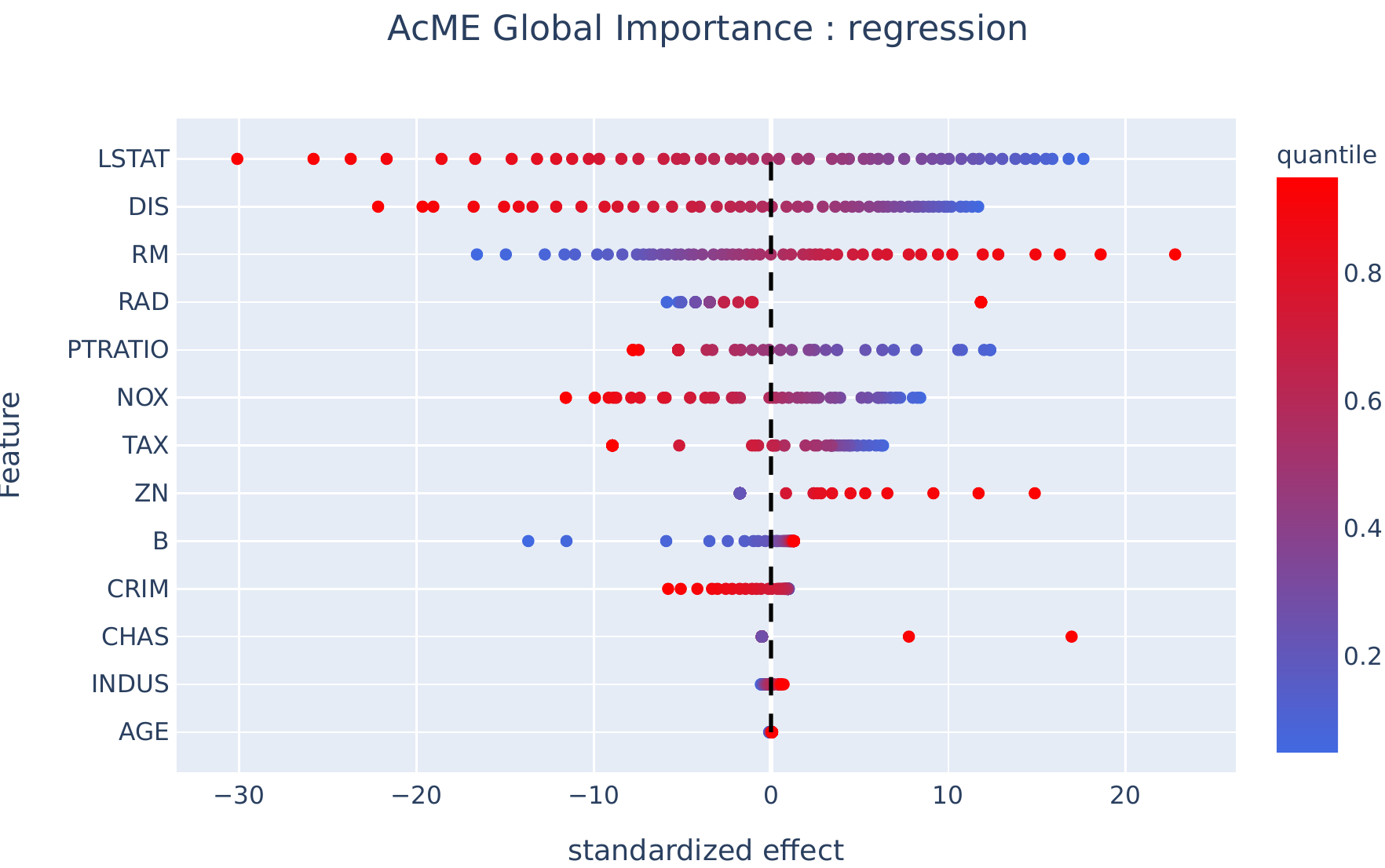}}
		\hspace{0.2cm}
		\subfloat[SHAP on Linear Model]{\includegraphics[width=0.48\textwidth]{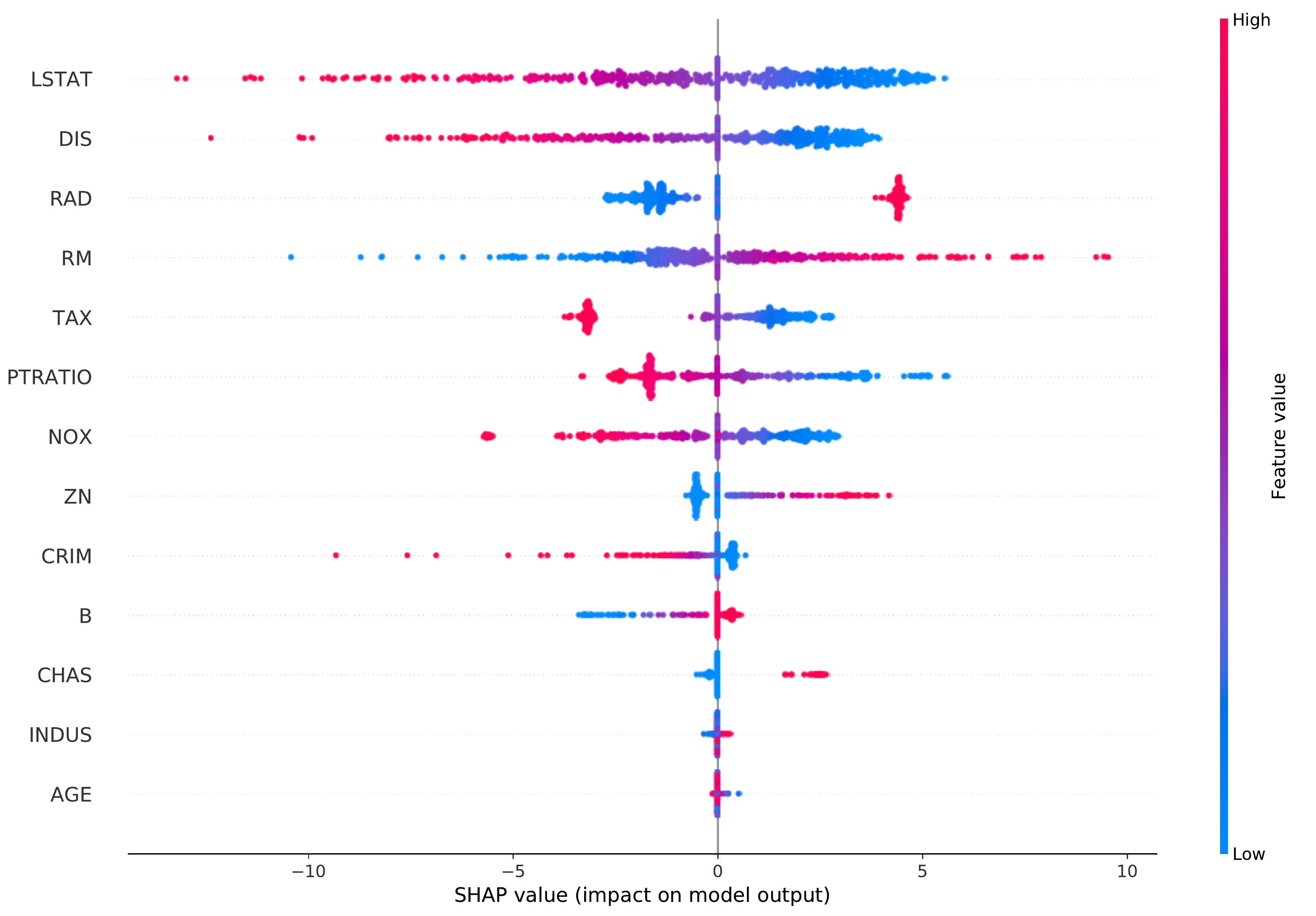}}\\
		\subfloat[\ourmethod Feature Importance]{\includegraphics[width=0.5\textwidth]{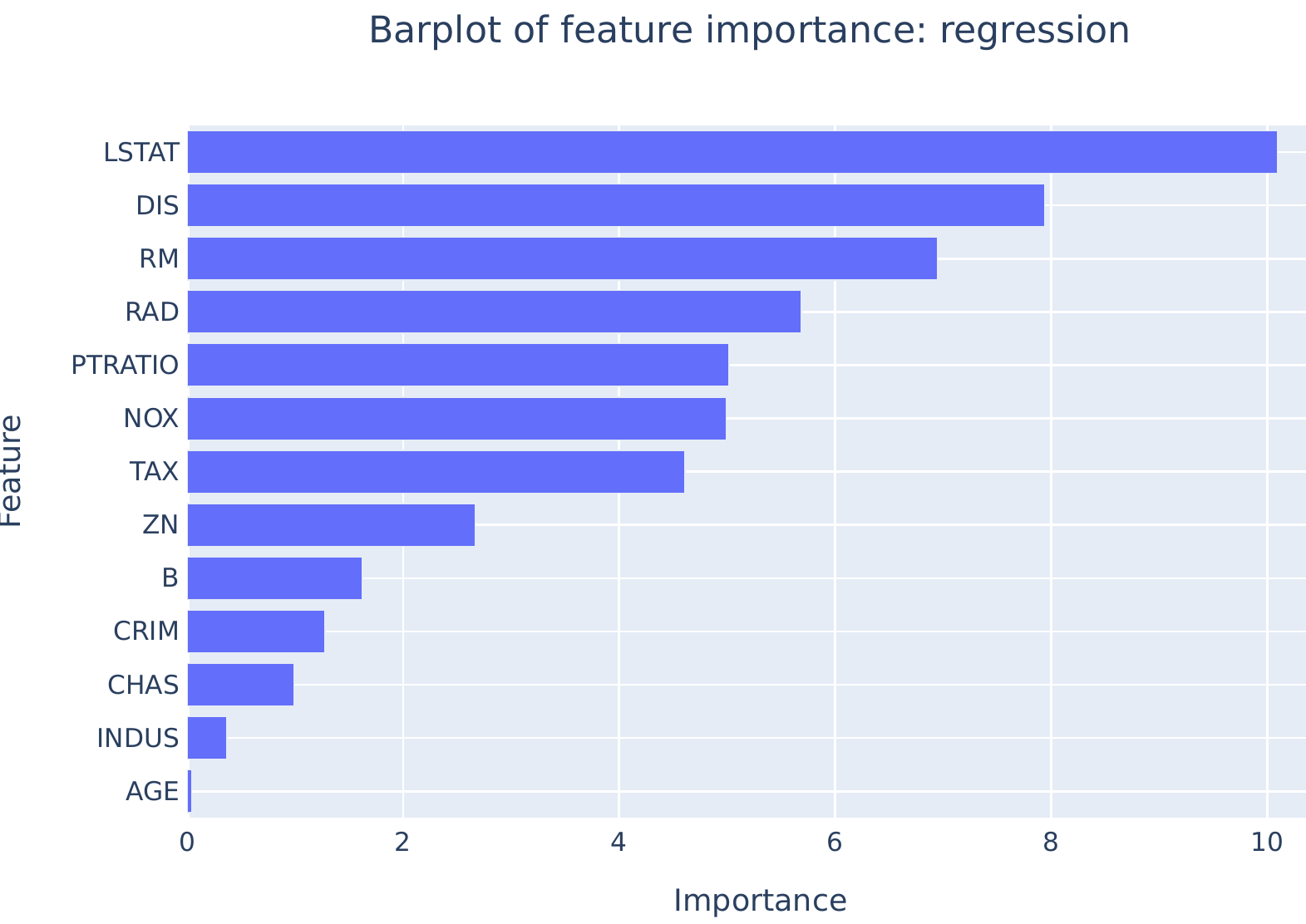}}
		\hspace{0.2cm}
		\subfloat[SHAP Feature Importance]{\includegraphics[width=0.48\textwidth]{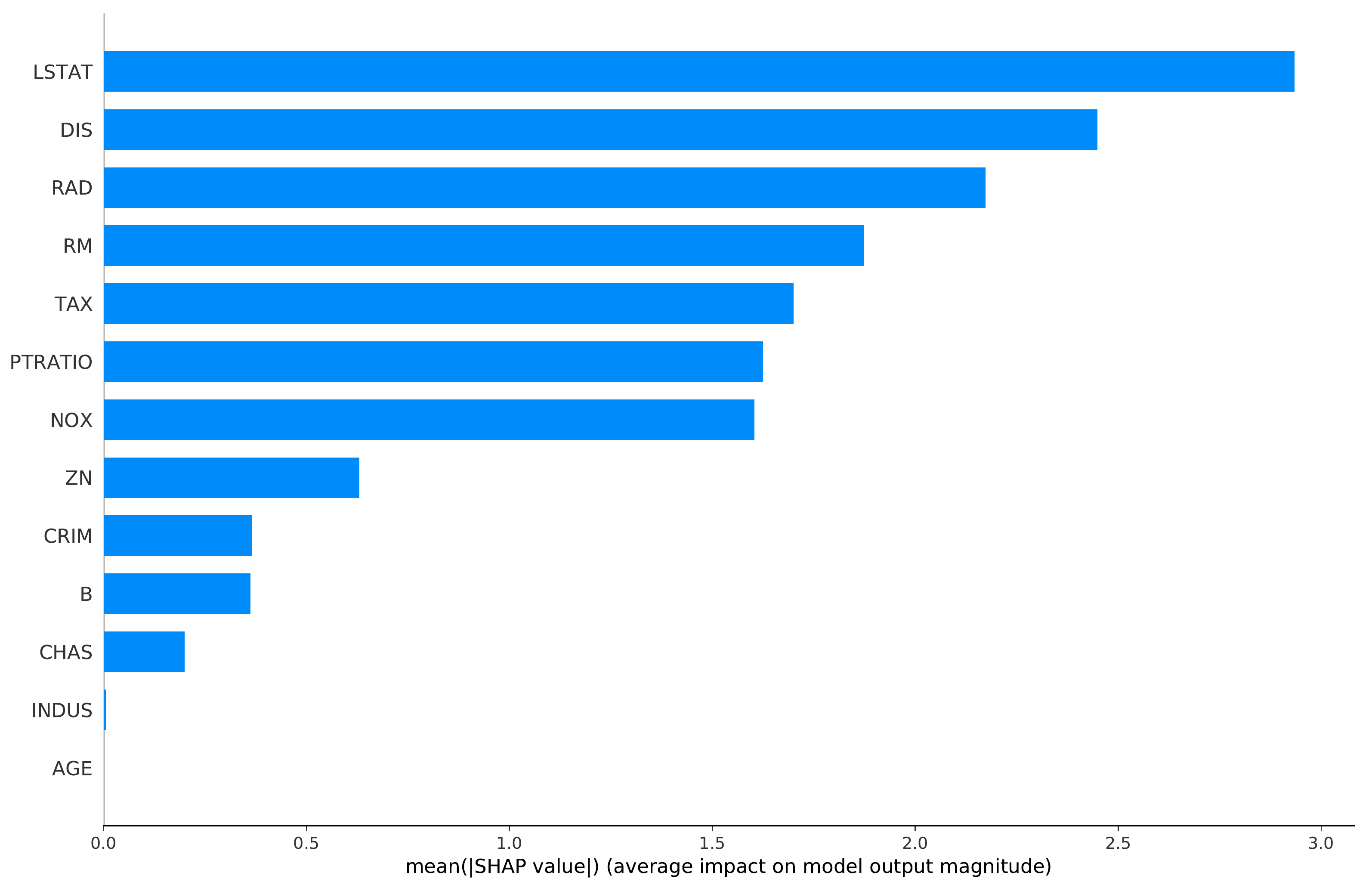}}
    \caption{[Boston Housing Dataset] LINEAR REGRESSION: Result comparison of KernelSHAP and \ourmethod.}
    \label{fig:LR}
\end{figure}

\begin{figure*}
		\subfloat[\ourmethod on Random Forest \label{rf_acme}]{\includegraphics[width=0.5\textwidth]{acme_rf_cropped.pdf}}
		\hspace{0.2cm}
		\subfloat[KernelSHAP on Random Forest \label{rf_shap} ]{\includegraphics[width=0.48\textwidth]{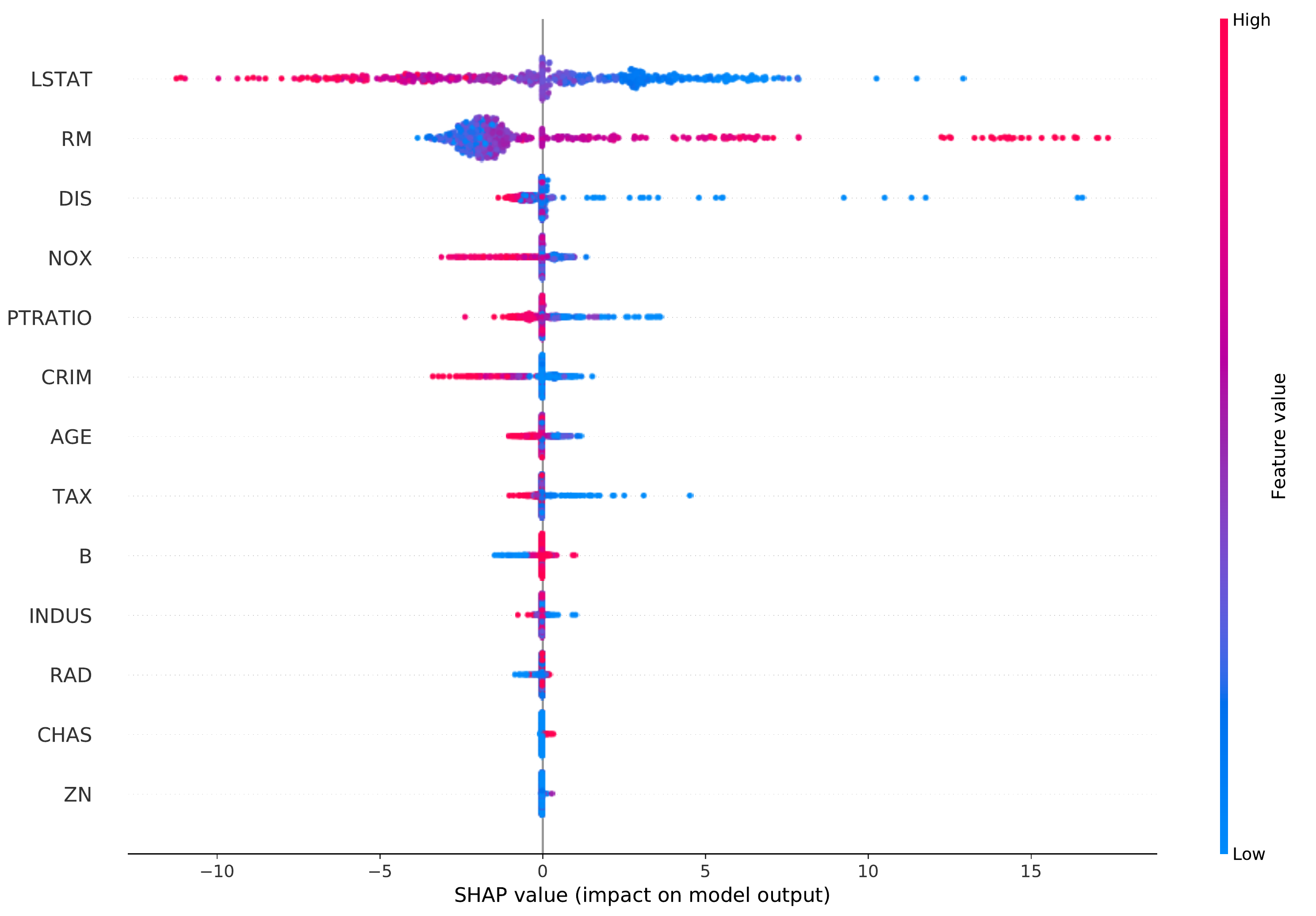}}\\
		\subfloat[\ourmethod Feature Importance]{\includegraphics[width=0.5\textwidth]{barplot/acme_bar_rf_cropped.pdf}}
		\hspace{0.2cm}
		\subfloat[KernelSHAP Feature Importance]{\includegraphics[width=0.48\textwidth]{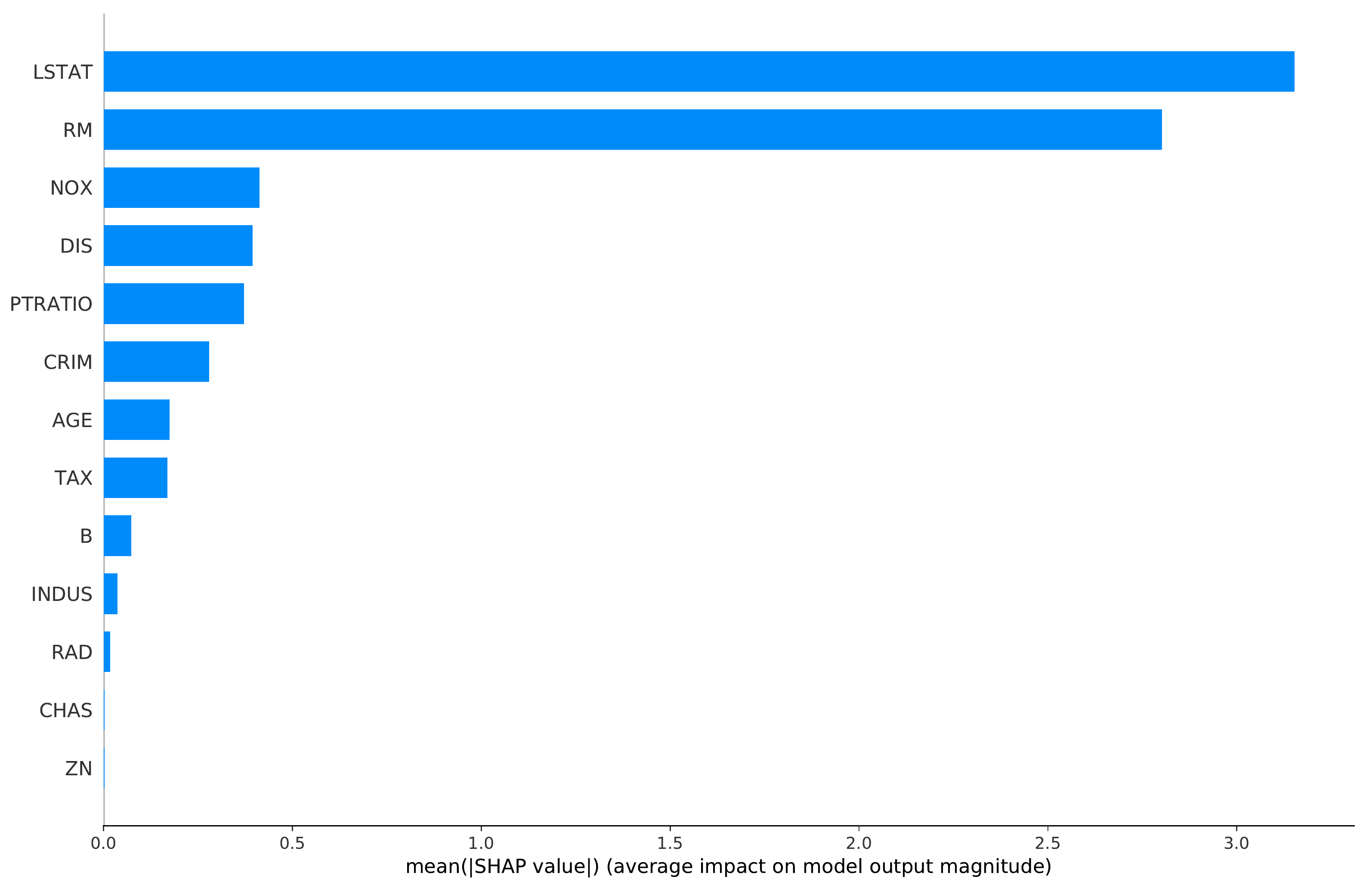}}
    \caption{[Boston Housing Dataset] RANDOM FOREST : Result comparison of KernelSHAP and \ourmethod }
    \label{fig:RF_boston_comparison}
\end{figure*}

\begin{figure*}
		\subfloat[\ourmethod on Catboost Regression]{\includegraphics[width=0.5\textwidth]{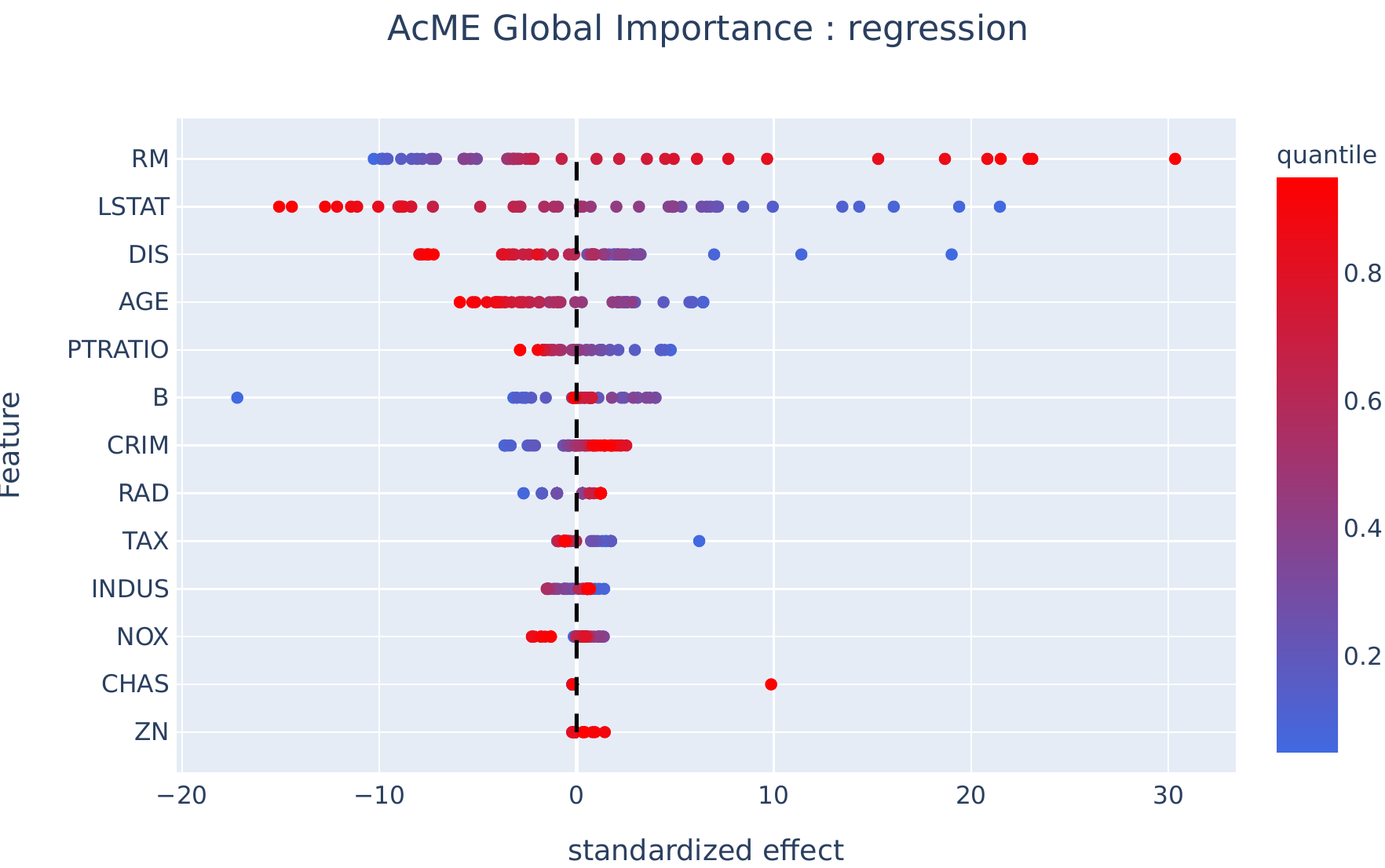}}
		\hspace{0.2cm}
		\subfloat[SHAP on Catboost Regression]{\includegraphics[width=0.48\textwidth]{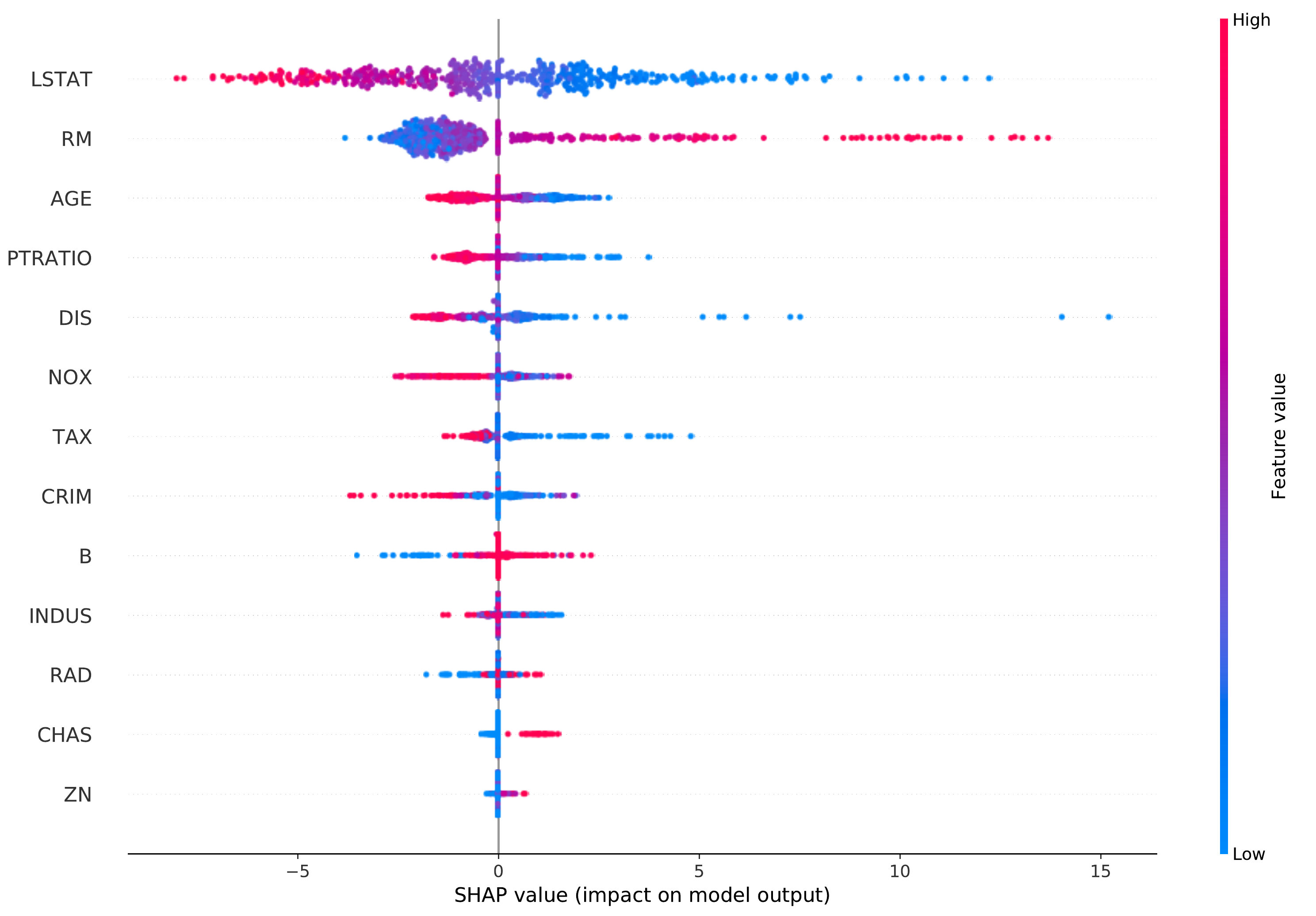}}\\
		\subfloat[\ourmethod Feature Importance]{\includegraphics[width=0.5\textwidth]{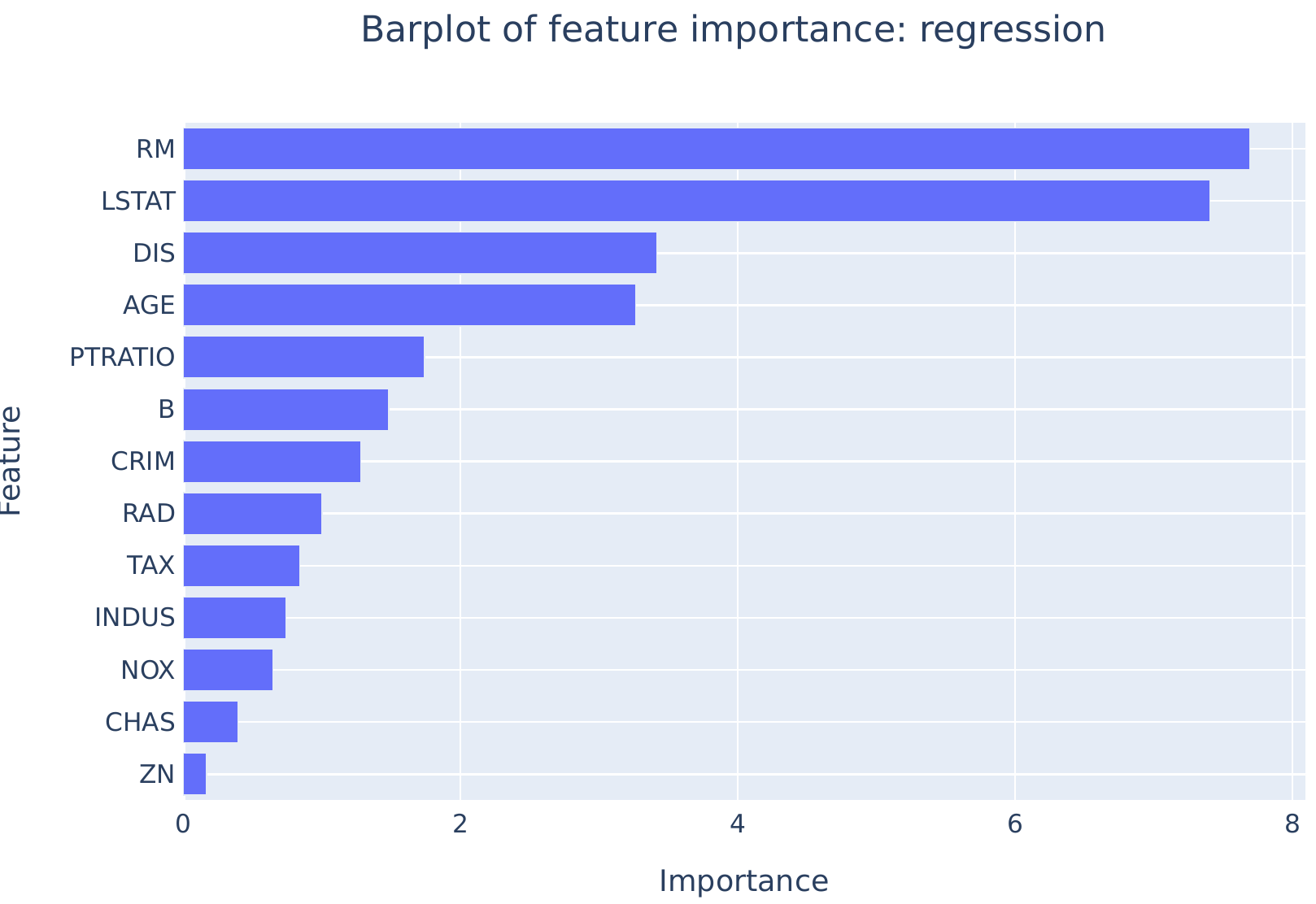}}
		\hspace{0.2cm}
		\subfloat[SHAP Feature Importance]{\includegraphics[width=0.48\textwidth]{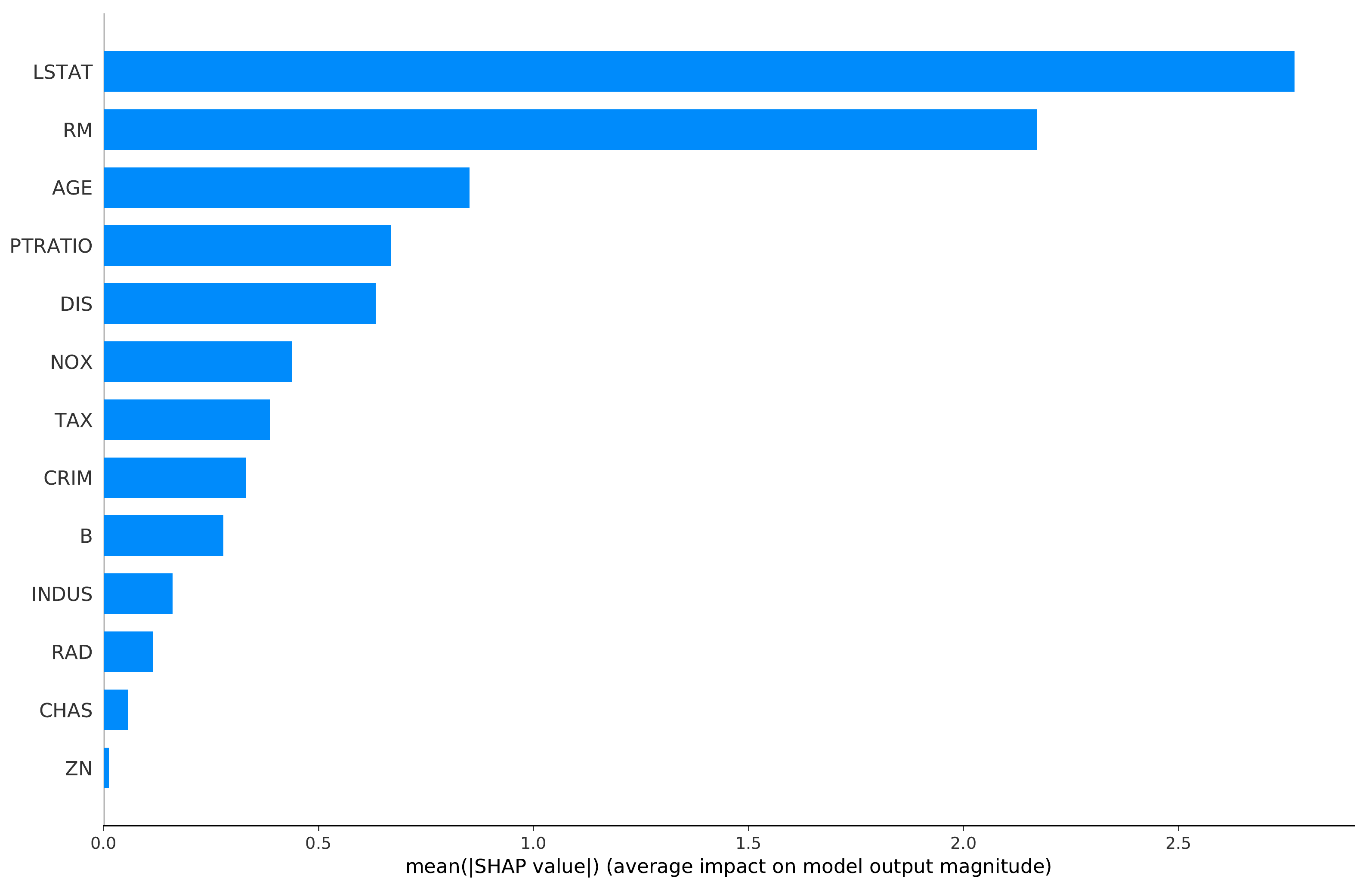}}
    \caption{[Boston Housing Dataset] CatBoost REGRESSION: Result comparison of KernelSHAP and \ourmethod }
    \label{fig:CT}
\end{figure*}

\chiara{Additional experiments with other real-world dataset are reported in the Appendix.}


\subsubsection{Local interpretability}
\begin{figure}[t]
    \centering 
    {\includegraphics[width=0.75\textwidth]{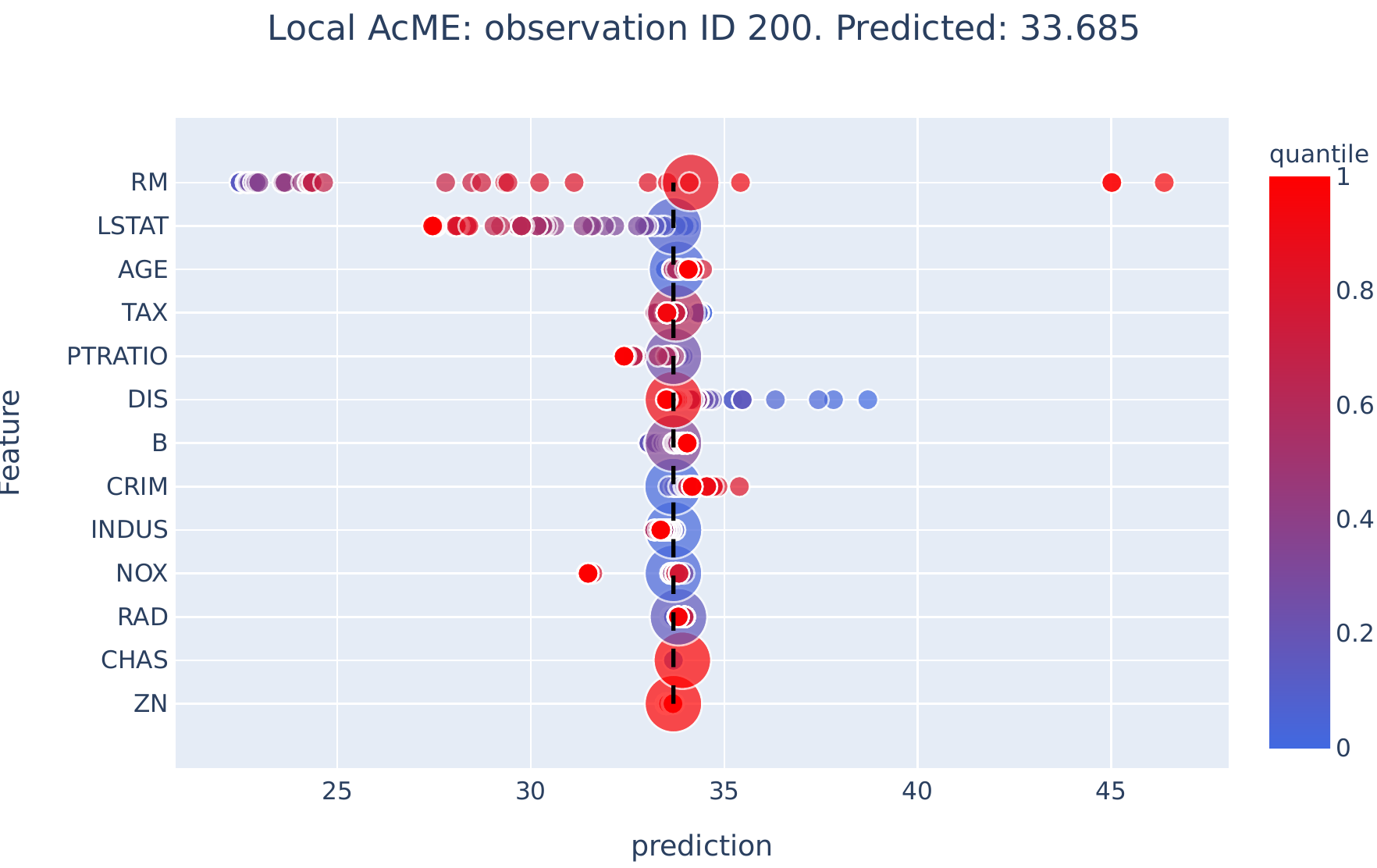}}
    \caption{\ourmethod result plot for single observation.
    The underlying model is a Random Forest with 100 tree, trained on the Boston Housing dataset }\label{fig:acme_local} 
\end{figure}

\david{Figure \ref{fig:acme_local} depicts local interpretability results for observation with $ID = 200$, when considering a Random Forest model. 
From the plot, we can understand how each feature impacts prediction and explore how variations in input values can affect estimates using a \textit{what-if} approach:
\begin{itemize}
    \item variable \texttt{RM} is the number of rooms, it has a high value (big red bubble), and this increases the estimated value of the house. We could easily see from the plot that a house with the same characteristics but with a lower number of rooms will have a lower price, going from actual 33.685 to less than 25;
    \item variable \texttt{LSTAT} represent the proportion of the population that is of lower status (low education or low income). The estimated value is higher due to the low value of this feature. It is clearly seen that as the value of this proportion increases, the estimated price lowers considerably.
    \item variable \texttt{DIS} is weighted distance to five Boston employment centres, and for this house it is near to the highest encountered in the dataset. For the model, this is a negative factor that reduces the house estimated value, and as we can see, a house with the same characteristics could have a much higher value by reducing the distance: from the actual value of 33.685 it could go near to 38-39.
\end{itemize}
}

\subsection{Experiments on a classification task: Glass dataset}
\label{sec:real_data_classification}
\begin{figure}[t]
\centering
    \subfloat[\ourmethod Feature Importance]{\includegraphics[width=0.48\textwidth]{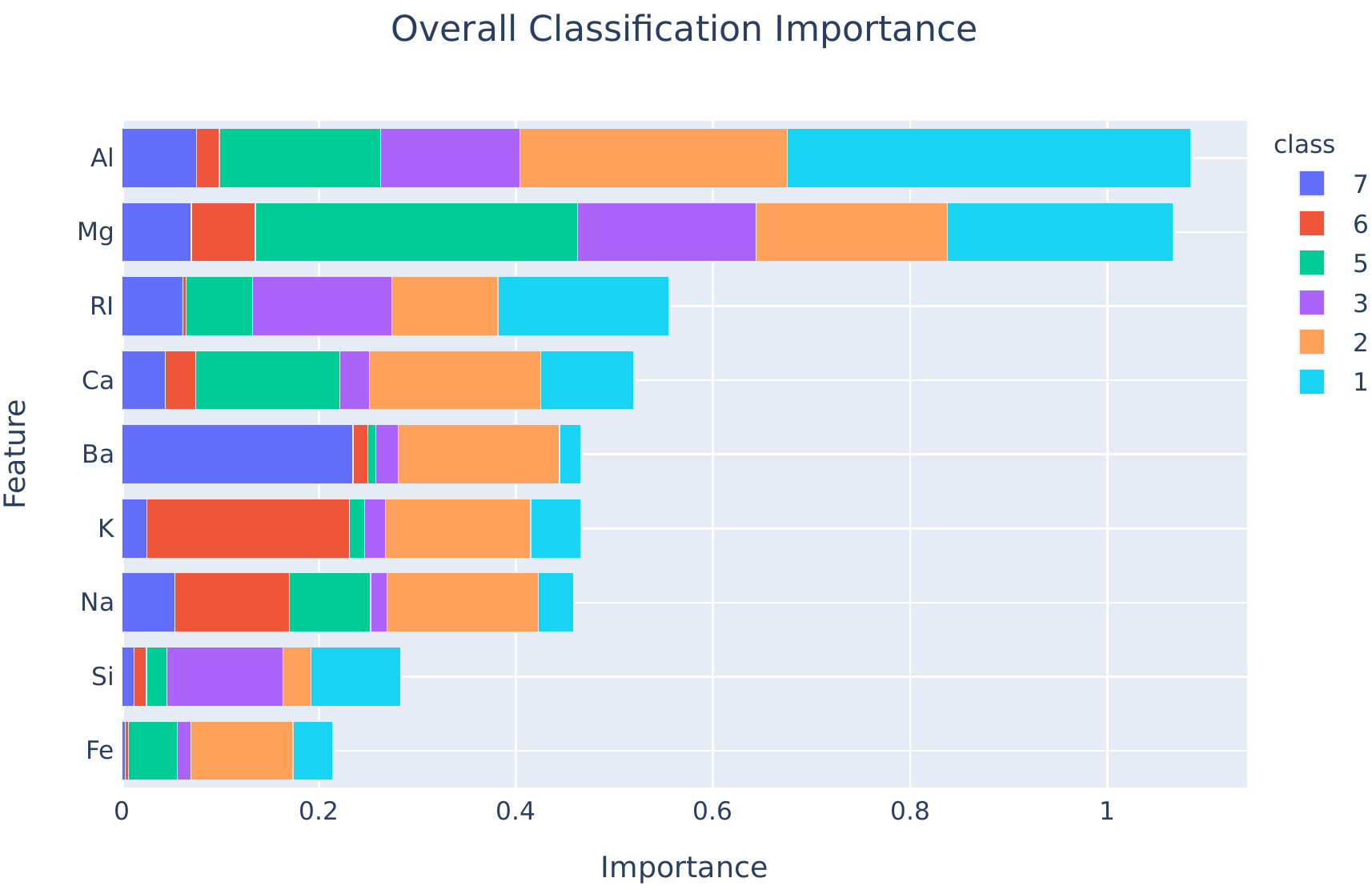}}
	\subfloat[KernelSHAP Feature Importance]{\includegraphics[width=0.52\textwidth]{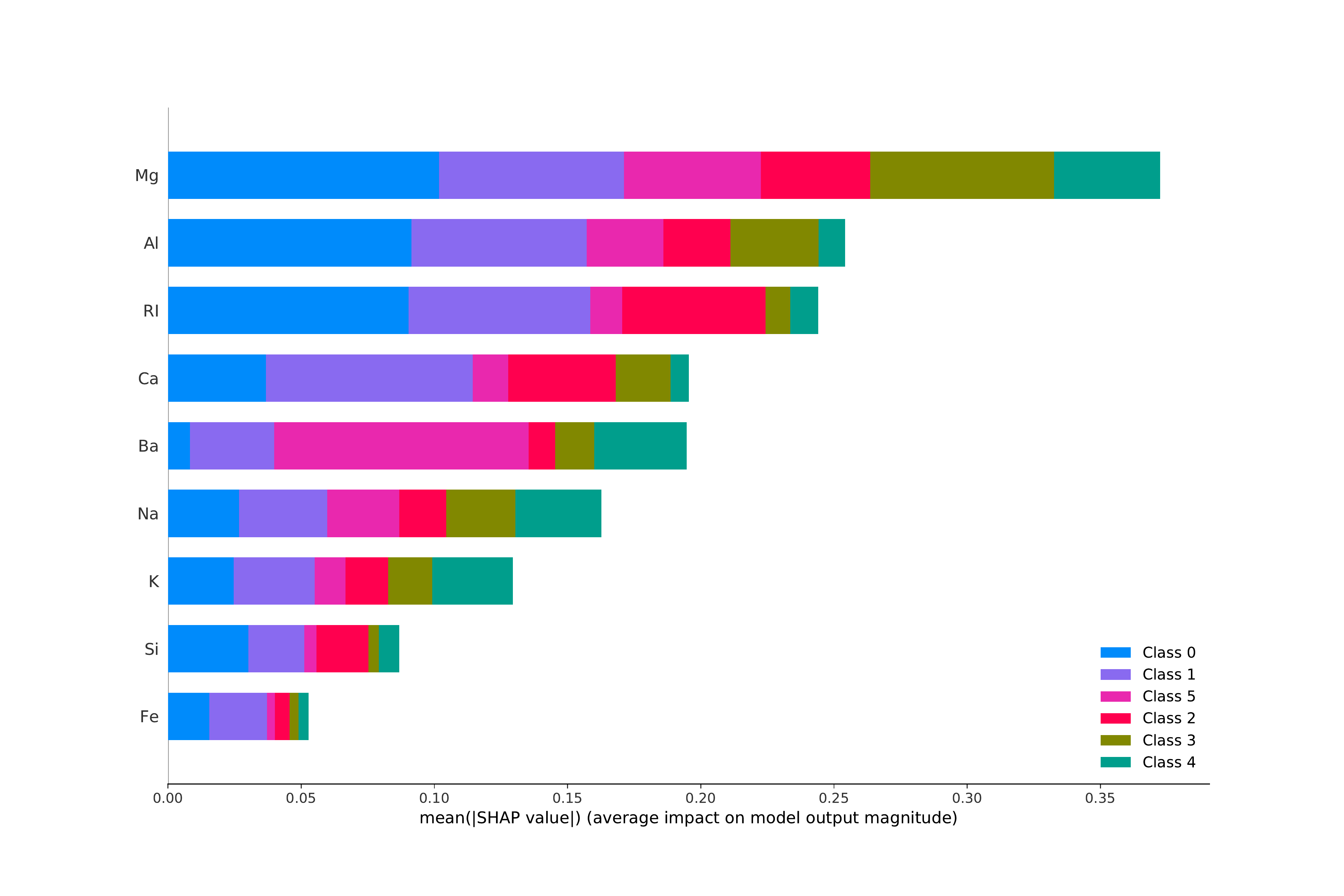}}
    \caption{[Glass dataset] Comparison of feature importance provided by  KernelSHAP and \ourmethod.}
    \label{fig:Glass}
\end{figure}

\chiara{This experiment aims at evaluating the performance of \ourmethod on a classification task. To this aim we consider a well-known publicly available dataset, Glass\footnote{https://archive.ics.uci.edu/ml/datasets/glass+identification}\label{glass}. The goal is to classify six different types of glass based on their chemical features. 
We use a CatBoost model to resolve the multiclass classification problem. Then, we compare \ourmethod with KernelSHAP}.
From Figure \ref{fig:Glass} we cans see that \ourmethod (with $K=20$) and KernelSHAP provide similar explanations about global importance for input features.
However, while \ourmethod runs in 0.58 seconds, it takes 186.5 seconds to obtain the results from KernelSHAP.

\begin{figure}[t]
\centering
		\subfloat[\ourmethod on the Glass dataset: glass type 5]{\includegraphics[width=0.5\textwidth]{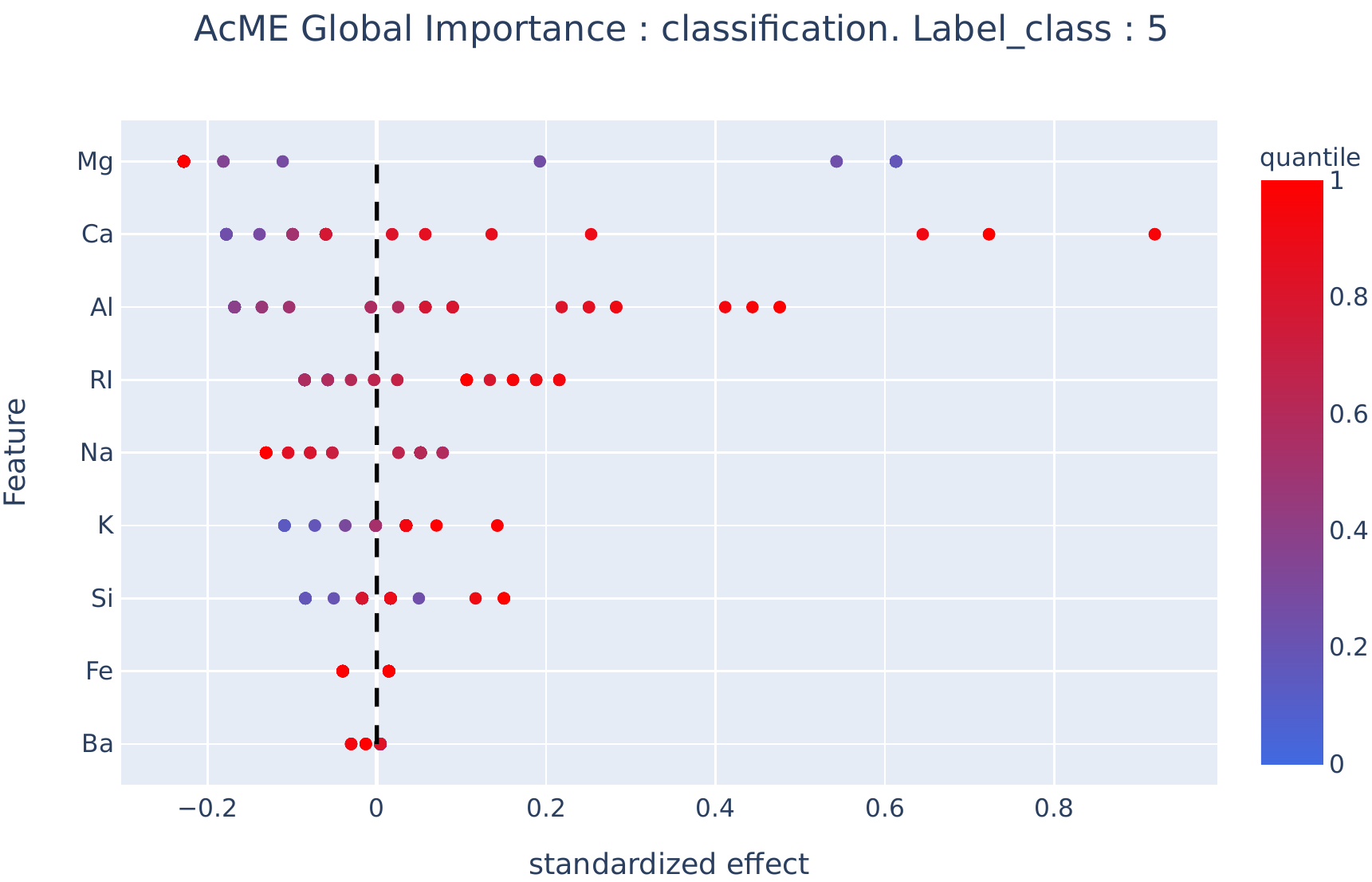}}
		\hspace{0.2cm}
		\subfloat[SHAP on the Glass dataset: glass type 5]{\includegraphics[width=0.48\textwidth]{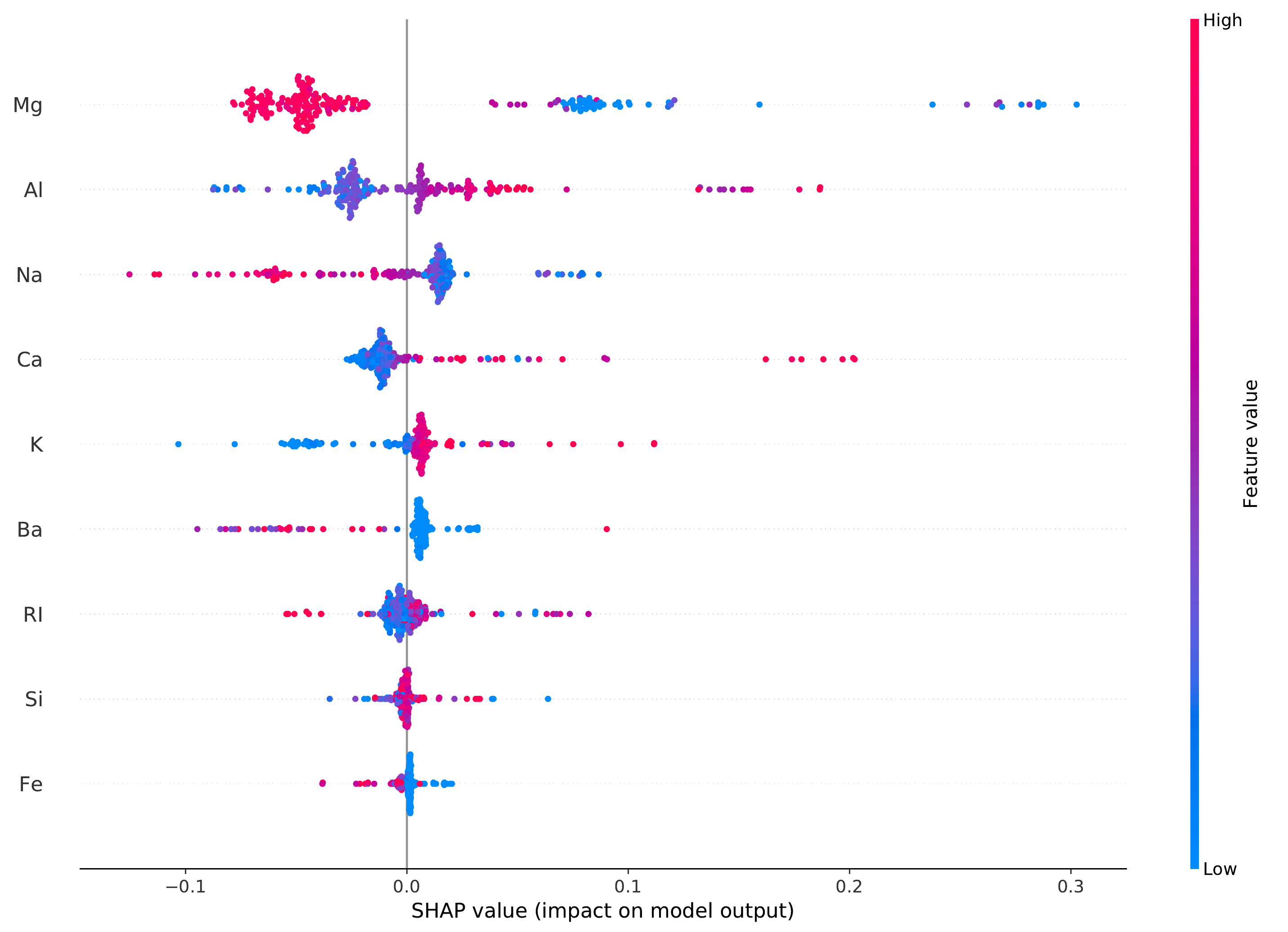}}\\
		\subfloat[\ourmethod on the Glass dataset: glass type 6]{\includegraphics[width=0.5\textwidth]{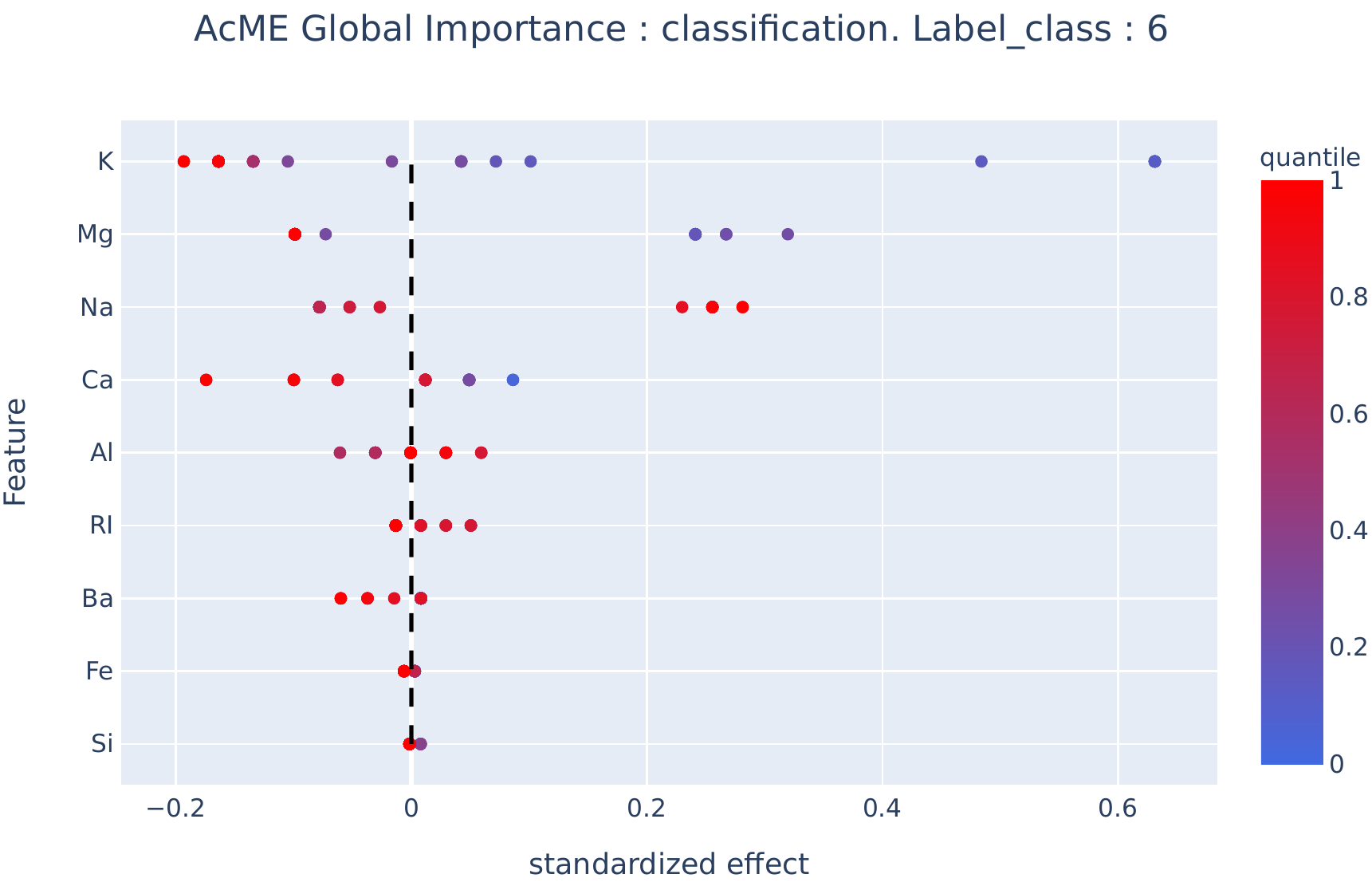}}
		\hspace{0.2cm}
		\subfloat[SHAP on the Glass dataset: glass type 6]{\includegraphics[width=0.48\textwidth]{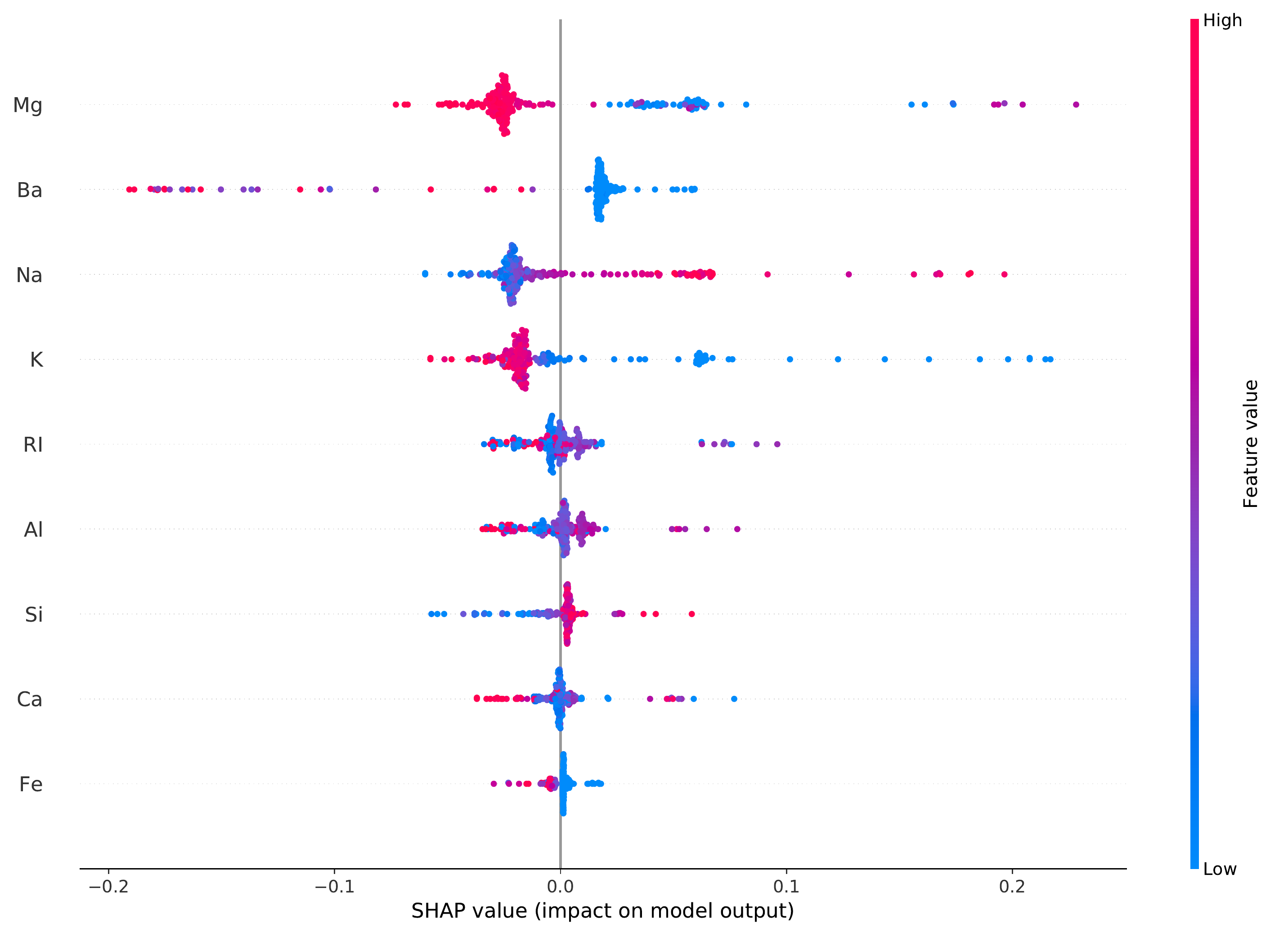}}\\
		\subfloat[\ourmethod on the Glass dataset: glass type 7]{\includegraphics[width=0.5\textwidth]{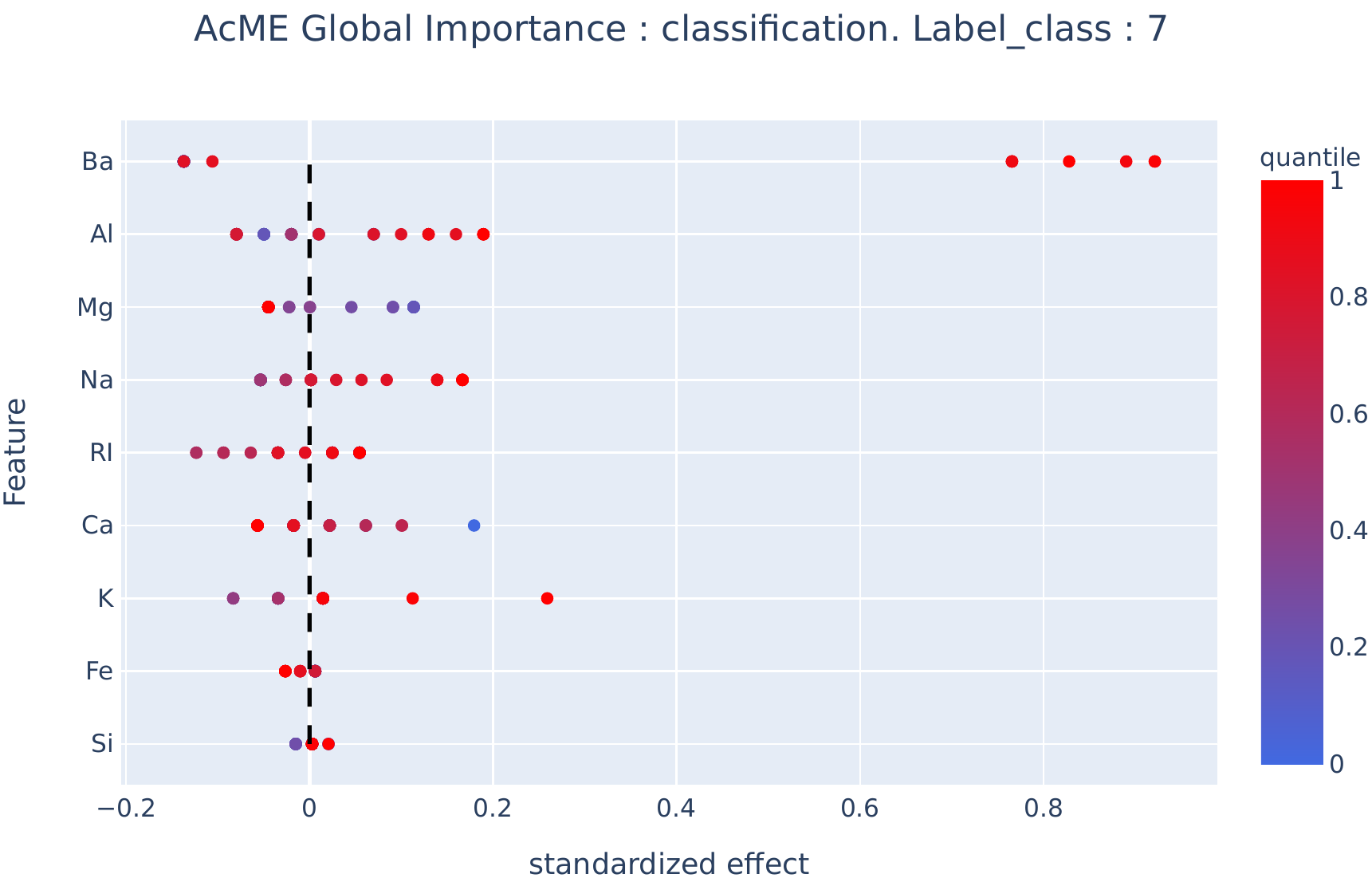}}
		\hspace{0.2cm}
		\subfloat[SHAP on the Glass dataset: glass type 7]{\includegraphics[width=0.48\textwidth]{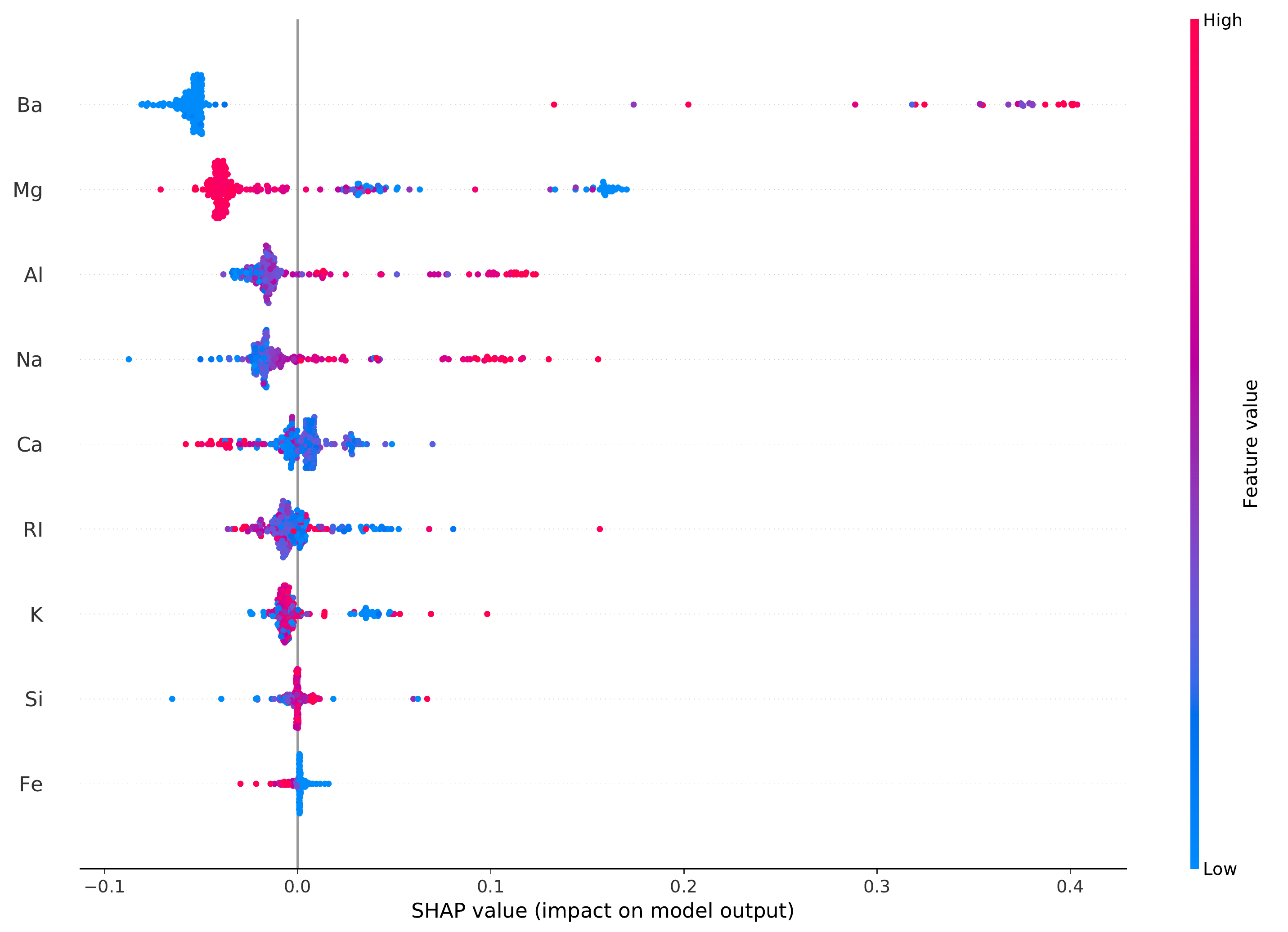}}
			\caption{[Glass dataset]: Feature effect according to \ourmethod (left) and KernelSHAP (right)}
	\label{fig:GlassSingleClasses}
\end{figure}

In Figure \ref{fig:GlassSingleClasses}, instead, we can see how the detailed visualizations provided by \ourmethod can be adapted to multi-class classification tasks by considering each class predicted probability, in the same vein as SHAP.

\begin{figure}[th]
    \centering
    \includegraphics[scale=0.65]{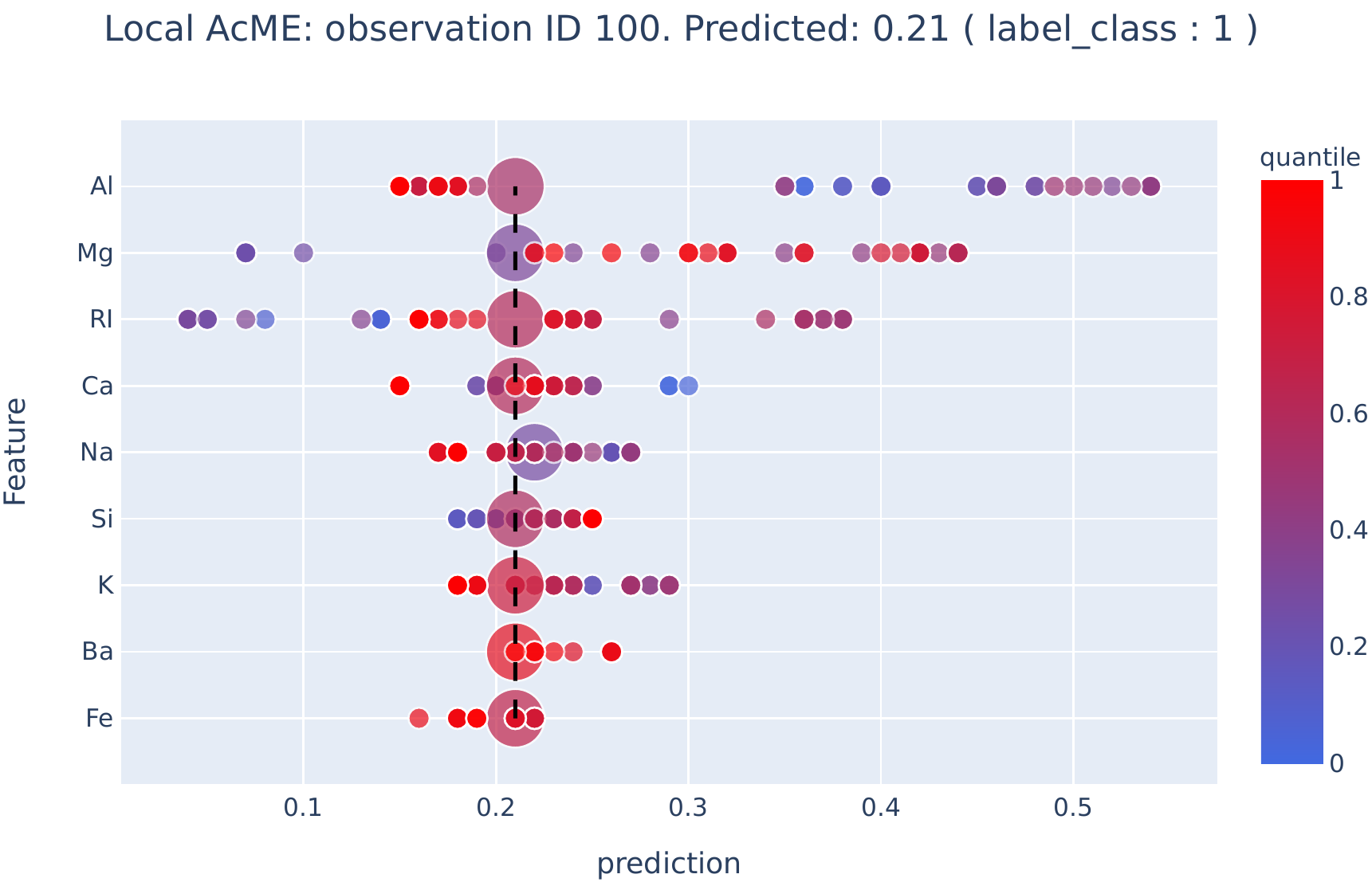}
    \caption{\ourmethod local importance scores visualization for Glass classification task.}
    \label{fig:acme_local_class}
\end{figure}

\chiara{Finally, in Figure \ref{fig:acme_local_class} we show how local interpretability can be used as a \textit{what-if} analysis tool. For example, we see that observation with $ID = 100$ has probability 0.21 of belonging to class 1, according to the CatBoost model. Everything else remaining fixed, this probability would grow over 0.5 if we decrease the aluminium content (Al) from quantile 0.55 (bigger bubble corresponding to the current observation value) to quantile 0.47, corresponding to the rightmost dot first row of the chart.
Instead, if we reduced Refractive Index (RI) from current value (quantile 0.59) to quantile 0.33, the resulting class 1 probability would be 0.04.} 


\begin{figure}[t]
\centering
		\subfloat[\ourmethod on the Glass dataset: glass type 1]{\includegraphics[width=0.5\textwidth]{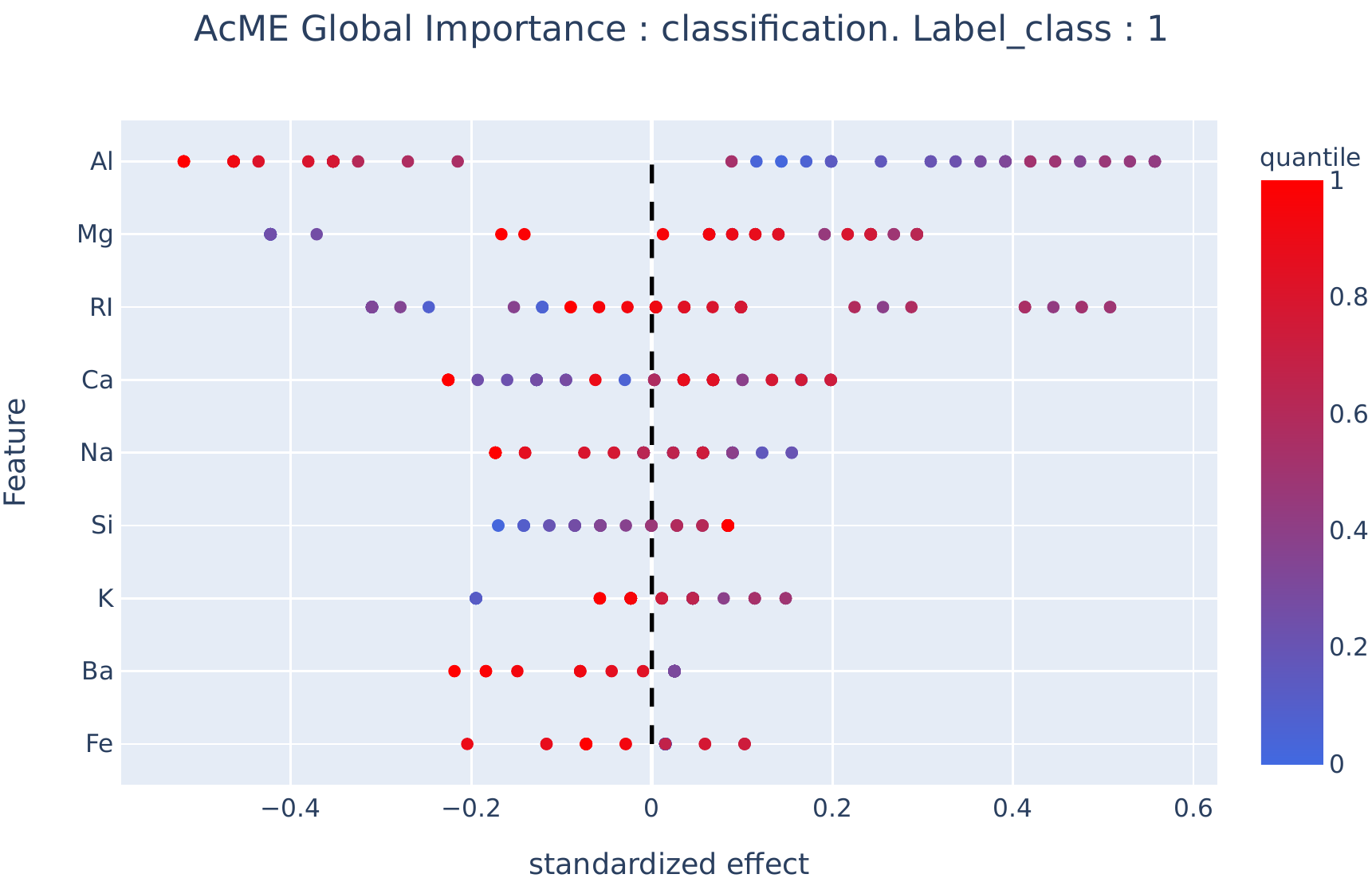}}
		\hspace{0.2cm}
		\subfloat[SHAP on the Glass dataset: glass type 1]{\includegraphics[width=0.48\textwidth]{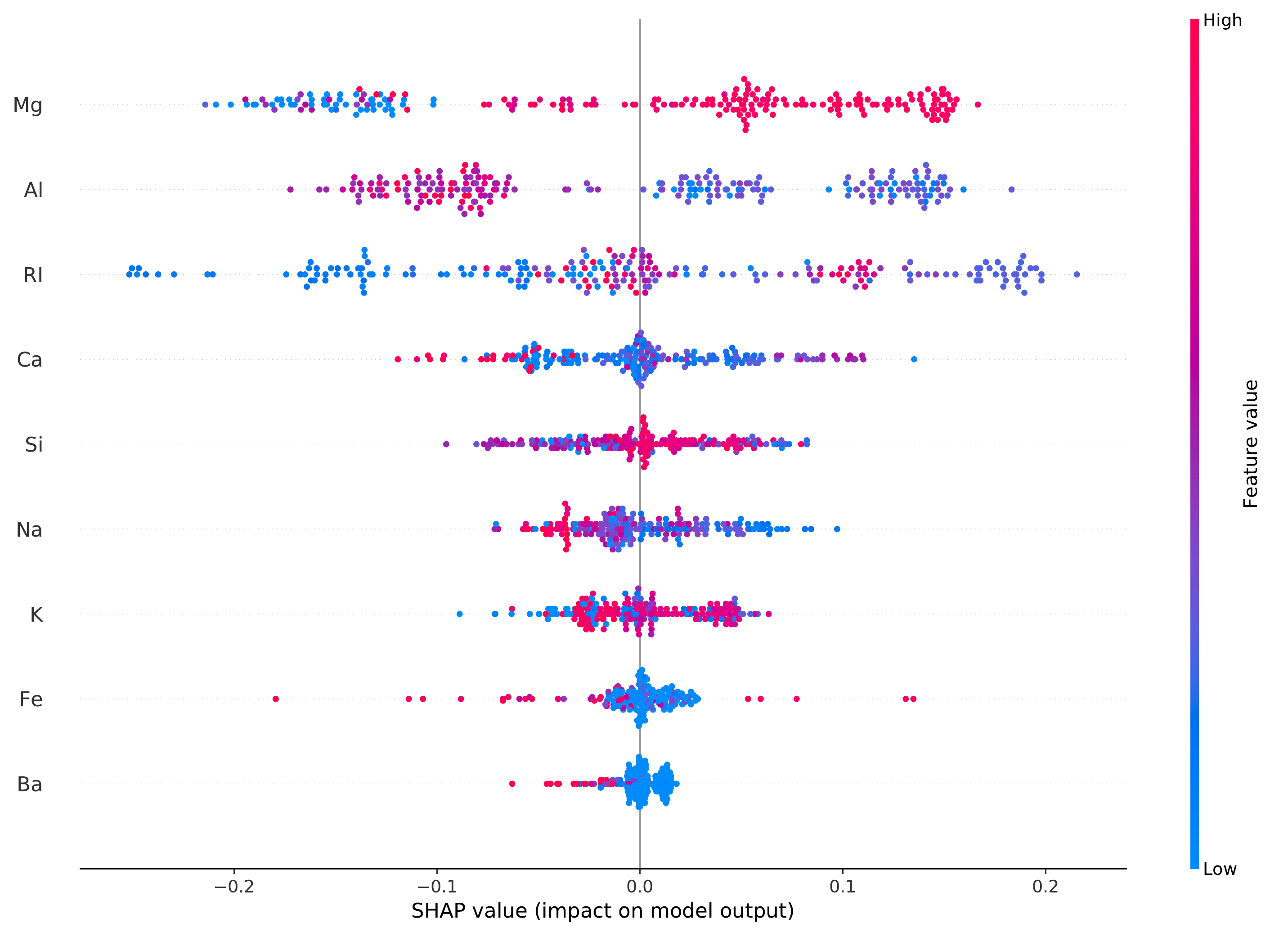}}\\
		\subfloat[\ourmethod on the Glass dataset: glass type 2]{\includegraphics[width=0.5\textwidth]{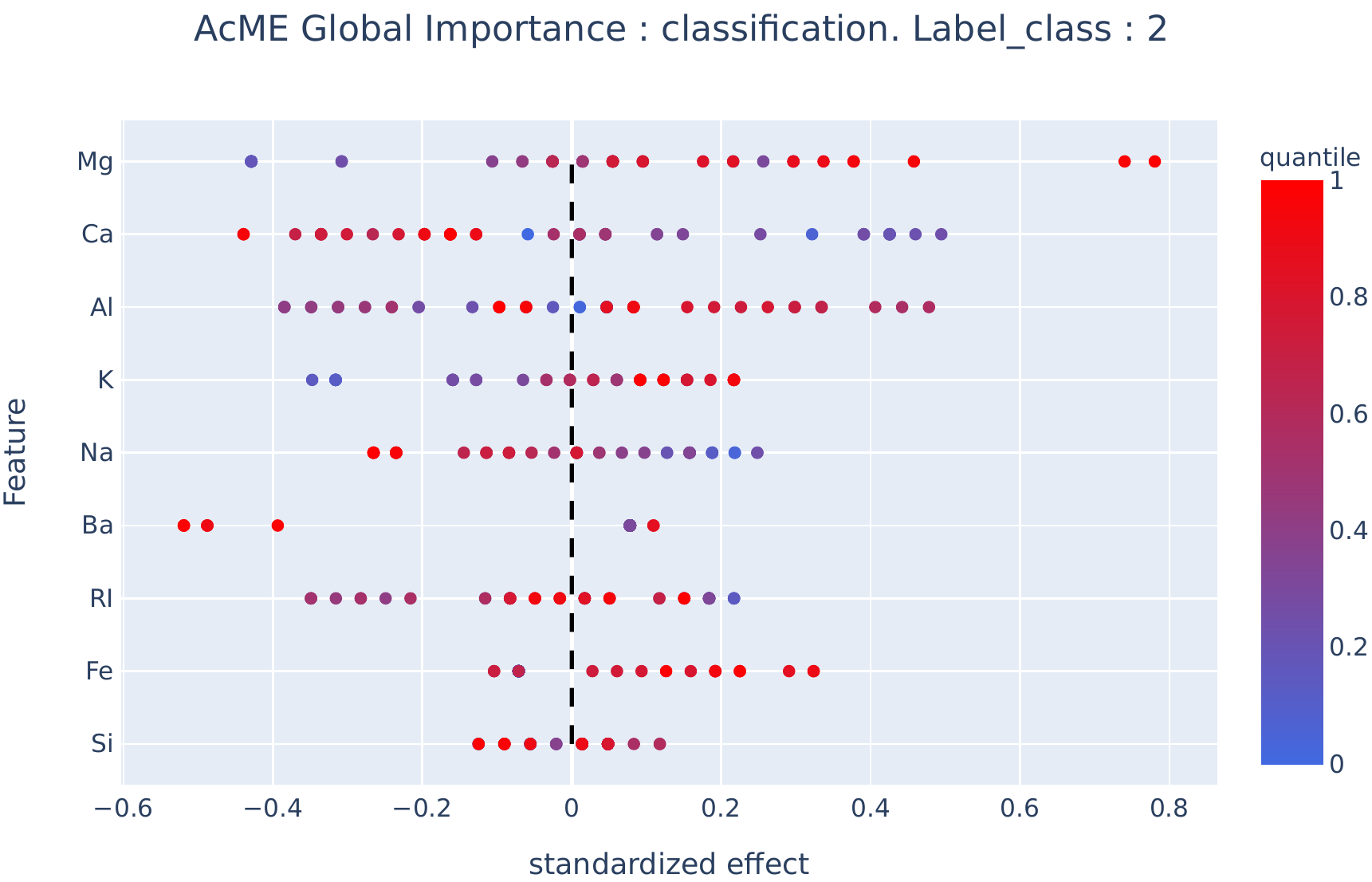}}
		\hspace{0.2cm}
		\subfloat[\ourmethod on the Glass dataset: glass type 2]{\includegraphics[width=0.48\textwidth]{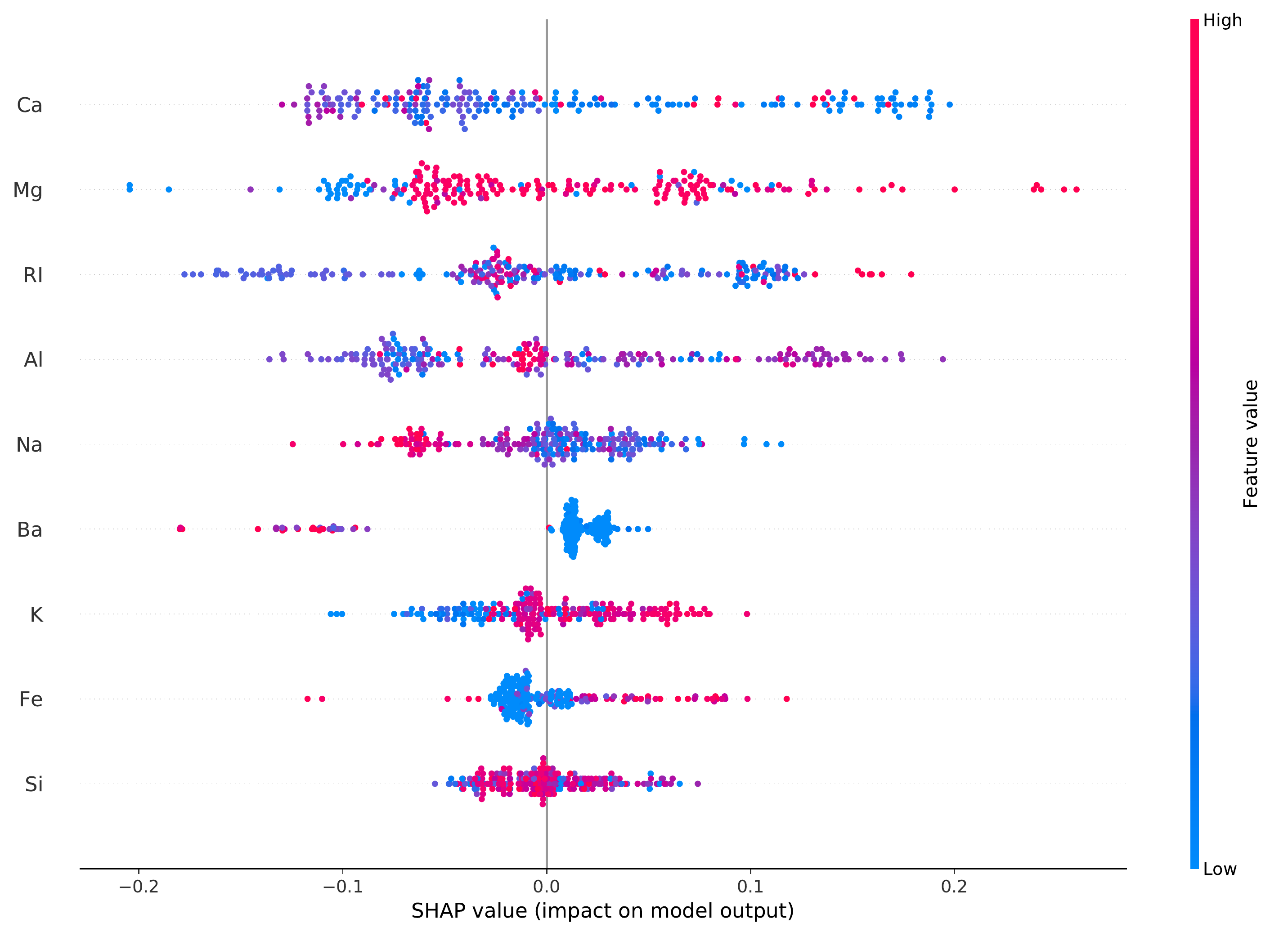}}\\
		\subfloat[\ourmethod on the Glass dataset: glass type 3]{\includegraphics[width=0.5\textwidth]{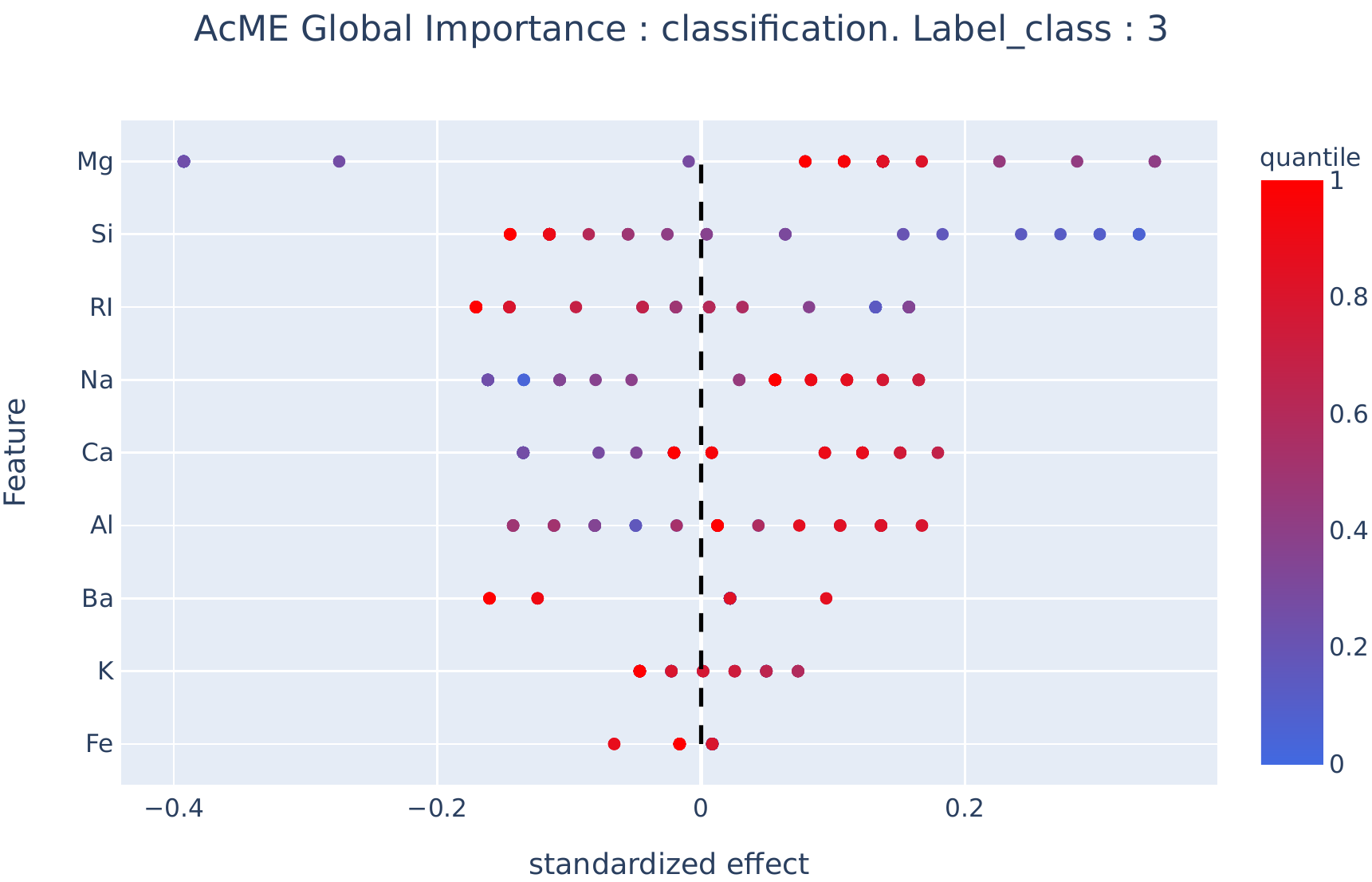}}
		\hspace{0.2cm}
		\subfloat[\ourmethod on the Glass dataset: glass type 3]{\includegraphics[width=0.48\textwidth]{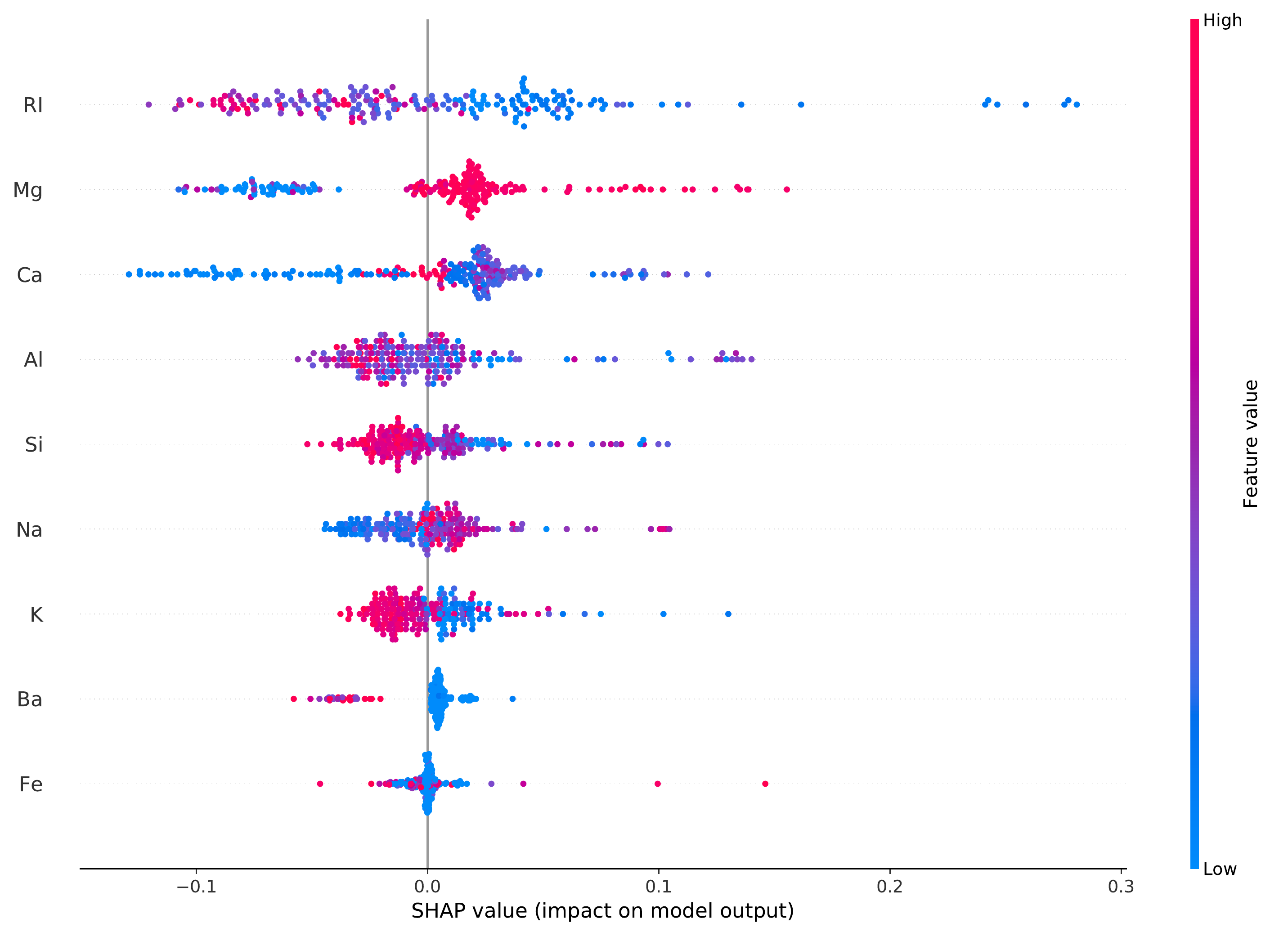}}
\end{figure}%
\newpage

\FloatBarrier
\section{Conclusions and future research directions}

\chiara{We introduced \ourmethod, a model-agnostic interpretability approach designed with execution speed in mind. This requirement is of paramount importance in human-in-the-loop applications where users need to take corrective actions quickly, e.g., Decision Support Systems for Fraud Detection or Fault Classification, to name a few examples. \ourmethod fulfils this requirement, as it is very efficient and scales up well with the dataset size. Indeed, \ourmethod does not require any model retraining step, nor it needs to apply model prediction to each element in the dataset. 
We designed the proposed method to explain predictions both at the global and local levels. Experimental results suggest that \ourmethod produces global explanations similar to those provided by SHAP, in a fraction of the computation time. As for local interpretability, differently from SHAP, \ourmethod provides a simple \textit{what-if} tool that allows users to figure out how changes in feature values may affect the predictions.
\\
In this work, we focused on regression and classification. However, we can extend \ourmethod to every task that requires estimating a numerical score. For example, \ourmethod is suitable also for unsupervised Anomaly Detection, where observations are assigned an anomaly score to detect the most abnormal behaviours. Exploiting the local interpretability, we could focus on a sample and use \ourmethod to evaluate how changes in the input feature values would impact the corresponding anomaly score. Thus, \ourmethod would act as a Root Cause Analysis tool, in the sense that it may help users to figure out why the algorithm deemed a sample as normal or abnormal. In the scenario of Anomaly Detection, where prompt corrective actions can translate into a reduced waste of money, time, and materials, the execution speed of interpretability procedures is even more relevant.
\\
Besides the application to Anomaly Detection, another interesting future research direction is estimating and visualizing the combined effect of variables. Currently, \ourmethod has no specific strategy for managing the relations among features since perturbation analysis involves only one variable at a time. This limitation is a common shortcoming for most model-agnostic interpretability methods, and deserves further investigation. 
}


\bibliography{mybibfile}

\newpage
\appendix
\section{SHAP and ACME comparison in other dataset}
To show the quality of the results obtained by \ourmethod,   \ourmethod, we tested it on others public datasets:
\begin{itemize}
    \item Wine\footnote{https://archive.ics.uci.edu/ml/datasets/wine}\label{wine}: the data describes chemicals quality of different wines, ranked with a quality score. We use a random forest to predict the quality score in a regression problem. This is another example of interpretability in regression tasks in addition to the one in \ref{sec:real_data}.
    \item Exams\footnote{https://www.kaggle.com/spscientist/students-performance-in-exams}\label{exams}: this dataset keeps record of the marks obtained by students in the writing, reading, and math tests. It also adds information about the students gender, race and other socio-economic variables. We use CatBoost model to predict the math test score using all the other variables in regression problem.
\end{itemize}

\begin{figure}[H]
		\subfloat[\ourmethod on the Wine dataset]{\includegraphics[width=0.48\textwidth]{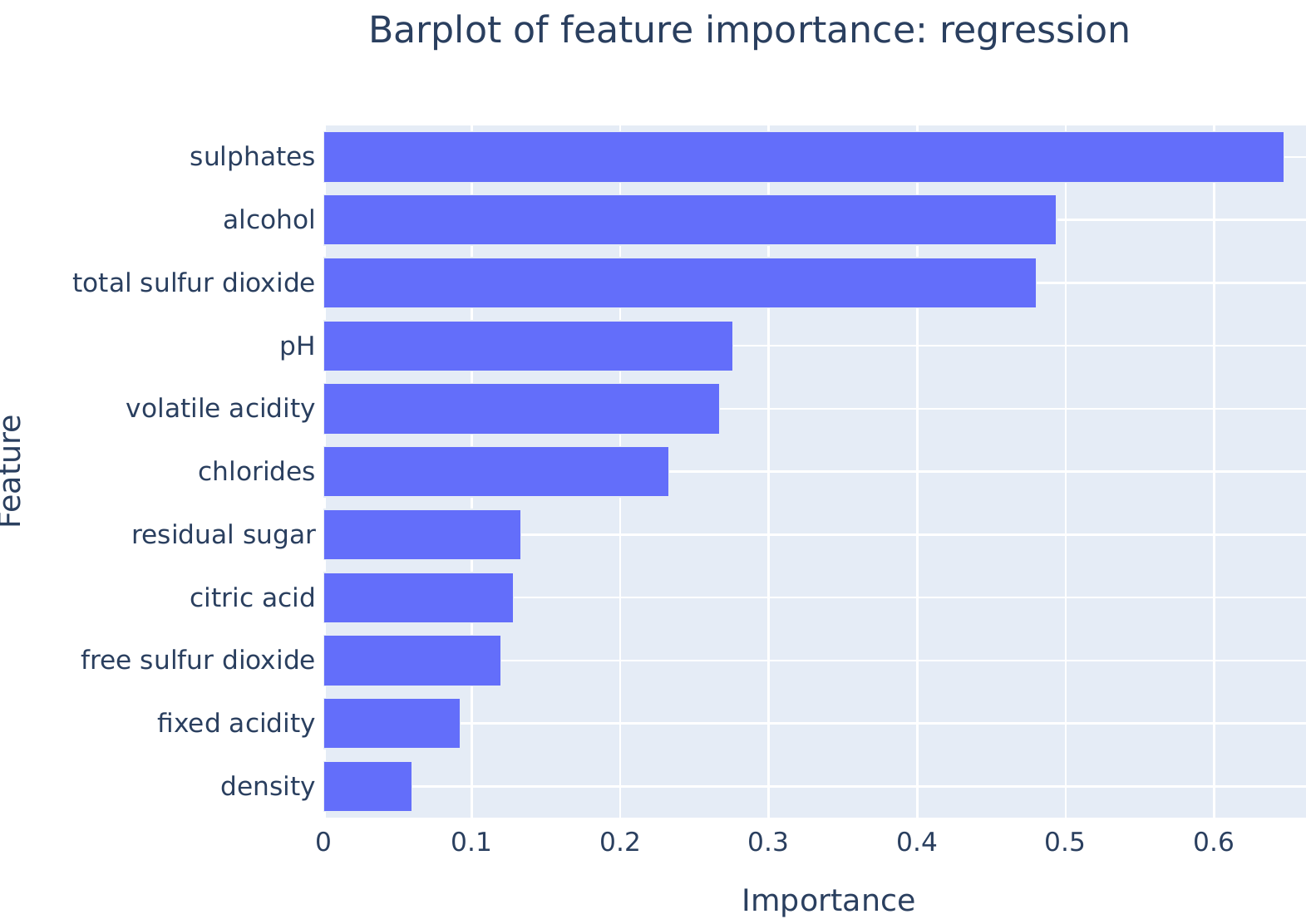}}
		\hspace{0.2cm}
		\subfloat[KernelSHAP on the Wine dataset]{\includegraphics[width=0.5\textwidth]{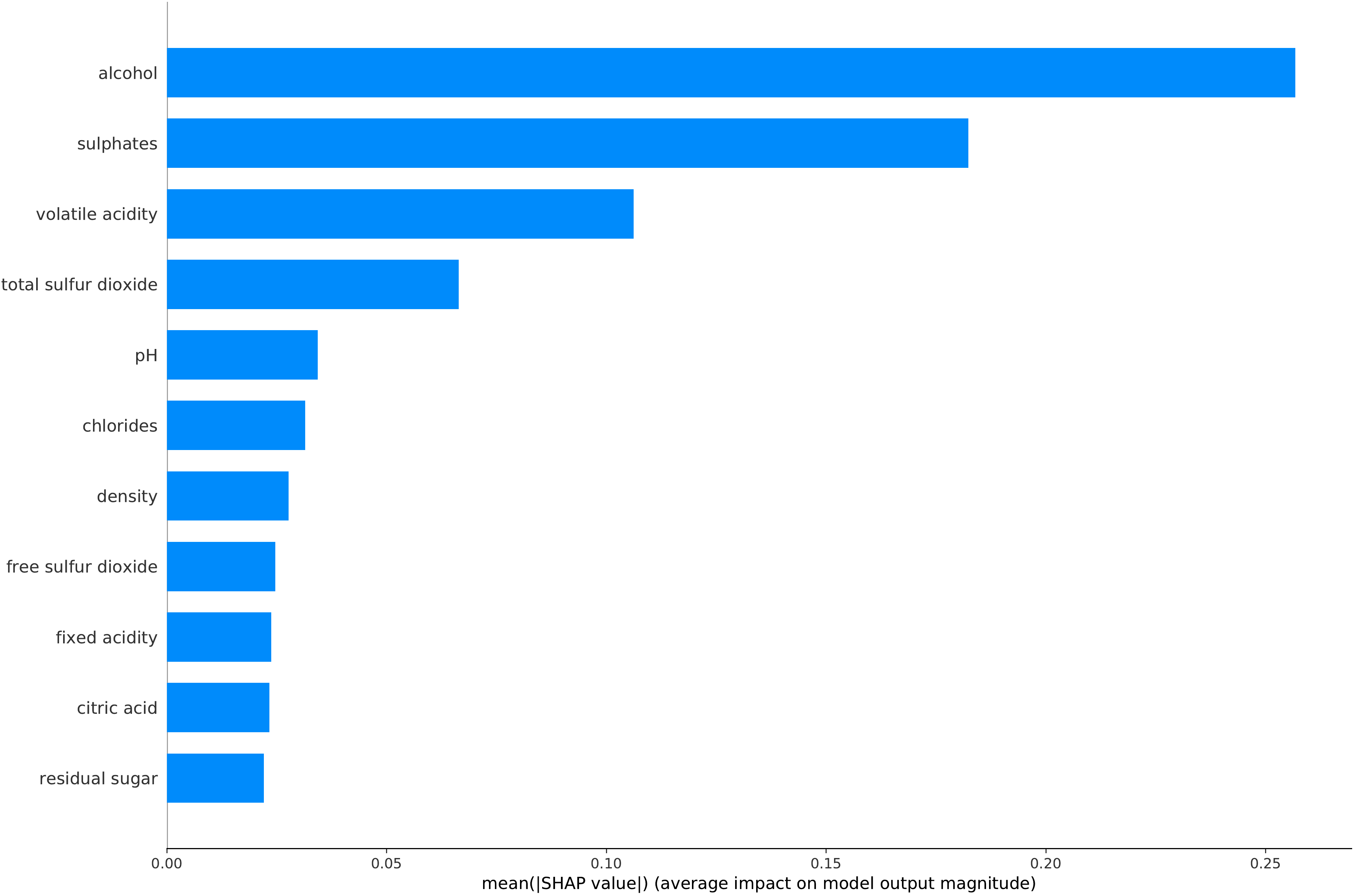}}\\
		\subfloat[\ourmethod Feature Importance]{\includegraphics[width=0.48\textwidth]{wine/wine_acme_bar_cropped.pdf}}
		\hspace{0.2cm}
		\subfloat[KernelSHAP Feature Importance]{\includegraphics[width=0.5\textwidth]{wine/wine_shap_bar_cropped.pdf}}
    \caption{[Wine dataset] Comparison of feature importance provided by  KernelSHAP and \ourmethod.}
    \label{fig:Wine}
\end{figure}

\begin{figure}[H]
    \subfloat[\ourmethod on the Exams dataset]{\includegraphics[width=0.5\textwidth]{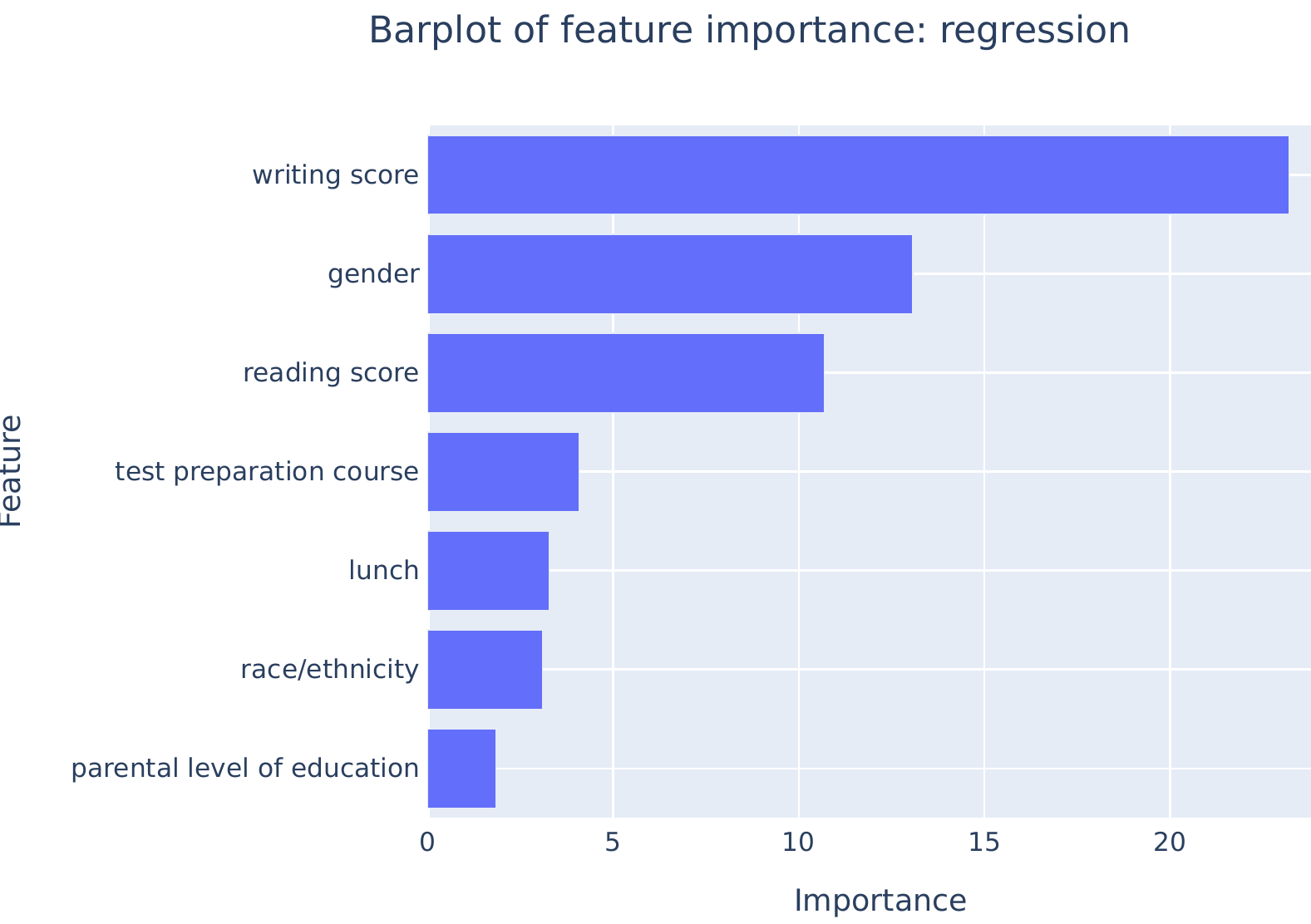}}
	\hspace{0.2cm} \subfloat[KernelSHAP on the Exams dataset]{\includegraphics[width=0.5\textwidth]{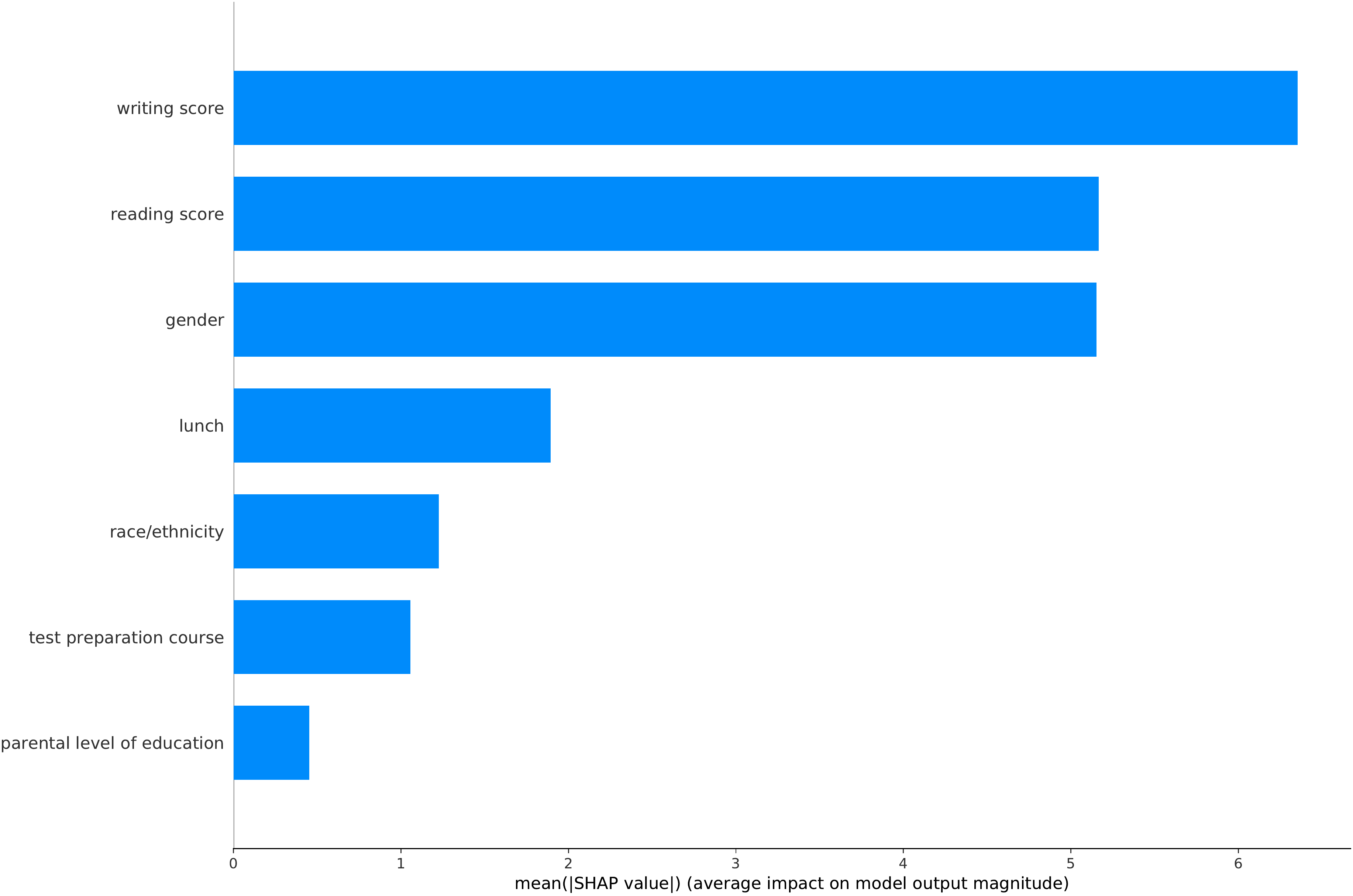}}\\
	\subfloat[\ourmethod Feature Importance]{\includegraphics[width=0.5\textwidth]{exams/exams_acme_bar_cropped.pdf}}
	\hspace{0.2cm}
	\subfloat[KernelSHAP Feature Importance]{\includegraphics[width=0.5\textwidth]{exams/exams_shap_bar_cropped.pdf}}
    \caption{[Exams dataset] Comparison of feature importance provided by  KernelSHAP and \ourmethod.}
    \label{fig:Exams}
\end{figure}

\david{ For each dataset, we first fit a black-box model, then we compute both \ourmethod and SHAP. What we were looking for is a sort of similarity between the set of results. As example, the first $K$ features of both the method should be pretty much the same, and the importance barplot should be at least similar. We must remember that \ourmethod is not an approximation or a simplification of KernelSHAP, but has a different method to compute the impact of each feature, and this could lead to some differences in the order, however the major sense of how the model works should be the same. The two methods results are absolutely comparable for all the datasets (Figure \ref{fig:Wine} -  \ref{fig:Exams}), revealing how both of the techniques understand the model predictions.  }

\end{document}